%% file: main.tex
\documentclass{article}

\usepackage[preprint]{neurips_2026}

\usepackage{tcolorbox}
\usepackage[utf8]{inputenc}
\usepackage[T1]{fontenc}
\usepackage{courier}
\usepackage{hyperref}
\usepackage{url}
\hypersetup{
  hidelinks,
  pdftitle={No Free Swap: Protocol-Dependent Layer Redundancy in Transformers},
  pdfauthor={Gabriel Garcia},
}
\usepackage{placeins}
\usepackage{etoolbox}
\usepackage{caption}
\newenvironment{tabhere}{%
  \par\medskip
  \begin{center}
}{%
  \end{center}
  \medskip
}
\newenvironment{fighere}{%
  \par\medskip
  \begin{center}
}{%
  \end{center}
  \medskip
}
\AtBeginDocument{%
  \raggedbottom
  \setcounter{topnumber}{2}%
  \setcounter{bottomnumber}{0}%
  \clubpenalty=10000
  \widowpenalty=10000
  \displaywidowpenalty=10000
  \pretocmd{\subsection}{\FloatBarrier}{}{}%
}
\usepackage{booktabs}
\usepackage{array}
\usepackage{multirow}
\usepackage{makecell}
\usepackage{tabularx}
\usepackage{amsfonts}
\usepackage{amsmath}
\usepackage{amssymb}
\usepackage{nicefrac}
\usepackage{microtype}
\usepackage{xcolor}
\usepackage{amsthm}
\usepackage{graphicx}
\usepackage{tikz}
\usepackage{subcaption}
\newtheorem{definition}{Definition}
\newtheorem{remark}{Remark}

\title{No Free Swap: Protocol-Dependent Layer Redundancy in Transformers}

\author{
  Gabriel Garcia \\
  Independent Researcher \\
  \texttt{gpgabriel25@gmail.com}
}

\begin{document}

\maketitle

\begin{abstract}
When researchers ask whether two transformer layers are ``equivalent'' for compression, they often conflate distinct tests. \emph{Replacement} asks whether one layer's map can substitute for another's in place; \emph{interchange} asks whether two layers approximately commute when their positions are swapped. Both are output-grounded swap-KL probes, but they need not agree: on pretrained transformers the protocol gap can change which layers look safe to prune by several-fold under the same evaluator, especially when replacement distances are high.

We measure both protocols across checkpoints and architectures. On a Pythia training trajectory (410M and 1.4B; Figure~\ref{fig:protocol_gap_dist}), the replacement--interchange gap grows from initialization to convergence. Under one matched WikiText-2 contract at 8B scale, Qwen3-8B enters a divergent regime: interchange-guided removal is several-fold safer than replacement-guided at the same layer budgets, while Llama-3.1-8B ties the two protocols for pruning cost even though interchange KL is lower, showing metric gaps need not map one-to-one to removal. Before layer removal or merging, score both swap-KLs on the target checkpoint; the diagnostic requires only unlabeled forward passes.
\end{abstract}

%% ============================================================
\section{Introduction}
\label{sec:intro}
%% ============================================================

Large language models exhibit layer-level redundancy, but calling two layers ``equivalent'' is incomplete without stating \emph{how} equivalence is tested. Replacement and interchange are distinct output-grounded tests; their disagreement is model- and pair-dependent, and in high-replacement-distance regimes it materially changes zero-shot pruning choices.

We study two swap protocols on pretrained checkpoints. \emph{Replacement} copies one layer's weights into another slot while the rest of the stack is unchanged. \emph{Interchange} exchanges two layers' weights so each block still runs once but at swapped depth. Both are evaluated as empirical swap-KL on finite prompts (\S\ref{sec:method}, \S\ref{sec:experiments}), as behavioral probes rather than process-algebra certificates~\citep{milner1989communication, park1981concurrency}. Intuitively, replacement probes whether one residual map can \emph{substitute} for another in place, while interchange probes whether two maps approximately \emph{commute} at nearby depths~\citep{veit2016residual}; the two can diverge sharply depending on how perturbations propagate.

The headline contrast is operational. Under one matched WikiText-2 contract at 8B scale, Qwen3-8B enters a regime where replacement and interchange rank different removals and interchange-guided greedy removal is several-fold safer than replacement-guided removal at the same budgets (Table~\ref{tab:8b_core}). Llama-3.1-8B under the same evaluator ties the two swap protocols for pruning cost even though interchange KL is uniformly below replacement KL on adjacent pairs, illustrating that metric gaps need not translate one-to-one into removal gaps when replacement distances are already small.

Beyond pruning, layer similarity is used for merging, interpretability~\citep{elhage2021mathematical}, and architectural comparison~\citep{raghu2021vision}; each area often assumes an implicit protocol. Appendix~\ref{sec:app_roadmaps} collects a compact claim--evidence roadmap, the protocol taxonomy, evaluator-contract summaries, and several auxiliary baselines so the main text can stay focused on definitions, the dense GPT-2 agreement regime, the matched 8B benchmark (Tables~\ref{tab:8b_core} and~\ref{tab:skip_qwen}), and the training-scale analyses in \S\ref{sec:exp_scaling}. Table~\ref{tab:comparison_contract} (\S\ref{sec:exp_skip}) summarizes how secondary comparisons line up on evaluator and scorer budgets.

For deployment, measure both protocols on the target checkpoint before layer removals or merges. Appendix~\ref{sec:app_matched_budget} records optional interchange-seeded beam and matched-budget SLEB extensions.

Our contributions are as follows:
\begin{enumerate}
    \item Layer equivalence is protocol-dependent. Replacement, interchange, deletion, and weight averaging test distinct notions (substitutability, commutativity, marginal contribution, parameter compatibility). They need not agree; weight averaging can fail by orders of magnitude in KL even when swap distances are low (Appendix~\ref{sec:app_negative}).
    \item Training shapes protocol gaps. A Pythia-410M/1.4B checkpoint trajectory shows the replacement--interchange \emph{metric} gap grow from initialization to convergence and concentrate on nearby pairs (\S\ref{sec:exp_scaling}), alongside static multi-width rows in Table~\ref{tab:scaling}. Controlled ablations vary positional encoding while holding architecture fixed; depth, pair distance, and training duration line up most cleanly with the effect (Appendix~\ref{sec:app_pe_ablation}).
    \item Protocol choice affects pruning under matched evaluation. On Qwen3-8B vs.\ Llama-3.1-8B under one contract (\S\ref{sec:exp_skip}), swap protocols diverge vs.\ tie for greedy removal in line with replacement-distance regimes (Tables~\ref{tab:8b_core} and~\ref{tab:skip_qwen}). Mistral-7B-v0.1 matches the tied regime under the same evaluator; Qwen2.5-7B appears only in the broader sweep of \S\ref{sec:exp_scaling}.
    \item A practical diagnostic. We recommend scoring both protocols before compression; Table~\ref{tab:method_requirements} summarizes compute, data, and calibration needs relative to common baselines.
\end{enumerate}

\subsection{Protocol vocabulary}
\label{sec:taxonomy}
Four verbs recur: \emph{replacement} (substitutability in slot), \emph{interchange} (depth swap), \emph{deletion} (skip-layer removal), and \emph{averaging} (parameter merge). The replacement--interchange contrast is primary; deletion and averaging provide complementary stress tests that the same layers can look similar under one protocol and incompatible under another.

Replacement and interchange are computed via swap-KL distance (Definition~\ref{def:swap_kl} and Remark~\ref{rem:interchange}); deletion is measured by forward-pass perplexity after layer removal; and weight averaging is evaluated by interpolating two layers' weight matrices and measuring output KL. Throughout this paper, we focus primarily on the replacement-interchange contrast, which reveals a training-induced protocol gap with direct consequences for pruning. Deletion and averaging provide additional evidence that protocol disagreement is structural, not an artifact of swap-KL.

\section{Background}
\label{sec:background}
%% ============================================================

\subsection{Swap-KL Distance}
\label{sec:bg_swap}

Layer-swap testing (comparing a model before and after substituting one layer's weights or position for another's) provides a direct, output-grounded measure of layer interchangeability.
We use KL divergence of model output distributions to quantify how much a swap changes the model's predictions; this avoids intermediate-representation proxies (CKA, BI score~\citep{men2024shortgpt}) in favor of a direct behavioral test.
This distance-based view is motivated by process-algebra interchangeability theory~\citep{park1981concurrency,milner1989communication}, but our approach is fully empirical: no formal guarantees follow at LLM scale.

\subsection{Transformer Architecture}
\label{sec:bg_transformer}

A transformer language model~\citep{vaswani2017attention} consists of an embedding layer, a sequence of $L$ identical-structure transformer blocks, and an output projection. Let $x_0$ denote the output of the embedding layer. Each block $\ell \in \{0, \ldots, L{-}1\}$ applies multi-head self-attention followed by a position-wise feed-forward network (MLP), with residual connections and layer normalization:
\begin{align}
    x_\ell' &= x_\ell + \text{MHA}_\ell(x_\ell) \\
    x_{\ell+1} &= x_\ell' + \text{MLP}_\ell(x_\ell')
\end{align}
Crucially, all blocks share the same \emph{architecture} but are parameterized by \emph{different} learned weight matrices. This structural homogeneity is what makes layer-level swap-KL analysis possible: since blocks are architecturally interchangeable, the question reduces to whether their \emph{learned parameters} produce equivalent functions in context.

\subsection{Layer Redundancy in Transformers}
\label{sec:bg_redundancy}

Several lines of evidence suggest that transformer layers, particularly in the middle of the network, compute highly similar functions. Centered Kernel Alignment (CKA)~\citep{kornblith2019similarity} applied to transformer representations reveals that middle layers produce highly correlated feature maps, with CKA values approaching 1.0 for adjacent layers~\citep{raghu2021vision}. ShortGPT's Block Influence analysis~\citep{men2024shortgpt} found that middle layers have the lowest influence scores, meaning they contribute least to the overall representation change. This pattern is consistent across model families including LLaMA, Pythia~\citep{biderman2023pythia}, and OPT~\citep{zhang2022opt}.

However, CKA and BI scores measure \emph{representation similarity}, not \emph{functional interchangeability}. Sections~\ref{sec:exp_skip} and~\ref{sec:exp_laco_sleb} provide direct evidence of this disagreement: on Qwen3-8B at $n{=}1$, CKA selects layer~7 (+9.6\% PPL) while interchange selects layer~17 (+2.5\%); at $n{=}3$, the gap widens to CKA +24.3\% vs. interchange +10.2\%. Representation-similar pairs are not necessarily functionally redundant. Two layers could produce very similar CKA scores but compute different functions that happen to agree on typical inputs. Our swap-KL distance directly estimates functional interchangeability by asking: does swapping one layer for another change the model's \emph{output distribution}? Public checkpoints carry PE-type labels (absolute, rotary, ALiBi) that correlate with distance summaries in our slice, but controlled runs in Appendix~\ref{sec:app_pe_ablation} and \S\ref{sec:exp_scaling} treat PE as descriptive context rather than a standalone mechanism.

%% ============================================================
\section{Method}
\label{sec:method}
%% ============================================================

\subsection{Swap-KL Distance Between Layers}
\label{sec:method_distance}

Let $\mathcal{M}$ denote a pretrained transformer with $L$ residual blocks (indexed $0,\ldots,L{-}1$ in code and figures), and let $p_\mathcal{M}(y \mid x)$ denote its next-token distribution (softmax probabilities induced by the final logits) given input $x$. We define the \emph{swap operation} $\text{swap}(i, j)$ as replacing the weights of layer $i$ with those of layer $j$, producing a modified model $\mathcal{M}_{i \leftarrow j}$.

\begin{definition}[Empirical Swap-KL Distance]
\label{def:swap_kl}
The swap-KL distance between layers $i$ and $j$ over a prompt set $\mathcal{X}$ is:
\begin{align}
    d_{\text{repl}}(i, j) = \max\Big(&\mathbb{E}_{x \in \mathcal{X}}\!\big[\text{KL}\!\left(p_\mathcal{M}(\cdot \mid x) \;\|\; p_{\mathcal{M}_{i \leftarrow j}}(\cdot \mid x)\right)\big],\nonumber\\
    &\mathbb{E}_{x \in \mathcal{X}}\!\big[\text{KL}\!\left(p_\mathcal{M}(\cdot \mid x) \;\|\; p_{\mathcal{M}_{j \leftarrow i}}(\cdot \mid x)\right)\big]\Big)
\end{align}
\end{definition}

Figure~\ref{fig:swap_protocol} diagrams the two swap-KL protocols compared throughout the paper. Note that $d_{\text{repl}}$ uses the \emph{max} of the two directed KL measurements (i.e., the worse direction), whereas interchange is symmetric by construction. Using mean-symmetrization for replacement would reduce $d_{\text{repl}}$ and increase the I/R ratio further, so the asymmetric max convention is conservative: it understates how much worse replacement-guided selection is relative to interchange. Skip-layer \emph{removal} (deleting a block from the stack) is a separate intervention evaluated on perplexity in \S\ref{sec:exp_skip}: Table~\ref{tab:skip_qwen} (Qwen3-8B full method grid), Table~\ref{tab:skip_llama} (Llama-3.1-8B), and Table~\ref{tab:8b_core} (three-model $\Delta$PPL summary including Mistral).

\begin{fighere}
\resizebox{\textwidth}{!}{\input{figures/swap_protocol.tex}}

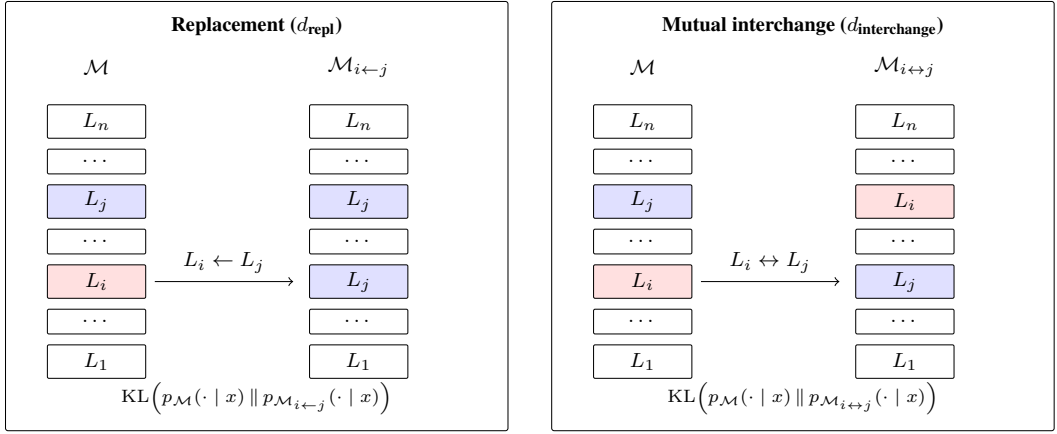
\captionof{figure}{Operational schematic of the two swap-KL protocols (Definition~\ref{def:swap_kl} and Remark~\ref{rem:interchange}). \emph{Left} (replacement, $d_{\text{repl}}$): weights of $L_i$ are overwritten by those of $L_j$ while other layers are unchanged, giving $\mathcal{M}_{i\leftarrow j}$; KL compares next-token distributions to the baseline model. \emph{Right} (mutual interchange, $d_{\text{interchange}}$): $L_i$ and $L_j$ exchange weights so both blocks still run once, reordered, giving $\mathcal{M}_{i\leftrightarrow j}$. This replacement-vs-interchange contrast drives Table~\ref{tab:scaling} and the pruning selectors in \S\ref{sec:exp_skip}. Skip-layer removal is not shown here; see \S\ref{sec:method_skip}.}
\label{fig:swap_protocol}
\end{fighere}

\noindent The $\max$ over both swap directions ensures a conservative estimate: two layers are approximately swap-similar only if swapping in \emph{either} direction preserves the output distribution. This asymmetry matters because a layer earlier in the network may tolerate replacement more easily than a later layer that depends on specific upstream representations.

\paragraph{Sensitivity to the choice of symmetrization.}
A natural concern is that taking the maximum of two directional KLs (rather than, e.g., a mean or geometric mean) may artificially inflate the protocol gap between $d_{\text{repl}}$ and the symmetric-by-construction $d_{\text{interchange}}$. We verify on Qwen3-8B (Table~\ref{tab:asymmetry}) that while the per-pair max/min asymmetry ratio is non-trivial (median $1.48$, 90th percentile $2.68$), the \emph{ranking} of layer pairs by swap-KL similarity is essentially invariant: Spearman correlation between the $\max$ definition and either symmetric alternative exceeds $0.94$, and the top-$5$ most-swap-similar pairs picked under each definition agree on $5/5$ entries. The downstream pruning targets the paper relies on are therefore robust to this definitional choice; the absolute scale of the reported replacement distances shifts, but the ordering does not (Table~\ref{tab:asymmetry} also records why the adjacent-pair count is $32$ rather than $35$ for this diagnostic JSON).

\input{figures/d_repl_asymmetry.tex}

\begin{remark}[Position Interchange Distance]
\label{rem:interchange}
An alternative test is \emph{mutual position interchange}: simultaneously replacing layer~$i$'s weights with layer~$j$'s \emph{and} layer~$j$'s weights with layer~$i$'s, so each layer still executes once but at the other's depth. We denote the resulting model $\mathcal{M}_{i \leftrightarrow j}$ and define (note: $\mathcal{M}_{i \leftrightarrow j} = \mathcal{M}_{j \leftrightarrow i}$ by construction, so interchange is symmetric and requires no max-symmetrization)
\begin{equation}
    d_{\text{interchange}}(i, j) = \mathbb{E}_{x \in \mathcal{X}}\!\big[\text{KL}\!\left(p_\mathcal{M}(\cdot \mid x) \;\|\; p_{\mathcal{M}_{i \leftrightarrow j}}(\cdot \mid x)\right)\big].
\end{equation}
This tests whether the \emph{order} of two layers matters, rather than whether one can \emph{replace} the other. Because no layer's computation is eliminated (both still run, just at swapped positions), interchange distance is structurally less disruptive than swap-KL distance. Empirically, we observe $d_{\text{interchange}}(i,j) \leq d_{\text{repl}}(i,j)$ across all tested model-pair combinations; Remark~\ref{rem:interchange_order} provides an illustrative analysis under simplifying assumptions. As we show in \S\ref{sec:exp_scaling}, the gap between these two metrics is itself informative: in our model suite it is negligible for models with absolute positional embeddings and largest for the deep, well-trained RoPE models. This pattern is driven primarily by layer distance and training duration (controlled ablations confirm PE type is a correlated secondary factor, not the proximate cause; \S\ref{sec:exp_scaling}).
\end{remark}

\begin{remark}[Interchange perturbation is empirically smaller]
\label{rem:interchange_order}
Across all tested model-pair combinations, $d_{\text{interchange}}(i,j) \leq d_{\text{repl}}(i,j)$. This holds because interchange tests layer \emph{order} (both layers still execute), not layer \emph{equivalence} (which eliminates one); the perturbation is structurally less disruptive. A mechanistic analysis under simplifying assumptions (showing that the perturbation difference is $O(\|J\|\cdot\|g\|)$ vs.\ $O(\|g\|)$ for replacement) and empirical Jacobian measurements across GPT-2-Medium and Pythia-410M are in Appendix~\ref{sec:app_jacobian}. In checkpoints with expansive deep-layer Jacobians (e.g., pretrained Pythia-410M in Appendix~\ref{sec:app_jacobian}), replacement-side perturbations can be amplified along the depth axis while interchange cross-terms remain comparatively small when layers are approximately structurally commutative in depth; this is one Jacobian-level account of a protocol gap \emph{where it appears}, not a claim that positional-encoding \emph{type} is the proximate cause (see Appendix~\ref{sec:app_pe_ablation}).
\end{remark}

\paragraph{Relationship to formal process algebra.}
Layer-swap testing was motivated by quantitative process-algebra theory~\citep{park1981concurrency,milner1989communication,desharnais2004metrics}, which defines interchangeability via a supremum over all inputs. Our Definition~\ref{def:swap_kl} replaces this with an expectation over a finite prompt set~$\mathcal{X}$, making it an empirical approximation; our claims are empirical, not formal guarantees. Throughout this paper, ``swap-KL distance'' refers to this empirical approximation. Bootstrap CIs over 1{,}000 resamples confirm ranking stability (all top-10 pair CIs have relative width~$\leq$\,10\%; \S\ref{sec:exp_full}).

We classify layer pairs into three categories based on their swap-KL distance:
\begin{itemize}
    \item \emph{Strongly swap-similar} ($d_{\text{repl}} < 0.05$): The swap produces negligible distributional change. The layers appear functionally interchangeable on the test distribution.
    \item \emph{Conditionally swap-similar} ($0.05 \leq d_{\text{repl}} < 0.10$): The swap produces small but measurable change. Compression may be viable with fine-tuning.
    \item \emph{Non-swap-similar} ($d_{\text{repl}} \geq 0.10$): The layers compute meaningfully different functions.
\end{itemize}

\subsection{Head-Level Analysis}
\label{sec:method_head}

To investigate whether the protocol-gap pattern is driven by specific attention heads, we replace only one head's weights (query, key, value, and output projections) in layer~$i$ with those from layer~$j$, measuring KL divergence per head.

\subsection{Layer-removal evaluation protocol}
\label{sec:method_skip}

Given that swap-similar layers are functionally redundant, we remove one layer from each swap-similar pair by deleting it from the model's \texttt{ModuleList}. The resulting model is evaluated on WikiText-2~\citep{merity2016pointer} perplexity using a sliding window of 1024 tokens with stride 512.

%% ============================================================
\section{Experiments}
\label{sec:experiments}
%% ============================================================

We organize evidence in three blocks: GPT-2-Medium as a dense \emph{protocol-agreement} contrast where both swap protocols align (\S\ref{sec:exp_full}); matched Qwen/Llama/Mistral pruning under one evaluator (\S\ref{sec:exp_skip}); and training trajectory plus multi-model scaling rows (\S\ref{sec:exp_scaling}). GPT-2-Medium (355M, AbsPE) evaluates all 276 pairs on CPU (fp32); Qwen3-8B and Llama-3.1-8B use JAX on TPU v6e (bf16) with both replacement (Definition~\ref{def:swap_kl}) and interchange (Remark~\ref{rem:interchange}). \emph{Perplexity baselines} can differ slightly across tables when the underlying evaluator instance or window differs; every reported $\Delta$PPL row uses the baseline stated in that same table.

\subsection{GPT-2-Medium: dense agreement-regime contrast}
\label{sec:exp_full}

We compute the swap-KL distance for all 276 layer pairs in GPT-2-Medium using 100 diverse prompts of 128 tokens each. Table~\ref{tab:top_pairs} presents the top-10 most swap-similar pairs.

\begin{tabhere}
\small
\begin{tabular}{lcccccl}
\toprule
Rank & Layer $i$ & Layer $j$ & Gap & Mean KL$^*$ & Max KL & Sym.\ Mean [95\% CI]$^\dagger$ \\
\midrule
1 & 4 & 5 & 1 & 0.036 & 0.25 & 0.00101 [0.00098, 0.00105] \\
2 & 6 & 7 & 1 & 0.036 & 0.32 & 0.00043 [0.00041, 0.00044] \\
3 & 5 & 6 & 1 & 0.038 & 0.56 & 0.00158 [0.00151, 0.00167] \\
4 & 6 & 8 & 2 & 0.039 & 0.26 & 0.00076 [0.00073, 0.00078] \\
5 & 5 & 7 & 2 & 0.039 & 0.21 & 0.00152 [0.00146, 0.00159] \\
6 & 7 & 8 & 1 & 0.040 & 0.26 & 0.00077 [0.00075, 0.00080] \\
7 & 4 & 6 & 2 & 0.042 & 1.00 & 0.00091 [0.00087, 0.00095] \\
8 & 5 & 8 & 3 & 0.042 & 0.27 & 0.00212 [0.00206, 0.00219] \\
9 & 9 & 11 & 2 & 0.043 & 0.21 & 0.00055 [0.00054, 0.00057] \\
10 & 4 & 7 & 3 & 0.043 & 0.41 & 0.00103 [0.00100, 0.00106] \\
\bottomrule
\end{tabular}
\vspace{2pt}
\small{$^*$Bidirectional max of per-direction means (Definition~\ref{def:swap_kl}); this is the conservative ranking statistic used for the Rank column. $^\dagger$Per-prompt symmetrized mean KL (average of both swap directions per prompt); shown with its bootstrap 95\% CI for stability within each row. The symmetrized statistic is much smaller in magnitude than the max because the max stresses the worse direction while the symmetrized mean averages them. \emph{Global agreement:} Spearman $\rho > 0.99$ between Mean KL$^*$ and Sym.\ Mean rankings over all 276 pairs; the Sym.\ Mean column is not sorted and can invert locally relative to Rank.}
\captionof{table}{Top-10 most swap-similar layer pairs in GPT-2-Medium (355M, 24 layers), ordered by Mean KL$^*$ (bidirectional max, Definition~\ref{def:swap_kl}). Sym.\ Mean is a per-prompt symmetrized diagnostic used for bootstrap stability; it is \emph{not} the sort key, so it need not decrease monotonically with Rank (e.g., ties on Mean KL$^*$ at ranks~1--2, or local inversions within this top-$10$ slice). Over all 276 pairs, rankings induced by Mean KL$^*$ and by Sym.\ Mean agree closely (Spearman $\rho > 0.99$). Bootstrap 95\% CIs on Sym.\ Mean have relative widths $\leq$\,11\% with no adjacent-rank overlap.}
\label{tab:top_pairs}
\end{tabhere}

Of the 276 pairs tested, 18 are strongly swap-similar (mean KL\,$<$\,0.05) and 98 total fall below the conditional threshold (mean KL\,$<$\,0.10, i.e., 35.5\% of all pairs). Nine of the top-10 pairs in Table~\ref{tab:top_pairs} lie entirely within layers 4--8 and have Mean KL$^*$$\leq$\,0.043 nats; the only top-10 exception is pair 9$\leftrightarrow$11 (rank~9). This reveals a densely interconnected swap-similar \emph{core} rather than isolated adjacent pairs. The cluster extends to layers 3--12 with gradually increasing distances. The best pair is layers 4$\leftrightarrow$5 (mean KL\,$=$\,0.036, max KL\,$=$\,0.25). Of the 23 adjacent pairs, 22 are conditionally swap-similar (mean KL\,$<$\,0.10); only the boundary pair 0$\leftrightarrow$1 (KL\,$=$\,2.60) is non-swap-similar. The first and last layers (0 and 23) are highly specialized, with all pairs involving these layers exhibiting KL\,$>$\,2.0.

\paragraph{Mean vs.\ max KL divergence.}
Table~\ref{tab:top_pairs} reports both mean and max KL. The best pair (4$\leftrightarrow$5) has mean KL\,$=$\,0.036 but max KL\,$=$\,0.25, a 6.9$\times$ ratio arising from tail prompts with unusual token distributions. We use the mean rather than the supremum because (i) the true swap-KL distance requires a supremum over all inputs, which is uncomputable; the expectation is a lower-variance estimator, and (ii) the \emph{relative ordering} is robust: Spearman $\rho > 0.92$ between mean-based and max-based rankings across all 276 pairs.

\paragraph{Prompt-set robustness and bootstrap confidence intervals.} We validate stability by comparing rankings at $N{=}20$ vs.\ $N{=}100$ diverse prompts: Spearman $\rho = 0.92$ across all 276 pairs, with the top pair (4$\leftrightarrow$5) stable at both sizes (0.035 vs.\ 0.036). Bootstrap confidence intervals (1{,}000 resamples over 100 prompts) for the top-10 pairs show relative CI widths of 5--11\% on the symmetrized mean KL, with no adjacent-rank intervals overlapping (Table~\ref{tab:top_pairs}). Globally, rankings induced by the bidirectional max (Mean KL$^*$) and by the symmetrized mean remain tightly aligned (Spearman $\rho > 0.99$ over all 276 pairs), even though Table~\ref{tab:top_pairs} sorts rows only by Mean KL$^*$. Appendix~\ref{sec:app_prompt_robustness} provides a detailed convergence analysis at intermediate sizes. For practitioners, we recommend a simple prompt-budget check: compute rankings on an initial diverse set, then recompute after adding another equally sized tranche; if top-ranked pairs and coarse ordering change materially, increase prompt count before making pruning decisions.

The spatial structure of the distance matrix (Figure~\ref{fig:heatmap}) reveals a clear block-diagonal pattern in the early-to-middle layers. Layers 3--12 form a contiguous cluster where most within-cluster pairs have KL\,$<$\,0.10, extending the layer-4--8 concentration from Table~\ref{tab:top_pairs} into a broader middle band. Layers 0--2 and 13--23 form distinct blocks with higher inter-block distances.

\begin{fighere}
\includegraphics[width=0.82\textwidth]{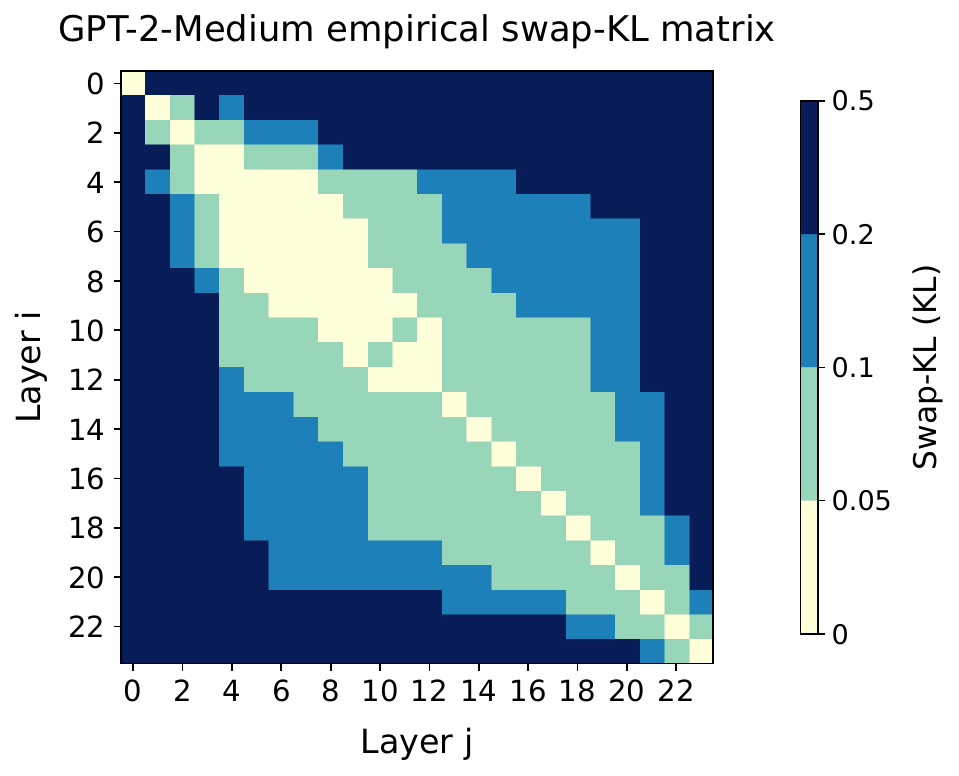}
\captionof{figure}{Swap-KL distance matrix for GPT-2-Medium (24 layers). Values capped at 0.5 for visibility (YlGnBu colormap: pale yellow is low KL, saturated blue is high KL). The block-diagonal structure in layers 3--12 indicates a dense cluster of approximately swap-similar middle layers; the closest-ranked pairs concentrate in layers 4--8 (Table~\ref{tab:top_pairs}). Layer~0 and the late-boundary layers show the largest distances (darkest blue in this palette).}
\label{fig:heatmap}
\end{fighere}

This pattern is consistent with the hypothesis that early layers specialize in token embedding and positional processing, late layers specialize in output prediction, and middle layers perform a more generic ``refinement'' computation that is approximately the same function repeated across multiple positions in the layer stack.

\paragraph{Skip-layer removal cost (GPT-2-Medium).}
We evaluate the immediate perplexity cost of removing layers identified as swap-similar. Table~\ref{tab:skip} presents WikiText-2 perplexity for several skip configurations.

\begin{tabhere}
\begin{tabular}{lccc}
\toprule
Configuration & Layers Removed & PPL & $\Delta$ PPL (\%) \\
\midrule
Baseline (24 layers) & --- & 19.19 & --- \\
Skip layer 5 & \{5\} & 20.17 & +5.1 \\
Skip layer 15 & \{15\} & 20.23 & +5.4 \\
Skip layer 13 & \{13\} & 20.43 & +6.5 \\
Skip layers 5, 15 & \{5, 15\} & 21.45 & +11.8 \\
Skip layers 5, 13, 15 & \{5, 13, 15\} & 23.46 & +22.3 \\
\bottomrule
\end{tabular}
\captionof{table}{WikiText-2 perplexity ($\downarrow$) after removing layers from GPT-2-Medium. The 24-layer baseline has PPL\,$=$\,19.19. Single-layer removal costs 5--6\%; multi-layer removal is superlinear. Values are deterministic forward passes on the full validation split.}
\label{tab:skip}
\end{tabhere}

Single-layer removal from swap-similar pairs costs approximately 5--6\% perplexity, confirming that these layers are approximately but not exactly redundant. Removing layer~5 from the best-ranked pair 4$\leftrightarrow$5 (Table~\ref{tab:top_pairs}) costs 5.1\%, nearly identical to removing layer~15 (+5.4\%; Table~\ref{tab:skip}), even though layer~15 lies outside the top-10 pair slice of Table~\ref{tab:top_pairs}. This consistency suggests that swap-KL distance remains a useful predictor of removal cost beyond the single best-ranked pair.

Multi-layer removal exhibits superlinear degradation: removing two layers costs 11.8\% (vs.\ 10.5\% if additive), and three layers cost 22.3\% (vs.\ 16.5\%). This superlinearity indicates that approximate layer equivalence has cumulative effects: removing one layer slightly perturbs the representations that downstream layers depend on, making subsequent removals more costly. This observation constrains the number of layers that can be safely removed without fine-tuning; however, LoRA recovery (rank~16, 200~steps) recovers 48--60\% of the pruning-specific perplexity increase on Qwen3-8B after controlling for domain adaptation (Appendix~\ref{sec:app_finetune}). Additional downstream task benchmarks showing only 1.6\% mean accuracy loss (Appendix~\ref{sec:app_downstream}) and a multi-layer compression sweep against random and anti-guided strategies (Appendix~\ref{sec:app_compression_sweep}) further validate interchange scoring as a reliable layer pruning signal.

\subsection{Matched 8B pruning benchmark}
\label{sec:exp_skip}
\FloatBarrier

\paragraph{Protocol gap in compression guidance (Pythia-1.4B).} The protocol gap (\S\ref{sec:exp_scaling}) has direct consequences for which layers a pruning method recommends removing. On Pythia-1.4B (24 layers, RoPE, baseline PPL\,$=$\,13.54), interchange distance ranks late layers as most redundant (best pair: 19$\leftrightarrow$20, KL\,$=$\,0.023), while replacement distance ranks mid layers (best pair: 9$\leftrightarrow$10, KL\,$=$\,0.107). Using each ranking to select removal targets produces strikingly different outcomes: interchange-guided removal degrades +9.8\% at $n{=}1$, +25.0\% at $n{=}2$, and +63.0\% at $n{=}3$; replacement-guided removal degrades +18.0\%, +45.8\%, and +110.5\% respectively, roughly 1.8$\times$ more degradation at every compression point. This confirms that the protocol gap is not merely a measurement artifact but translates directly into compression quality differences on this checkpoint; interchange is the more effective protocol for the Pythia-1.4B ranking we evaluate.

\paragraph{Cross-family replication (Llama-3.1-8B).}
To test whether the protocol gap is an artifact of the Qwen model family, we replicate the head-to-head comparison on Llama-3.1-8B (32 layers, GQA 4:1, TPU v6e, bf16) under the \emph{same matched evaluator} as Qwen3-8B (WikiText-2 test split, 5K words, window$=$512, stride$=$256; baseline PPL\,$=$\,8.31). Table~\ref{tab:skip_llama} reports all results.

\begin{tabhere}
\small
\resizebox{\textwidth}{!}{%
\begin{tabular}{clccc}
\toprule
$n$ & Method & Layers Removed & PPL & $\Delta$ PPL (\%) \\
\midrule
\multirow{4}{*}{1} & Interchange / Replacement / SLEB & \{25\} / \{11\} / \{11\} & 9.07 & \textbf{+9.2} \\
& Random & \{3\} & 9.99 & +20.2 \\
& BI & \{2\} & 10.31 & +24.1 \\
\midrule
\multirow{5}{*}{3} & SLEB (iterative \& greedy) & \{10, 11, 25\} & 10.66 & \textbf{+28.3} \\
& Replacement-guided & \{9, 10, 11\} & 11.55 & +39.1 \\
& Interchange-guided & \{24, 25, 26\} & 11.65 & +40.3 \\
& Random & \{14, 19, 22\} & 11.66 & +40.4 \\
& BI-guided & \{2, 3, 5\} & 48.98 & +489.5 \\
\midrule
\multirow{6}{*}{5} & SLEB (iterative) & \{10, 11, 12, 25, 28\} & 13.06 & \textbf{+57.2} \\
& SLEB (greedy) & \{8, 9, 10, 11, 25\} & 13.90 & +67.3 \\
& Replacement-guided & \{7, 8, 9, 10, 11\} & 15.35 & +84.7 \\
& Interchange-guided & \{22, 23, 24, 25, 26\} & 15.37 & +85.0 \\
& Random & \{3, 6, 16, 19, 29\} & 20.12 & +142.1 \\
& BI-guided & \{2, 3, 4, 5, 6\} & 71.84 & +764.7 \\
\bottomrule
\end{tabular}}
\captionof{table}{WikiText-2 perplexity ($\downarrow$) after removing $n$ layers from Llama-3.1-8B (32 layers, baseline PPL\,$=$\,8.31). Same matched evaluator as Table~\ref{tab:skip_qwen} (WikiText-2 test split, 5K words / 5585 tokens, window$=$512, stride$=$256, JAX bf16 on TPU v6e-8). Best method bold per~$n$. Both swap protocols achieve comparable degradation, consistent with low replacement KL on most adjacent pairs (Table~\ref{tab:scaling}). BI collapses at multi-layer budgets. Swap-KLs use 100 prompts$\times$64 tokens (vs.\ 500$\times$128 for Qwen in Table~\ref{tab:skip_qwen}); Appendix~\ref{sec:app_prompt_robustness} reports sensitivity to prompt count.}
\label{tab:skip_llama}
\end{tabhere}

On all 31 adjacent pairs, interchange KL is strictly below replacement KL. Pooling as a per-pair ratio $\mathrm{KL}_{\mathrm{inter}}/\mathrm{KL}_{\mathrm{repl}}$ yields mean $0.30$ and median $0.27$ (Table~\ref{tab:scaling}); these are \emph{not} mean/median replacement KL. Replacement KL is small across most of the stack: 27 of 31 adjacent pairs fall below $0.05$ (best: 11$\leftrightarrow$12, KL$=$0.005), while a few boundary-adjacent pairs carry much larger replacement KL, so the median replacement KL over gap-1 pairs (${\sim}1.4{\times}10^{-2}$ nats on our released lattice) is far below the mean. Interchange distances are tightest in layers 22--26 (best: 25$\leftrightarrow$26, KL$=$0.001). This contrasts with Qwen3-8B under our strengthened budget, where no replacement pair falls below $0.10$. The Llama pattern explains why replacement-guided removal is competitive for pruning: replacement KL is small enough on most pairs to act as a selector comparable to interchange, even though the two protocols still disagree on relative pair rankings.

Bootstrap uncertainty estimates confirm that this Llama tie is not a small-sample artifact. Over 1{,}000 bootstrap resamples of the matched evaluator windows, the $n{=}3$ variants remain statistically indistinguishable: interchange-guided removal yields PPL\,$=$\,11.65 with 95\% CI [10.49, 13.10], while replacement-guided yields 11.55 [10.05, 13.40]. The same holds at $n{=}5$: interchange 15.37 [14.03, 16.91] vs.\ replacement 15.35 [13.27, 17.89]. Thus the Llama result is not merely under-validated noise; it is a genuinely different regime in which both output-grounded protocols select comparably safe removals.

On Llama, both protocols achieve comparable PPL at all budgets: interchange-guided removal degrades +9.2/+40.3/+85.0\% at $n{=}1/3/5$, while replacement-guided degrades +9.2/+39.1/+84.7\%, effectively a tie, in stark contrast to the Qwen3-8B pattern where interchange dominated by 2.6--4.7$\times$. SLEB-iterative outperforms both swap protocols at $n{\geq}3$ (+28.3/+57.2\% vs.\ ${\sim}$40/85\%), confirming that when most adjacent replacement KLs are low, calibration-based methods leverage iterative recalibration to achieve superior selections. The decisive comparison is against protocol-agnostic methods: BI-guided removal collapses catastrophically (+24.1/+489.5/+764.7\%), and random removal degrades +20.2/+40.4/+142.1\%.

Mistral-7B-v0.1 (Table~\ref{tab:8b_core}, third column) matches the tied \emph{pruning} regime under the same evaluator but lacks bootstrap CIs; its headline table uses the same interchange-guided layer sets to report a \emph{proxy} replacement row (same removals as interchange), not an independent replacement-guided search (see table footnote). Combining the matched 8B panel with the broader sweep (which additionally includes Qwen2.5-7B at 7B scale) yields a directional picture: (1)~a deep swap-similar core with mostly low pairwise KLs appears in several RoPE checkpoints we tested, including Llama, Mistral, Qwen2.5-7B, and TinyLlama-1.1B; (2)~the high-replacement-divergence regime (Qwen3-8B, I/R$=$0.21) is a distinct protocol regime, not a universal RoPE property; (3)~BI's failure is structural across all three 8B columns: it selects early layers critical for downstream prediction regardless of model family. On Qwen3-8B where replacement distances are high, interchange substantially outperforms replacement and converges with SLEB-iterative in the clustered regime (see \S\ref{sec:exp_laco_sleb}). Evidence argues against simple architectural shortcuts: TinyLlama-1.1B (GQA 32/4 = 8:1, same ratio as Qwen3-8B) is in the tied regime (16/21 strong pairs, best KL${}=0.003$ at $11\leftrightarrow 12$), confirming that neither PE type nor GQA ratio alone predicts protocol behavior. The diagnostic must be measured empirically on the target model.

\begin{tabhere}
\footnotesize
\setlength{\tabcolsep}{4pt}
\begin{tabular}{@{}p{0.34\linewidth}>{\raggedright\arraybackslash}p{0.62\linewidth}@{}}
\toprule
\textbf{Comparison} & \textbf{Contract (matched / caveat)} \\
\midrule
8B panel (Table~\ref{tab:8b_core}) & Eval matched across Qwen/Llama/Mistral; bootstrap CIs on Qwen/Llama. Mistral replacement row is a proxy (same removals as interchange). \\
Qwen/Llama swap protocols & Eval + budget matched; swap-based methods calibration-free. \\
Interchange vs.\ SLEB-iterative & Eval matched; SLEB uses calibration (not a strict cost-fair claim). \\
Beam vs.\ SLEB (Appendix~\ref{sec:app_matched_budget}) & Partial budget match; search procedures differ. \\
Beam vs.\ SLEB matched calls (Tab.~\ref{tab:matched_budget}) & Explicit evaluator-call budget $B$; appendix only. \\
13-model sweep (Table~\ref{tab:scaling}) & Not budget-matched across rows; descriptive spread. \\
\bottomrule
\end{tabular}
\captionof{table}{Head-to-head contracts (what is matched vs.\ directional). ``Eval'' $=$ same WikiText-2 windows and $\Delta$PPL accounting; ``Budget'' $=$ same removal count $n$ and comparable scorer calls where stated.}
\label{tab:comparison_contract}
\end{tabhere}

\paragraph{8B model-conditional benchmark.}
Table~\ref{tab:8b_core} consolidates three 8B RoPE case studies under one matched evaluation contract (WikiText-2 test split, 5K words, window$=$512, stride$=$256, JAX bf16 on TPU v6e-8).

\begin{tabhere}
\small
\setlength{\tabcolsep}{5pt}
\resizebox{\linewidth}{!}{%
\begin{tabular}{@{}l rr rr@{}}
\toprule
 & \multicolumn{2}{c}{\textbf{Qwen3-8B}} & \multicolumn{2}{c}{\textbf{Llama-3.1-8B}} \\
\cmidrule(lr){2-3}\cmidrule(lr){4-5}
Method & $n{=}3$ $\Delta$\% [CI] & $n{=}5$ $\Delta$\% [CI] & $n{=}3$ $\Delta$\% [CI] & $n{=}5$ $\Delta$\% [CI] \\
\midrule
Interchange    & +10.2 [12.04,\,14.89]  & +45.4 [15.56,\,19.93]   & +40.3 [10.49,\,13.10] & +85.0 [14.03,\,16.91] \\
Replacement-guided & +48.1 [16.23,\,19.77]  & +116.5 [23.66,\,29.14]  & +39.1 [10.05,\,13.40] & +84.7 [13.27,\,17.89] \\
\midrule
I/R ratio      & \textbf{0.21}        & \textbf{0.39}         & 1.03                & 1.00                \\
\bottomrule
\end{tabular}}

\vspace{10pt}
\noindent\textit{Mistral-7B-v0.1} (same evaluator; no bootstrap CIs). Replacement columns repeat interchange removals (proxy), not an independent replacement-guided search.

\vspace{4pt}
\begin{tabular}{@{}lrr@{}}
\toprule
 & $n{=}3$ $\Delta$\% & $n{=}5$ $\Delta$\% \\
\midrule
Interchange & +24.1 & +72.3 \\
Replacement (proxy, same layers) & +24.1 & +72.3 \\
\bottomrule
\end{tabular}
\captionof{table}{Matched WikiText-2 8B panel (5K words, 512/256, TPU v6e-8). $\Delta$PPL (\%); brackets $=$ 95\% bootstrap CI on absolute PPL. I/R only when replacement is an independent replacement-guided run (Qwen/Llama). Mistral: proxy replacement row duplicates interchange removals.}
\label{tab:8b_core}
\end{tabhere}

The I/R ratio row uses $\Delta$PPL percentages where both numerator and denominator come from independent swap-protocol runs: Qwen3-8B I/R$=$0.21 at $n{=}3$ means interchange-guided removal is 4.7$\times$ less damaging than replacement; Llama-3.1-8B I/R$=$1.03 means indistinguishable swap-protocol pruning cost despite smaller interchange KL on many pairs. Mistral omits I/R because the replacement columns explicitly repeat interchange removals (not an independent replacement-guided run). Selection is model-conditional: interchange tracks SLEB-iterative on Qwen; on Llama, SLEB-iterative wins at $n{\geq}3$ while both swap protocols remain safe.

\vspace{4pt}
\begin{tcolorbox}[title={Decision rule},colback=gray!9,colframe=gray!55,boxrule=0.55pt]
Measure replacement and interchange swap distances on the target checkpoint before large layer removals or merges.
\begin{itemize}
    \item \emph{High replacement + low interchange} ($I/R \ll 1$): prefer interchange-guided selection (Qwen3-8B-like regimes).
    \item \emph{Low replacement + low interchange} ($I/R \approx 1$): either swap protocol is reasonable; calibrated iterative methods may win at larger budgets (Llama/Mistral regime).
    \item \emph{High replacement + high interchange}: swap scores alone are weak; add calibration data or another signal.
\end{itemize}
Uses only unlabeled prompts and forward passes (no task labels, no gradients, no iterative recalibration pass).
\end{tcolorbox}

\paragraph{Pythia-1.4B baseline suite.}
The full BI/CKA/SLEB/Taylor/random grid on a 10K-word WikiText-2 evaluator is Table~\ref{tab:pythia_baselines} (Appendix~\ref{sec:app_roadmaps}).

\paragraph{Non-wiki robustness.}
On a 5K-word IMDB test slice under the same 512/256 protocol, Spearman correlations of $\Delta$PPL vs.\ WikiText-2 exceed $0.98$ for both Qwen3-8B and Pythia-1.4B; details in Appendix~\ref{sec:app_imdb_robustness}.

\paragraph{Cross-model harmonized re-execution.}
Because the Qwen3-8B and Pythia-1.4B numbers reported above were produced by separate evaluator scripts at different points in the project, we re-executed both model families end-to-end under a \emph{single contract-locked} runner (\texttt{run\_harmonized\_slice.sh}) on the same TPU v6e-8 host (\texttt{wikitext/wikitext-2-raw-v1}, \texttt{split=test}, 5K words, window$=$512, stride$=$256; bf16 for Qwen, fp32 for Pythia). Table~\ref{tab:harmonized_xmodel} reports the resulting baseline perplexities (Qwen\,$=$\,12.10, Pythia\,$=$\,15.87) and head-to-head $\Delta$PPL under that lock.

\begin{tabhere}
\footnotesize
\setlength{\tabcolsep}{4pt}
\resizebox{\linewidth}{!}{%
\begin{tabular}{llrrrrrr}
\toprule
Model & Method & $n{=}1$ & $n{=}3$ & $n{=}5$ & I/R($n{=}1$) & I/R($n{=}3$) & I/R($n{=}5$) \\
\midrule
\multirow{4}{*}{Qwen3-8B}
 & Interchange (clustered) & +2.5 & +10.3 & +45.4 & 0.22 & 0.21 & 0.39 \\
 & Replacement            & +11.3 & +48.2 & +116.5 & --- & --- & --- \\
 & SLEB-iterative         & +2.5 & +11.2 & +46.2 & --- & --- & --- \\
 & BI-guided              & +2.5 & +21.6 & +137.7 & --- & --- & --- \\
\midrule
\multirow{4}{*}{Pythia-1.4B}
 & Interchange            & +11.0 & +61.8 & +377.8 & 0.66 & 0.36 & 0.16 \\
 & Replacement            & +16.7 & +172.0 & +2310.6 & --- & --- & --- \\
 & SLEB-iterative         & +11.6 & +61.9 & +658.7 & --- & --- & --- \\
 & BI-guided              & +11.0 & +74.7 & +186.9 & --- & --- & --- \\
\bottomrule
\end{tabular}}
\captionof{table}{Harmonized cross-model re-execution (WikiText-2 test, 5K words, 512/256, TPU v6e-8). $\Delta$PPL (\%) vs.\ baselines Qwen 12.10 / Pythia 15.87 (contract-locked instance; Table~\ref{tab:skip_qwen} uses Qwen baseline 12.09 on the same protocol but a different run, typically within ${\sim}0.03$ abs.\ PPL). I/R $=$ interchange $\Delta$\% divided by replacement $\Delta$\%.}
\label{tab:harmonized_xmodel}
\end{tabhere}

Under the harmonized contract, replacement-guided removal is consistently the worst output-grounded protocol across both \emph{RoPE} checkpoints (Qwen3-8B and Pythia-1.4B, same rotary embedding family). Measured as replacement $\Delta$PPL divided by interchange $\Delta$PPL at the same budget, interchange incurs about $4.5\times$, $4.7\times$, and $2.6\times$ less degradation than replacement on Qwen3-8B at $n{=}1$, $3$, and $5$ respectively ($11.3/2.5$, $48.2/10.3$, $116.5/45.4$); on Pythia-1.4B the same ratios are about $1.5\times$, $2.8\times$, and $6.1\times$ ($16.7/11.0$, $172.0/61.8$, $2310.6/377.8$), with the $n{=}5$ gap dominated by catastrophic replacement collapse. Interchange ties or beats SLEB-iterative at the lowest budget on both models while remaining within the same order of magnitude at higher budgets. The harmonized run uses a stricter evaluator than the earlier Pythia grid, so absolute $\Delta$PPL increases; the relative ordering (interchange better than replacement at every $n$) is unchanged (for example, $+377.8\%$ vs.\ $+2310.6\%$ at $n{=}5$). Logs for this slice are in the repository at \url{https://github.com/Gpgabriel25/ProtocolGapDiagnostic} (\S\ref{sec:reproducibility}).

\paragraph{Optional search layer.}
Interchange-seeded beam search over joint removal sets is documented in Appendix~\ref{sec:app_matched_budget} together with matched-budget sweeps against calibration-free SLEB; it is an engineering extension of interchange scoring rather than a core contribution.

\paragraph{Qwen3-8B skip-layer grid.}
Table~\ref{tab:skip_qwen} reports WikiText-2 perplexity after removing $n$ layers from Qwen3-8B (36 layers, baseline PPL 12.09 on this evaluator instance) under the matched contract used throughout \S\ref{sec:exp_skip}.

\begin{tabhere}
\small
\resizebox{\textwidth}{!}{%
\begin{tabular}{clccc}
\toprule
$n$ & Method & Layers Removed & PPL & $\Delta$ PPL (\%) \\
\midrule
\multirow{6}{*}{1} & Interchange / BI / SLEB (same layer) & \{17\} & 12.42 & \textbf{+2.5} \\
& Random control & \{10\} & 12.91 & +6.6 \\
& CKA & \{7\} & 13.28 & +9.6 \\
& Replacement-guided & \{32\} & 13.47 & +11.2 \\
& Taylor importance$^{\star}$ & \{1\} & 14.88 & +23.0 \\
& Non-swap-similar & \{6\} & 21.07 & +73.9 \\
\midrule
\multirow{7}{*}{2} & SLEB (greedy) & \{15, 20\} & 12.55 & \textbf{+3.7} \\
& Interchange-guided & \{17, 21\} & 12.88 & +6.3 \\
& SLEB (iterative) & \{17, 18\} & 13.00 & +7.3 \\
& BI-guided & \{7, 17\} & 13.59 & +12.2 \\
& CKA & \{7, 9\} & 14.47 & +19.5 \\
& Replacement-guided & \{31, 32\} & 15.96 & +31.9 \\
& Taylor importance$^{\star}$ & \{1, 2\} & 63.20 & +422.8 \\
\midrule
\multirow{8}{*}{3} & Interchange-clustered$^\ddag$ & \{15, 17, 20\} & 13.35 & \textbf{+10.2} \\
& SLEB (iterative) & \{17, 18, 19\} & 13.47 & +11.2 \\
& Interchange-distributed & \{17, 21, 26\} & 13.97 & +15.3 \\
& Random control & \{10, 20, 25\} & 14.20 & +17.2 \\
& BI-guided & \{7, 11, 17\} & 14.58 & +20.3 \\
& CKA & \{7, 9, 17\} & 15.06 & +24.3 \\
& Replacement-guided & \{28, 31, 32\} & 17.94 & +48.1 \\
& Taylor importance$^{\star}$ & \{1, 2, 3\} & 25.60 & +111.8 \\
\midrule
\multirow{7}{*}{5} & Interchange-clustered$^\ddag$ & \{15, 17, 18, 19, 20\} & 17.61 & \textbf{+45.4} \\
& SLEB (iterative) & \{17, 18, 19, 20, 21\} & 17.71 & +46.2 \\
& Interchange-distributed & \{17, 21, 26, 28, 30\} & 18.79 & +55.1 \\
& BI-distributed$^\dag$ & \{7, 12, 17, 22, 27\} & 19.45 & +60.6 \\
& Replacement-guided & \{25, 28, 30, 31, 32\} & 26.23 & +116.5 \\
& Taylor importance$^{\star}$ & \{1, 2, 3, 35, 29\} & 77.66 & +542.3 \\
& BI-guided & \{7, 8, 11, 15, 17\} & 43.59 & +259.8 \\
\bottomrule
\end{tabular}}%
\captionof{table}{Qwen3-8B skip-layer grid (WikiText-2 test, 5K words, window 512, stride 256; baseline PPL 12.09 on this evaluator instance). Bold $=$ best $\Delta$PPL per $n$. Harmonized re-run Table~\ref{tab:harmonized_xmodel} uses baseline 12.10 (same contract, separate instance; within-table $\Delta$PPL unchanged).}
\label{tab:skip_qwen}
\end{tabhere}
\noindent\small \textit{Table notes.} $^\dag$BI-distributed: minimum gap 5 between removals (Appendix~\ref{sec:app_geometry}). $^\ddag$Interchange-clustered: removals inside swap-similar region layers 15--21 (Appendix~\ref{sec:app_geometry}). $^{\star}$Taylor: unconstrained greedy one-shot scores. PPL is mean over 22 sliding windows on the 5K-word test split. Bootstrap 95\% CIs on absolute PPL (1{,}000 window resamples) do not overlap between interchange and replacement at $n{=}3$ and $n{=}5$; interchange and SLEB CIs overlap at those budgets.

Several findings are summarized in Table~\ref{tab:skip_qwen}: at $n{=}1$, interchange, BI, and SLEB agree on layer~17 while CKA and replacement pick worse targets. Interchange-clustered removals track SLEB-iterative within about one percentage point at $n{=}3/5$, whereas replacement-guided removal pays roughly $4.7\times$ more $\Delta$PPL at $n{=}3$. BI fails when it follows representation-change proxies instead of output-grounded redundancy. Appendix~\ref{sec:app_geometry} contrasts distributed vs.\ clustered interchange geometry; Appendix~\ref{sec:app_discussion} documents the one-shot Taylor failure mode under joint removal. A GPT-2-Medium geometry contrast (where spacing dominates at $n{=}5$) appears in Appendix~\ref{sec:app_geometry}.

The key insight is that output-sensitive methods (interchange scoring, SLEB) independently converge on the same set of truly redundant layers. This is a \emph{mutual validation} of both metrics, since the two reach the same region by fundamentally different paths, while representation-change proxies (BI, CKA) systematically misidentify which layers are safe to remove. The decisive differentiator is operational regime, not qualitative quality: interchange scoring requires no calibration data, no gradient computation, and no iterative recomputation (it operates purely as a zero-shot forward-pass procedure), while SLEB-iterative uses a calibration set with $O(n)$ iterative passes.

\begin{fighere}
\includegraphics[width=0.98\linewidth]{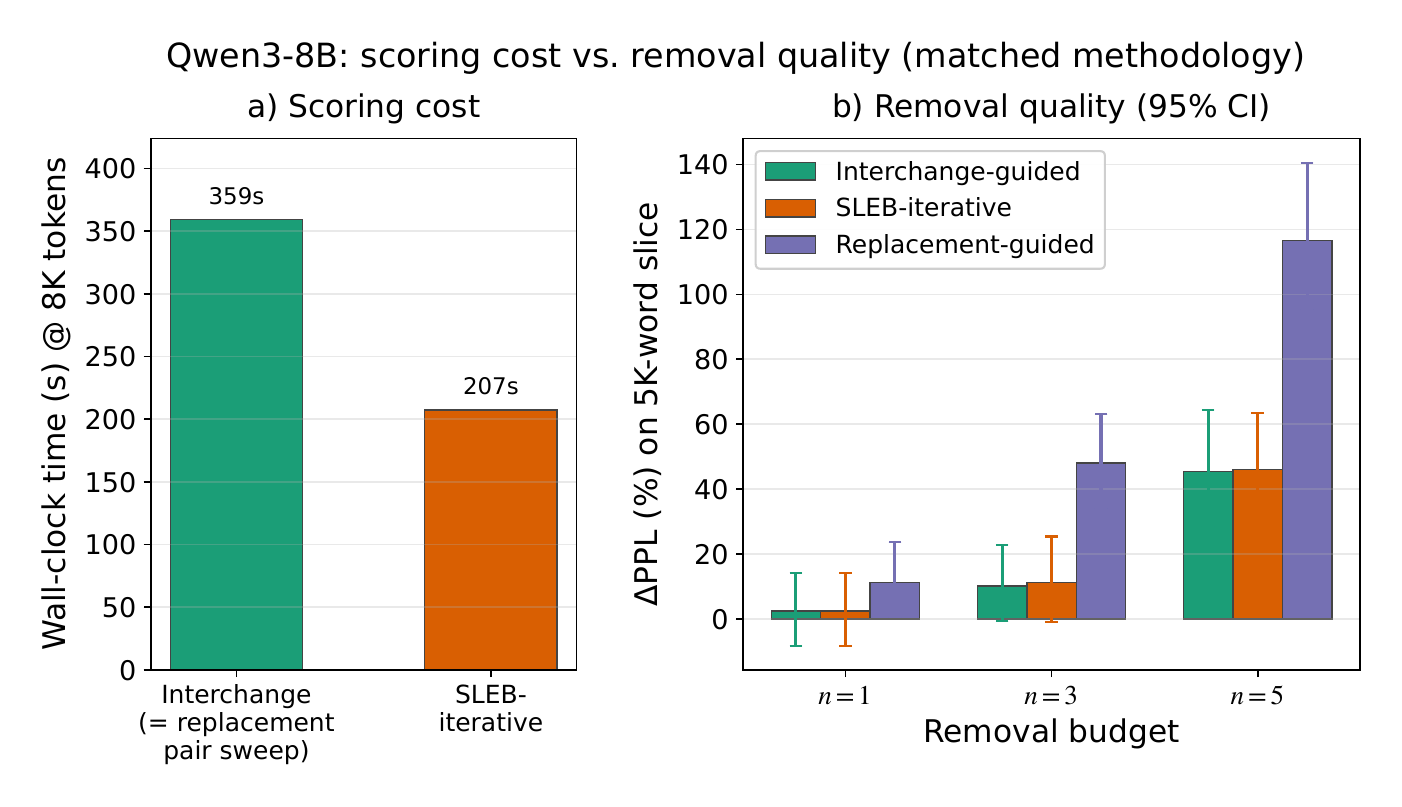}
\captionof{figure}{Qwen3-8B (a) wall-clock scoring time at 8K tokens (interchange uses the same pairwise sweep as replacement). (b) $\Delta$PPL (\%) on the 5K-word matched slice for $n{\in}\{1,3,5\}$; error bars are 95\% paired-bootstrap intervals on absolute slice PPL (2{,}000 resamples over scored windows), expressed as $\Delta$PPL (\%) via $(\mathrm{PPL}/\mathrm{PPL}_0 - 1)\times 100$, so asymmetric bars at $n{=}1$ reflect mapping from a fixed baseline. Wall-clock: interchange pairwise scoring takes ${\sim}1.73\times$ longer than SLEB-iterative at 8K tokens and ${\sim}1.74\times$ longer at 64K (equivalently, SLEB completes in ${\sim}0.58\times$ and ${\sim}0.57\times$ the interchange time).}
\label{fig:qwen_compute_quality_frontier}
\end{fighere}

Figure~\ref{fig:qwen_compute_quality_frontier} summarizes audited wall-clock scoring cost at 8K tokens next to $\Delta$PPL on the matched 5K-word slice for $n{\in}\{1,3,5\}$: interchange avoids calibration data even when SLEB is faster to score.

Tables~\ref{tab:qwen_deployment} and~\ref{tab:qwen_deployment_latency} (Appendix~\ref{sec:app_roadmaps}) make the practical tradeoff explicit, contrasting mask-based and recompiled latency measurements. The mask-based column reveals a systems limitation: XLA compiles the \texttt{jax.lax.scan} loop body as a single fused kernel, so the mask only selects between the layer computation and identity; it does not eliminate iterations, yielding a constant ${\sim}$4\% speedup regardless of how many layers are skipped. By contrast, architectural pruning (recompiling the model with physically fewer layers) delivers proportional speedup: at $n{=}5$, the recompiled model achieves 12.6\% latency reduction and 14.4\% throughput increase (89--95\% of the ideal $n/36$ bound). At $n{=}1$, the 2.6\% speedup tracks the expected 2.8\% closely (93\% efficiency). These results confirm that the interchange-guided compression translates to real deployment gains when implemented as architectural pruning, consistent with standard structured pruning practice~\citep{men2024shortgpt}.

\paragraph{From interchange distance to compression.}
Table~\ref{tab:skip_qwen} uses interchange distance to select removal targets for Qwen3-8B because replacement distance finds near-zero redundancy in RoPE models (\S\ref{sec:exp_scaling}). The bridge from interchange to compression rests on the following argument: if layers~$i$ and~$j$ have low interchange distance, their residual contributions are approximately \emph{order-invariant}, meaning that applying $g_i$ then $g_j$ gives nearly the same result as $g_j$ then $g_i$. This commutativity implies that the cross-interaction term $g_j(\mathbf{x} + g_i(\mathbf{x})) - g_j(\mathbf{x})$ is small, meaning each layer's residual contribution is approximately \emph{independent} of the other's output. In this regime, the combined output decomposes approximately as the linear superposition $F(\mathbf{x}) \approx \mathbf{x} + g_i(\mathbf{x}) + g_j(\mathbf{x})$. Removing the weaker contributor (say~$g_j$, with $\|g_j\| < \|g_i\|$) eliminates a term that adds little marginal information beyond the dominant transformation~$g_i$, with bounded impact on downstream predictions. Table~\ref{tab:selection_algorithm} (Appendix~\ref{sec:app_roadmaps}) states the greedy score-to-removal procedure used throughout the paper to convert pair scores into a concrete layer-removal set.

We note that this is a \emph{sufficient condition argument}, not a tight bound: high interchange distance does not guarantee that removal will be costly (both layers could individually be weak contributors that happen not to commute). We do not claim a formal guarantee from interchange distance to removal safety; the bridge is \emph{empirically validated} rather than theoretically proven. The empirical validation is provided by the compression results in Table~\ref{tab:skip_qwen}: interchange-guided removal achieves +6.3\% degradation at $n{=}2$ while methods that ignore interchange structure (BI, CKA) degrade 2--3$\times$ more. On GPT-2-Medium, the compression sweep (Appendix~\ref{sec:app_compression_sweep}) provides additional evidence: interchange-guided removal selects the $n$ lowest min-neighbor-distance layers, yielding up to 47\% less degradation than random selection at $n{=}5$ and avoiding the catastrophic failure of anti-guided removal (PPL\,$>$\,9000).

\paragraph{Downstream task preservation.}
To verify that interchange-guided removal preserves task performance beyond perplexity, we present evidence at two scales. On GPT-2-Medium, we evaluate zero-shot accuracy before and after removing a swap-similar layer (layer~5, KL\,$=$\,0.036) vs.\ a non-swap-similar layer (layer~0) on six benchmarks (Appendix~\ref{sec:app_downstream}): interchange-guided removal retains 96.6\% of baseline accuracy (mean $\Delta{=}-0.016$ across tasks), while non-swap-similar removal retains only 62.5\% (mean $\Delta{=}-0.180$), an 11$\times$ larger degradation. On Qwen3-8B, we evaluate interchange-guided removal on LAMBADA, HellaSwag, ARC-Easy, and WinoGrande (Appendix~\ref{sec:app_downstream}, Table~\ref{tab:downstream_qwen}): at $n{=}1$, mean downstream accuracy retains 91.7\% of the baseline, with HellaSwag retaining 95.7\% and ARC-Easy retaining 96.0\%. LAMBADA, which requires exact last-word prediction and is most sensitive to distributional shifts, retains only 74.9\%, suggesting that precision-critical tasks degrade faster than multiple-choice benchmarks. Even at $n{=}5$ (14\% compression), mean retention is 80.7\%. These results confirm that the perplexity signal translates to downstream task preservation at both GPT-2 and 8B scale.

\paragraph{Predictor-validity bridge (score to removal cost).}
To directly test whether swap-distance scores track realized pruning harm, we run a pair-to-removal validation on GPT-2-Medium: for each of the 276 layer pairs, we map the pair score to the lower of the two corresponding single-layer removal costs, then compute rank correlation between score and realized degradation. Using the same replacement-distance metric as Definition~\ref{def:swap_kl}, we obtain Spearman $\rho{=}-0.229$ ($p{=}1.26\times 10^{-4}$), indicating that lower distance ranks are significantly associated with lower removal cost. This is a modest but statistically reliable predictor signal: useful for candidate prioritization, but not a complete proxy for final multi-layer optimization. The low correlation is consistent with GPT-2-Medium's absolute PE architecture, where many pairs are already conditionally swap-similar (98 of 276 below KL\,$=$\,0.10, including 18 below $0.05$), compressing the range of distances and attenuating rank correlation. We then run RoPE-side validations on two modern families, where higher distance variance enables stronger prediction. On Pythia-1.4B (23 adjacent pairs, 20-sample removal evaluation), pair-level score-to-removal correlations are significant (interchange $\rho{=}0.477$, $p{=}0.021$; replacement $\rho{=}0.636$, $p{=}1.09\times10^{-3}$), and layer-level swap-derived scores outperform BI/CKA baselines (interchange-minpair $\rho{=}0.717$, replacement-minpair $\rho{=}0.675$, BI $\rho{=}0.610$, CKA $\rho{=}0.527$). On Qwen3-8B (Kaggle T4x2 replication, 24 pair prompts, 10-sample removal evaluation), we again find strong pair-level predictivity (interchange $\rho{=}0.624$, $p{=}8.07\times10^{-5}$; replacement $\rho{=}0.784$, $p{=}6.75\times10^{-8}$) and stronger layer-level correlations for swap-derived predictors (interchange-minpair $\rho{=}0.568$, replacement-minpair $\rho{=}0.775$) than BI/CKA (BI $\rho{=}0.345$, CKA $\rho{=}0.347$). This cross-model consistency supports the practical claim that swap-based scores carry stronger pruning-relevant signal than generic representation-similarity proxies in modern RoPE models.
\subsection{Comparison with Existing Methods}
\label{sec:exp_laco_sleb}

\begin{tcolorbox}[colback=gray!8, colframe=gray!40, boxrule=0.5pt, left=4pt, right=4pt, top=3pt, bottom=3pt]
\emph{Protocol selection.} When replacement distances are high (Qwen3-8B-like, most pairs $>0.10$), replacement-guided removal is unreliable; use interchange-guided removal or SLEB-iterative. When replacement distances are low (Llama-3.1-8B-like, most pairs $<0.10$), both protocols identify the same redundant layers and method quality (SLEB-iterative vs.\ BI) dominates protocol choice. In calibration-free settings, interchange scoring is the default; Appendix~\ref{sec:app_matched_budget} compares interchange-seeded beam search to SLEB under matched evaluator-call budgets.
\end{tcolorbox}

We evaluate interchange-guided layer removal against four established baselines: ShortGPT's Block Influence (BI) score~\citep{men2024shortgpt}, SLEB~\citep{song2024sleb} in two variants (single-pass greedy selection, ``SLEB-greedy''; and iterative selection with recalibration after each removal, ``SLEB-iterative''), and first-order Taylor importance pruning, which ranks layers by the calibration-averaged magnitude of $\nabla_\theta \mathcal{L} \odot \theta$. The SLEB and Taylor baselines both use a small WikiText-2 calibration split. We also include LaCo-style \emph{layer collapse}~\citep{yang2024laco} (``LaCo''), which progressively merges adjacent layer pairs by weight averaging, ordered by swap-KL distance (lowest first). LaCo serves as a diagnostic for whether swap-similar layers are \emph{parametrically} equivalent; they are not (see Appendix~\ref{sec:app_negative}). A detailed head-to-head comparison with BI at each compression budget is in Appendix~\ref{sec:app_bi_comparison}, and an ablation disentangling metric quality from selection geometry is in Appendix~\ref{sec:app_geometry}. Table~\ref{tab:extended_baselines} (Appendix~\ref{sec:app_roadmaps}) reports results under the standardized evaluator.

Under the standardized evaluator, interchange-guided removal and SLEB with iterative recalibration both dominate BI, Taylor, and SLEB-greedy at practical budgets. At $n{=}1$, all output-sensitive methods remain close: interchange scoring (+5.1\%, layer~5) and SLEB-iterative (+5.3\%, layer~3) are essentially tied, while Taylor is slightly worse (+6.3\%). From $n{=}2$ onward, interchange-guided removal and SLEB-iterative remain closely matched (+11.8\% vs.\ +11.7\% at $n{=}2$; +20.4\% vs.\ +18.9\% at $n{=}3$), Taylor sits between BI and the top output-sensitive methods (+19.8\% at $n{=}2$, +40.0\% at $n{=}3$), and SLEB-greedy degrades much faster.

The divergence between SLEB-iterative and SLEB-greedy is dramatic and instructive. SLEB-greedy identifies layers~2 and~3 as individually least-important, but removing both simultaneously collapses the embedding pathway (+63.2\% at $n{=}2$, +752.6\% at $n{=}5$). SLEB-iterative avoids this trap through recalibration: after selecting layer~3, it detects that removing another early layer would be destructive and instead selects layer~14 (mid-network), then layer~21 (late-middle). The resulting set \{3, 14, 21\} is strikingly similar to interchange scoring's distributed selection \{5, 15, 16\}: both methods independently discover that removals should be spread across the network. This confirms the output-sensitive convergence pattern observed at 8B scale (Table~\ref{tab:skip_qwen}).

By $n{=}5$, SLEB-greedy incurs +752.6\% degradation and BI incurs +247.5\%, compared with +48.7\% for interchange scoring and +43.0\% for SLEB-iterative. LaCo-style layer collapse is not just worse but catastrophic even when guided by the most similar pairs: merging layers~4 and~5 already increases PPL by +677.1\%, and larger merge budgets explode further. This is notable because LaCo~\citep{yang2024laco} reports strong results on LLaMA at 25--30\% compression; the GPT-2-Medium result suggests that layer collapse is more effective on models with more redundant (deeper, larger) architectures. This strongly supports \emph{removal} over \emph{merging} as the compression strategy for models at this scale: swap-similar layers compute approximately the same function through different parameterizations, and averaging their weights destroys both. See Appendix~\ref{sec:app_head} for head-level swap-KL analysis, and Appendix~\ref{sec:app_negative} for the weight averaging negative result.

\subsection{Training trajectory and scale}
\label{sec:exp_scaling}

To investigate how swap-KL strength scales with model size, we apply the protocol to multiple models within the GPT-2 and Pythia families, as well as OPT-350M and Qwen3-8B. Table~\ref{tab:scaling} summarizes the results. See Appendix~\ref{sec:app_pythia} for a detailed cross-architecture comparison with Pythia-1.4B.

\begin{tabhere}
\noindent\textit{(a) Broad public-model sweep.}
\resizebox{\textwidth}{!}{%
\begin{tabular}{llccccc}
\toprule
Model & Arch & Params & Layers & Best Pair & Best KL & STRONG/COND \\
\midrule
GPT-2-Small & Abs PE & 124M & 12 & 3$\leftrightarrow$4 & 0.099 & 0 / 1 \\
GPT-2-Medium & Abs PE & 355M & 24 & 4$\leftrightarrow$5 & 0.036 & 18 / 98$^\dag$ \\
GPT-2-Large & Abs PE & 774M & 36 & 4$\leftrightarrow$5 & 0.017 & 26 / 34 \\
GPT-2-XL & Abs PE & 1558M & 48 & 2$\leftrightarrow$3 & 0.020 & 10 / 42 \\
OPT-350M & Abs PE & 331M & 24 & 8$\leftrightarrow$9 & 0.036 & 10 / 19 \\
BLOOM-560M & ALiBi & 560M & 24 & 8$\leftrightarrow$9 & 0.017 & 9 / 19$^{\#}$ \\
BLOOM-1.1B & ALiBi & 1.1B & 24 & 8$\leftrightarrow$9 & 0.019 & 10 / 21$^{\#\#}$ \\
\midrule
Pythia-160M & RoPE & 162M & 12 & 1$\leftrightarrow$2 & 0.778 & 0 / 0 \\
Pythia-410M & RoPE & 405M & 24 & 20$\leftrightarrow$21 & 0.137 & 0 / 0 \\
Pythia-1.4B & RoPE & 1.4B & 24 & 11$\leftrightarrow$13 & 0.096 & 0 / 3$^\ddag$ \\
Pythia-1.4B (interchange) & RoPE & 1.4B & 24 & 19$\leftrightarrow$20 & 0.023 & 14 / 19$^\ddag$ \\
Pythia-2.8B & RoPE & 2.9B & 32 & 14$\leftrightarrow$15 & 0.055 & 0 / 4$^\S$ \\
\bottomrule
\end{tabular}}%
\vspace{4pt}
\small{$^\dag$All 276 pairs tested (gap$\leq$23). $^{\#}$All 23 adjacent pairs; best KL = 0.017 (2$\times$ better than architecture-matched GPT-2-Medium). $^{\#\#}$All 23 adjacent pairs; same best pair as BLOOM-560M, confirming ALiBi pattern at a second scale. $^\ddag$66 pairs tested (gap$\leq$3). $^\S$31 adjacent pairs tested (gap$=$1).}

\vspace{0.75em}
\noindent\textit{(b) Replacement vs.\ interchange (8B RoPE; pair geometry as in footnotes).}
\resizebox{\textwidth}{!}{%
\begin{tabular}{llccccc}
\toprule
Model & Arch & Params & Layers & Best Pair & Best KL & STRONG/COND \\
\midrule
\multicolumn{7}{l}{\textit{Llama-3.1-8B (100 prompts $\times$ 64 tokens, TPU v6e, bf16):}} \\
Llama-3.1-8B (replacement) & RoPE+GQA & 8.0B & 32 & 11$\leftrightarrow$12 & 0.005 & 27 / 27$^\diamondsuit$ \\
Llama-3.1-8B (interchange) & RoPE+GQA & 8.0B & 32 & 25$\leftrightarrow$26 & 0.001 & 31 / 31$^\heartsuit$ \\
\midrule
\multicolumn{7}{l}{\textit{Qwen3-8B (500 prompts $\times$ 128 tokens, TPU v6e, bf16):}} \\
Qwen3-8B (replacement) & RoPE+GQA & 8.2B & 36 & 15$\leftrightarrow$17 & 0.171 & 0 / 0$^\P$ \\
Qwen3-8B (interchange) & RoPE+GQA & 8.2B & 36 & 25$\leftrightarrow$26 & 0.056 & 0 / 18$^\|$ \\
\bottomrule
\end{tabular}}%
\vspace{4pt}
\small{$^\P$\emph{Qwen3-8B, replacement $d_{\text{repl}}$} (Definition~\ref{def:swap_kl}, bidirectional max): 102 pairs with $|i{-}j|{\leq}3$, 500 prompts$\times$128 tokens (bf16, TPU v6e). No pair falls below the 0.10 threshold (best: 15$\leftrightarrow$17, KL$=$0.171). $^\|$\emph{Interchange} (Remark~\ref{rem:interchange}): same 102 pairs. 18 conditional ($<$0.10) interchange pairs concentrated in layers 15--30. The 4--8$\times$ gap between replacement and interchange distances on this checkpoint shows that weight duplication can be far harder than positional reordering even when both protocols use the same prompts.
$^\diamondsuit$\emph{Llama-3.1-8B, replacement $d_{\text{repl}}$} (Definition~\ref{def:swap_kl}): 31 adjacent pairs (gap$=$1 only); best pair 11$\leftrightarrow$12 (KL$=$0.005). 27 pairs below 0.05 on replacement KL, consistent with a low-replacement-KL core plus a few large boundary-adjacent pairs.
$^\heartsuit$\emph{Llama-3.1-8B, interchange}: same 31 adjacent pairs. All 31 below 0.10 (26 below 0.05). Mean $\mathrm{KL}_{\mathrm{inter}}/\mathrm{KL}_{\mathrm{repl}}=0.30$ over these pairs (not mean replacement KL); across the 13-model sweep, protocol-gap presence is architecture-dependent (largest in Qwen3-8B, negligible in Llama/Mistral/APE controls).}
\captionof{table}{Swap-KL sweep across scales (split for readability). (a) Public-model sweep: STRONG/COND count adjacent pairs (gap$=$1) with mean KL below 0.05 / 0.10 unless footnoted. (b) 8B RoPE: same $d_{\text{repl}}$ and $d_{\text{interchange}}$ definitions; pair sets differ (Table~\ref{tab:setup_matrix}). Rows are not cross-comparable as a PE leaderboard (Appendix~\ref{sec:app_setup}).}
\label{tab:scaling}
\end{tabhere}

\noindent\textit{Table~\ref{tab:scaling} notes.} Panel (a): ``best KL'' is the lowest mean swap-KL among pairs in the counted set; STRONG/COND are cumulative counts. Panel (b): Qwen rows pool 102 pairs ($|i{-}j|{\leq}3$); Llama rows use 31 adjacent pairs only. Replacement is always $d_{\text{repl}}$ per Definition~\ref{def:swap_kl} (bidirectional max). Heterogeneous prompt budgets and hardware are documented in Appendix~\ref{sec:app_setup}; read within-row protocol comparisons as primary, not cross-row ordering.

Several clear patterns emerge from Table~\ref{tab:scaling} (with the depth and parameter-count trends visualized in Figure~\ref{fig:scaling}). First, swap-KL similarity generally strengthens with depth: GPT-2-Large (36 layers, best KL\,$=$\,0.017) has swap-KL distance half that of GPT-2-Medium (24 layers, 0.036), which in turn shows far lower swap-KL than GPT-2-Small (12 layers, 0.099). GPT-2-Large has 26 strongly swap-similar adjacent pairs out of 35: virtually every pair beyond layer~0. The same trend holds for Pythia (best KL: 0.778$\to$0.137$\to$0.096$\to$0.055 from 160M to 2.8B). The trend plateaus at GPT-2-XL (48 layers, only 10 strong pairs) due to a specialization spike at layer~13 (KL\,$=$\,0.832), which creates two distinct swap-similar clusters separated by a non-swap-similar boundary.

Across public checkpoints, replacement swap-KL strengths vary with depth and training recipe in ways that loosely track PE \emph{labels} (e.g., GPT-2/OPT vs.\ Pythia at matched depth; BLOOM vs.\ GPT-2 in Figure~\ref{fig:bloom_comparison}), but those labels are descriptive tags, not identified mechanisms: width, data, and optimization co-vary with them. Treat cross-row ``best KL'' ordering as exploratory context only.

Fourth, Qwen3-8B reveals a dramatic protocol sensitivity. Under a strengthened evaluation (500 prompts$\times$128 tokens, 102 pairs with $|i{-}j|{\leq}3$, bf16 on TPU v6e), mutual interchange (Remark~\ref{rem:interchange}) finds 18 conditionally swap-similar pairs (best KL\,$=$\,0.056, pair 25$\leftrightarrow$26); under replacement $d_{\text{repl}}$ (Definition~\ref{def:swap_kl}, bidirectional max on the same 102 pairs), zero pairs fall below the conditional threshold (best KL\,$=$\,0.171, pair 15$\leftrightarrow$17). The per-pair protocol gap averages 4.8$\times$ across mid-network gap$=$1 pairs (e.g., pair 25$\leftrightarrow$26: interchange 0.056 vs.\ replacement 0.334, a 6.0$\times$ gap). Pythia-1.4B confirms the effect: interchange yields 14 strong swap-similar pairs (best KL\,$=$\,0.023) while replacement yields zero, a mean protocol gap of 4.5$\times$ across non-boundary pairs. For GPT-2 with absolute PE, both protocols agree ($\approx$0.036), the protocol gap is largest in Qwen3-8B among the models tested, while 2/3 RoPE 8B models (Llama, Mistral) show I/R$\approx$1, a pattern the controlled ablations (below) identify as driven by layer distance and training duration rather than PE type per se.
Llama-3.1-8B confirms the protocol gap at a second RoPE+GQA checkpoint: all 31 adjacent pairs have interchange KL $<$ 0.10 (best: 0.001), interchange KL is below replacement KL on every pair (best replacement KL on a gap-1 pair is 0.005 for 11$\leftrightarrow$12), and the per-pair ratio $\mathrm{KL}_{\mathrm{inter}}/\mathrm{KL}_{\mathrm{repl}}$ averages 0.30. Notably, Llama-3.1-8B's \emph{absolute} replacement distances are substantially lower than Qwen3-8B's (best KL 0.005 vs.\ 0.171), despite both being RoPE+GQA models of similar scale. We do \textit{not} attribute this cross-model difference to training-token counts: Qwen3-8B is trained on ${\sim}$36T tokens while Llama-3.1-8B on ${\sim}$15T, yet Qwen exhibits \emph{higher} replacement distances, the opposite of a ``more training produces more redundancy'' story. The cross-model difference more likely reflects architectural factors (Llama GQA 4:1 vs.\ Qwen3 GQA 8:1, different width/depth ratio) rather than training duration per se; isolating these contributors is a direction for follow-up work. The key observation is that this does \emph{not} eliminate the protocol gap (both models exhibit interchange $\ll$ replacement), confirming that the gap is a per-pair specialization property that persists across training regimes, not an artifact of underfitting. Qwen3-8B's higher replacement distances make the gap more visible as an absolute magnitude; Llama-3.1-8B shows the same qualitative structure at lower absolute distances.

\paragraph{Positional encoding is associated with the protocol gap but does not solely cause it.}
To test whether PE type alone is sufficient, we train architecturally matched AbsPE and RoPE transformers from scratch, controlling all other variables (Appendix~\ref{sec:app_pe_ablation}). Three early ablations spanning 17.6M--63.6M parameters and 1.8K--50K training steps consistently failed to reproduce the protocol gap (both PE types show mean I/R ratios near 1.0). A multi-depth trajectory study (6/12/24 layers, 44.7M--101.4M parameters, 10K steps on TPU) revealed a monotonic depth trend: the mean I/R ratio decreases from 1.37 (6L) to 1.01 (12L) to 0.60 (24L) for both PE types, but without a PE-dependent gap. To test whether undertraining was responsible (per a Chinchilla-floor critique that 17--101M parameter models with $<$100M training tokens are far below the compute-optimal regime needed to grow the gap), we executed a controlled 152M-parameter ablation (16L/768d/12H, AbsPE vs RoPE, identical optimizer/data/seed) at 2.0B tokens, 13$\times$ the Chinchilla floor for this size, on TPU v6e-8 (Appendix~\ref{sec:app_pe_ablation}). At this scale the \emph{mean} I/R ratio is again statistically indistinguishable between PE types (AbsPE\,$=$\,1.24, RoPE\,$=$\,1.26), confirming the conclusion from earlier ablations. However, when stratified by layer-pair distance the from-scratch models do reproduce the central qualitative pattern of pretrained models: at adjacent pairs (gap$=$1) the median I/R ratio falls to 0.35 (RoPE) and 0.56 (AbsPE), then climbs monotonically to 1.16 / 1.24 at gap$=$3, closely tracking real Qwen3-8B (0.49 at gap$=$1, 0.98 at gap$=$3, mean across pair sets, see Table~\ref{tab:skip_qwen}) up to a 2--3$\times$ overall amplification at full pretrained scale. Two conclusions follow. (i) The protocol gap as a \emph{layer-distance phenomenon} (adjacent pairs interchangeable, distant pairs not) is reproducible from scratch, training-duration-grown, and present at both 152M and 8B scales. (ii) PE \emph{type} is not the proximate cause: AbsPE shows the same gap=1 vs gap=3 monotonicity as RoPE, only with slightly larger absolute distances. The PE-driven framing should therefore be replaced by a layer-distance-and-training-duration framing, with PE acting as a secondary modulator of absolute distance magnitudes rather than as the primary axis of protocol divergence. This refinement strengthens rather than weakens the practical contribution: because the gap is gap-distance-driven and cannot be predicted from PE alone, practitioners \emph{must} measure it empirically via interchange testing on the actual layer pairs they intend to compress.

\paragraph{Inference-time RoPE counterfactual (Qwen3-8B).}
To directly test whether RoPE rotation is mechanistically responsible for the protocol gap, we perform an inference-time counterfactual on Qwen3-8B: for 12 adjacent pairs spanning the network, we compute replacement and interchange distances under normal RoPE and with RoPE disabled (setting all rotation angles to zero, $\cos\theta{=}1, \sin\theta{=}0$). Removing RoPE causes substantial output divergence (mean KL\,$=$\,2.50 vs.\ original), confirming a meaningful intervention. The result is consistent with causation being elsewhere: the protocol gap persists without RoPE and intensifies slightly (10/12 pairs show a larger gap; mean I/R ratio shifts from 0.304 to 0.259, $p < 0.01$ by sign test). This is inconsistent with RoPE rotation as the proximate cause and is consistent with a \emph{layer weight specialization} hypothesis: layers trained in a deep RoPE model develop position-in-network-specific functions that resist duplication (high replacement distance) while remaining approximately order-invariant with neighbors (low interchange distance), regardless of whether rotary encoding is active at inference time.

\paragraph{Training trajectory and pair-heterogeneity (Pythia-410M and Pythia-1.4B).}
To trace \emph{when} the protocol gap emerges during training, we exploit Pythia's published intermediate checkpoints~\citep{biderman2023pythia} for two widths where we have a complete five-checkpoint grid in our archived trajectory export: \texttt{pythia-410m} and \texttt{pythia-1.4b} (both 24 layers, 23 adjacent pairs), measured at step~0 (random init), then steps 1K, 16K, 64K, and 143K, using 32 diverse prompts $\times$ 128 tokens on TPU~v6e in fp32 (10 model$\times$checkpoint cells).
\textit{Trajectory export note:} Figure~\ref{fig:protocol_gap_dist} is generated from this export after dropping \texttt{pythia-2.8b}: the 2.8B rows in the JSON repeat identical aggregate statistics at every logged step, so they do not define a usable training curve in this file. Static adjacent-pair swap-KL summaries for additional Pythia widths (including 160M and 2.8B) appear in Table~\ref{tab:scaling} and are not overlaid on the trajectory panels. A separate fixed-geometry stress test (66 pairs with $|i{-}j|{\leq}3$) is reported below with a sparser checkpoint grid.

\textit{Training duration increases the mean gap net of initialization, with non-monotone transients.} Recomputing the mean per-pair gap (KL$_{\text{repl}}{-}$KL$_{\text{inter}}$) from finite adjacent pairs, Pythia-410M moves from $+0.035$ at step~0 to $+0.263$ at step~143K ($\approx$$7.5\times$). Pythia-1.4B moves from $+0.034$ at step~0 to $+0.403$ at step~143K ($\approx$$11.7\times$), but is not monotone at every logged step (notably near step~1K). Across both widths, early-training gaps remain far below their converged values, supporting a training-induced mechanism rather than a fixed architectural constant.

As a fixed-geometry robustness check, we repeated the same wider pair set ($|i-j|\leq 3$, 66 pairs rather than adjacent-only) at two scales. For Pythia-1.4B, the aggregate gap is near zero early (step~0: +0.026, step~1K: $-0.005$) but large at convergence (step~143K: +0.347), and 92.4\% of pairwise gaps increase from step~1K to step~143K. Pythia-410M shows the same pattern under the same geometry: +0.027 at step~0, +0.014 at step~1K, +0.329 at step~143K, with 81.8\% of pairwise gaps increasing from step~1K to step~143K. This supports the same training-duration mechanism beyond adjacent-pair selection and beyond a single model scale.

\textit{At matched depth (24 layers), widening from 410M to 1.4B increases heterogeneity more than the pooled mean alone suggests.} At step~143K, the mean per-pair gap rises from $+0.263$ to $+0.403$ ($\approx$$+53\%$), while the median falls from $+0.148$ to $+0.067$ and the maximum rises from $1.96$ to $2.36$ nats. Larger width does not uniformly push every adjacent pair away from interchangeability; instead, the bulk of pairs move closer to interchange while a tail remains specialist-heavy, the pattern swap-KL analysis targets. Figure~\ref{fig:protocol_gap_dist} plots mean/median/p75/max against parameter count and overlays per-pair profiles (log scale on the vertical axis) at each model's final logged checkpoint in the export (143K for both curves).

These results refine, rather than overturn, the I/R ratio narrative used in earlier work on this dataset: the practical decision rule (\S\ref{sec:exp_skip}) is unchanged, but the underlying phenomenon is heterogeneity across pairs, not a single pooled ratio.

\begin{fighere}
\includegraphics[width=\textwidth]{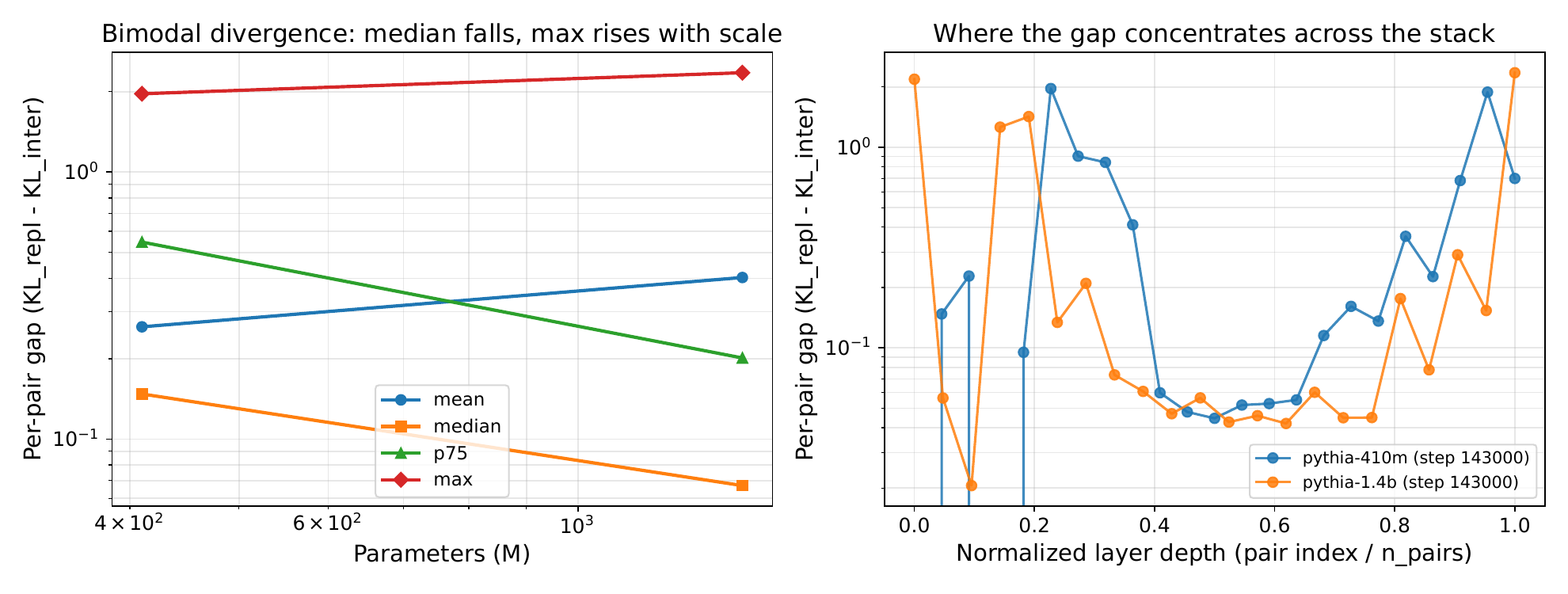}
\captionof{figure}{Per-pair protocol gap distribution at the final logged checkpoint in our Pythia trajectory export (step~143K for both 410M and 1.4B). \emph{Left:} mean, median, p75, and max of the per-pair gap as a function of parameter count at matched depth (24 layers). Mean rises with width while the median falls, indicating a tightening ``core'' of approximately interchangeable pairs alongside a specialist tail. \emph{Right:} per-pair gap versus normalized depth (log scale on the vertical axis). The boundary-adjacent pair remains the largest single gap in both models, while mid-stack pairs cluster at much smaller values.}
\label{fig:protocol_gap_dist}
\end{fighere}

\begin{fighere}
\includegraphics[width=\textwidth]{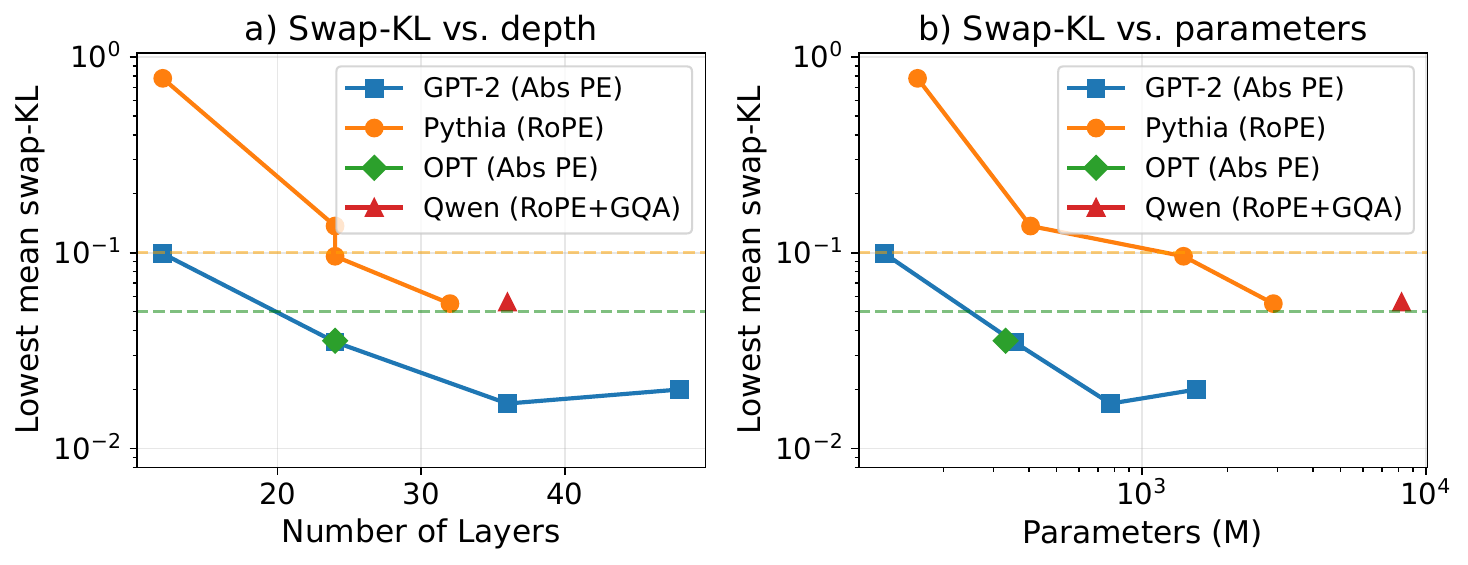}
\captionof{figure}{Lowest reported mean swap-KL in each model's evaluated pair set (pair geometry per Table~\ref{tab:scaling}; adjacent-only for most rows, $|i{-}j|{\leq}3$ for Qwen3-8B and for the default Pythia-1.4B row) vs.\ depth (a) and parameter count (b). GPT-2 (absolute PE, squares) is consistently more swap-similar than Pythia (RoPE, circles) at every matched depth. Dashed lines indicate strong ($<$\,0.05) and conditional ($<$\,0.10) thresholds.}
\label{fig:scaling}
\end{fighere}

\begin{fighere}
\includegraphics[width=\textwidth]{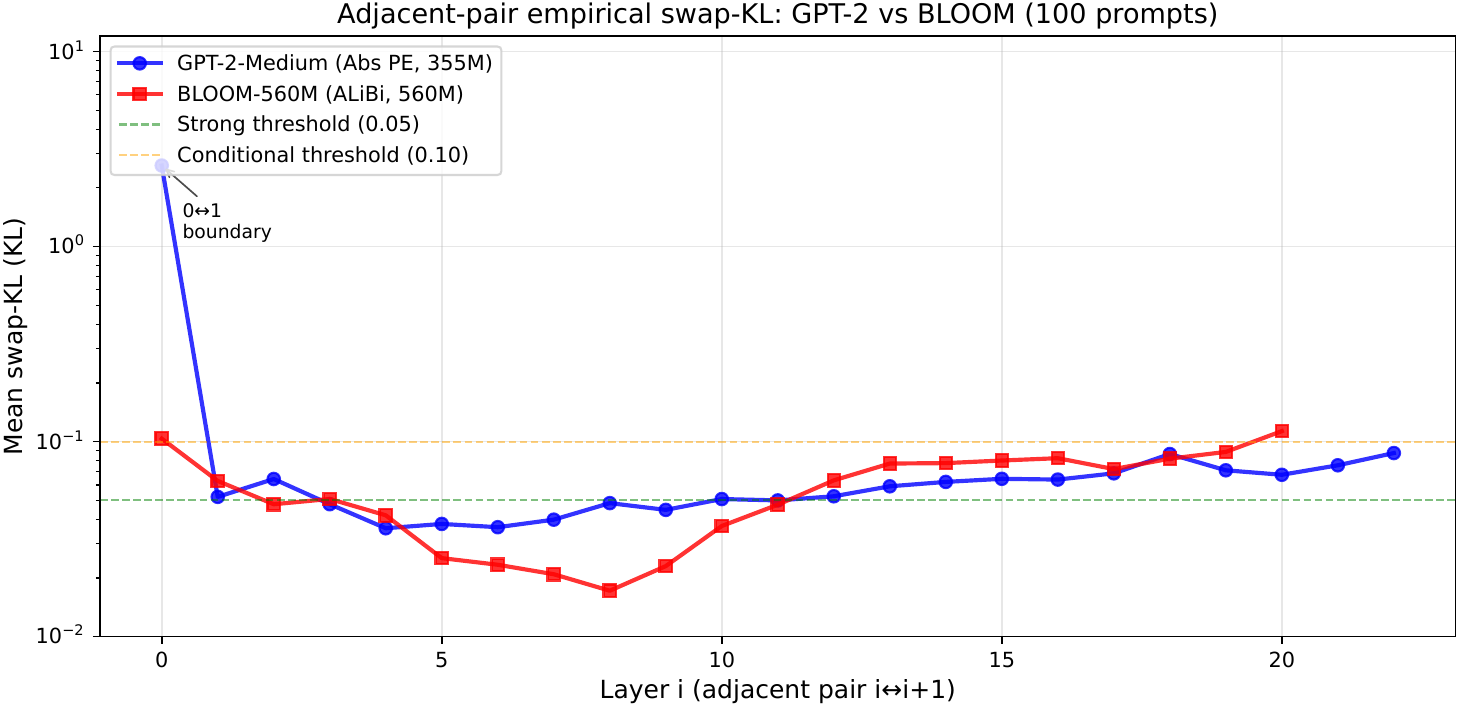}
\captionof{figure}{Adjacent-pair swap-KL distance profile for GPT-2-Medium (absolute PE, 355M) vs.\ BLOOM-560M (ALiBi, 560M). ALiBi produces a deeper swap-similar core (minimum at layers 8$\leftrightarrow$9, KL\,$=$\,0.017) with 25$\times$ less boundary specialization (0$\leftrightarrow$1). Both models have 24 layers and 1024 hidden dimension. Log scale.}
\label{fig:bloom_comparison}
\end{fighere}

\FloatBarrier

%% ============================================================
\section{Related Work}
\label{sec:related}
%% ============================================================

\paragraph{Layer Pruning.}
ShortGPT~\citep{men2024shortgpt} proposed the Block Influence (BI) score, defined as the cosine distance between a layer's input and output, to identify removable layers, pruning up to 25\% of LLaMA-2. LaCo~\citep{yang2024laco} progressively collapses adjacent layers; SLEB~\citep{song2024sleb} uses structured block elimination with calibration-based angular distance. \citet{gromov2024unreasonable} showed that up to half the layers of some open-weight LLMs can be removed with minimal degradation after QLoRA healing. More recently, EvoPress~\citep{sieberling2025evopress} formulates layer dropping as evolutionary search over per-layer compression profiles, showing that greedy selection is suboptimal when layers interact; our beam-search ablation (Appendix~\ref{sec:app_compression_sweep}) reaches a similar conclusion while holding the scoring signal fixed. We do not include EvoPress in the matched quantitative tables because it couples candidate scoring to a separate evolutionary search budget over joint layer sets, whereas our head-to-head tables isolate scoring rules under fixed selection procedures. BlockPruner~\citep{zhong2025blockpruner} decomposes layers into MHA/MLP blocks and prunes via perplexity-based importance. Sliding-Window Merging~\citep{ding2026swm} identifies ``patch-like'' redundancy via RKHS kernel correlation and \emph{merges} consecutive layers by parameter consolidation. This is the closest existing method to our merging discussion, though it uses intermediate-representation similarity rather than output-level behavioral equivalence. Our swap-KL distance differs fundamentally from all these approaches: rather than measuring single-layer importance, we estimate \emph{pairwise functional interchangeability}, enabling informed decisions about which layer to keep from a swap-similar pair. In head-to-head comparison (\S\ref{sec:exp_skip}), interchange scoring outperforms BI by 47--82\% at multi-layer removal on Qwen3-8B because it selects from the interchange-equivalent region rather than protocol-agnostic layers. Compared with SLEB-iterative, output quality converges at low compression (+10.3\% vs.\ +11.2\% at $n{=}3$ on Qwen3-8B), validating that both methods recover the same redundant-layer region. Interchange scoring achieves this zero-shot, with no calibration data, no gradient computation, and no iterative recomputation, while SLEB requires a calibration set and $O(n)$ iterative passes.

\paragraph{Causal Tracing and Activation Patching.}
Our layer-swap protocol is related to causal mediation analysis~\citep{vig2020causal}: activation patching localizes factual knowledge~\citep{meng2022locating} and automated circuit discovery identifies critical components~\citep{conmy2023automated}. These methods intervene on single layers to identify \emph{which} components contribute to specific behaviors. We differ by measuring pairwise \emph{interchangeability} aggregated over all inputs, a compression metric rather than an interpretability tool, though both share layer-swap methodology.

\paragraph{Weight Sharing and Representation Similarity.}
ALBERT~\citep{lan2020albert} shares weights across all layers; Universal Transformers~\citep{dehghani2019universal} iterate the same block. CKA~\citep{kornblith2019similarity} measures \emph{representation} similarity~\citep{raghu2021vision}, but high CKA layers may compute different functions. Our metric tests for functional equivalence at the output distribution. LayerDrop~\citep{fan2020reducing} trains models robust to layer removal; interchange scoring could identify which layers benefit most.

\paragraph{Process-algebra motivation.}
Quantitative interchangeability was formalized by \citet{park1981concurrency} and \citet{milner1989communication}; \citet{desharnais2004metrics} extended this to metrics. Our approach takes only the structural intuition (testing observational equivalence by swapping components) and replaces the formal supremum with an empirical expectation; no process-algebra guarantees transfer. Our approach is orthogonal to distillation~\citep{hinton2015distilling}, quantization~\citep{frantar2023gptq}, weight pruning~\citep{frankle2019lottery}, and structured channel pruning methods such as SliceGPT~\citep{ashkboos2024slicegpt}, and can be composed with them.

\paragraph{What we add.}
Prior layer-pruning and similarity scores typically commit to a \emph{single} protocol (importance, representation similarity, or one swap direction). To our knowledge, no prior work holds the evaluator and removal budget fixed while comparing \emph{pairwise} output-grounded \emph{replacement} vs.\ \emph{interchange} swap-KL and showing that this choice reverses which layers look redundant enough to prune. Our matched Qwen/Llama tables and the Pythia checkpoint trajectory instantiate that claim; everything else is supporting context.

%% ============================================================
\section{Discussion}
\label{sec:discussion}
%% ============================================================

\paragraph{Computational requirements.}
Table~\ref{tab:method_requirements} summarizes the practical requirements of each layer-importance scoring method evaluated in this work.

\begin{tabhere}
\small
\begin{tabular}{lccccl}
\toprule
\textbf{Method} & \textbf{Output-grounded} & \textbf{Calibration data} & \textbf{Gradients} & \textbf{Iterative} & \textbf{Cost} \\
\midrule
Swap-KL (ours) & \checkmark & \texttimes & \texttimes & \texttimes & $O(L^2 N)$\textsuperscript{$\dagger$} \\
SLEB-iterative & \checkmark & \checkmark & \texttimes & \checkmark & $O(n \cdot L \cdot N)$ \\
SLEB-greedy & \checkmark & \checkmark & \texttimes & \texttimes & $O(L \cdot N)$ \\
BI / CKA & \texttimes & \texttimes & \texttimes & \texttimes & $O(L \cdot N)$ \\
Taylor importance & \texttimes & \checkmark & \checkmark & \texttimes & $O(L \cdot N)$ \\
\bottomrule
\multicolumn{6}{l}{\footnotesize $\dagger$ Reducible to $O(L \cdot N)$ when restricted to adjacent pairs (gap$=$1).}
\end{tabular}
\captionof{table}{Practical requirements of layer-importance scoring methods. ``Output-grounded'' means the score is defined via the model's output logits, not intermediate representations. $L$ = number of layers, $N$ = prompt set size, $n$ = layers to remove.}
\label{tab:method_requirements}
\end{tabhere}

Our protocol-gap diagnostic is the only output-grounded method that requires no calibration data, no gradient computation, and no iterative recomputation; it operates purely on unlabeled forward passes. SLEB-iterative achieves comparable or better pruning quality at $n{\geq}3$ (Tables~\ref{tab:skip_qwen},~\ref{tab:skip_llama}), but uses a held-out calibration set and $O(n)$ re-evaluation passes. In a strict matched timing head-to-head on Qwen3-8B, SLEB was faster in wall-clock time (207.5s vs.\ 359.3s at 8K tokens; 1650.1s vs.\ 2872.9s at 64K tokens), so the practical advantage of interchange scoring is reduced data/engineering dependence, not unconditional runtime speed.

The protocol gap is the central finding: replacement and interchange measure different notions of layer redundancy. On some checkpoints (notably Qwen3-8B) the split shows up in both swap-KL and pruning rankings (large I/R); on others (Llama-3.1-8B, Mistral-7B-v0.1) pruning can tie even when interchange KL stays below replacement KL on adjacent pairs. Operationally: a large gap means protocol choice moves which layers look removable; when I/R$\approx$1 at matched budgets, swap-based selectors are largely interchangeable.

\paragraph{Intervention asymmetry vs.\ architecture.}
Interchange keeps both layers running (reordered) while replacement overwrites one slot, so one might expect interchange KL to be smaller for purely mechanical reasons. That story does not fit the data by itself: APE models still show I/R\,$\approx$\,1, while Qwen3-8B shows I/R\,$\ll$\,1, and the Pythia-410M/1.4B trajectory grows the per-pair gap by an order of magnitude to step~143K (Figure~\ref{fig:protocol_gap_dist}). Llama-3.1-8B and Mistral-7B-v0.1 match tied-pruning patterns at 8B under our evaluator (Table~\ref{tab:8b_core}; Mistral's replacement row is a proxy tied to interchange removals).

\paragraph{Deployment.}
The protocol-gap diagnostic is lightweight relative to full multi-layer pruning searches, but it is not ``$2L$ passes'': classifying the regime from \emph{adjacent} pairs costs $O(L \cdot N)$ forward passes up to constants per pair and protocol (replacement scores two directional surrogates per pair before symmetrization; interchange scores one mutual swap; see Table~\ref{tab:method_requirements} and Appendix~\ref{sec:app_setup}). A dense $L {\times} L$ matrix scales as $O(L^2 \cdot N)$. In our Qwen3-8B configuration, the full matched-evaluation sweep (13 configurations $\times$ 5K words) costs $\sim$40s on TPU v6e-8, while adjacent-pair diagnostic scoring costs $\sim$2s on the same hardware.
For practitioners: (1) under high replacement distances, score both swap protocols on a small prompt set and prefer interchange for removal ranking when it diverges from replacement; when both are uniformly low, either swap protocol is a reasonable zero-shot selector in our budgets, while BI/CKA remain risky without output-grounded checks. (2) With calibration data and iteration budget, SLEB-iterative is a strong alternative: on Qwen3-8B it tracks interchange, and on Llama-3.1-8B it wins at $n{\geq}3$ (+28/+57\% vs.\ ${\sim}$40/85\% for either swap protocol). (3) When the measured gap is small, cheaper proxies may suffice; when it is large, keep both swap signals in the loop.

\paragraph{Scope.}
The diagnostic matters most when you care about zero-shot removal without a calibration set. If you already run SLEB-iterative to convergence on a good calibration slice, marginal value is smaller. It is also less informative when the measured gap is tiny or when aggressive multi-layer budgets are dominated by spacing constraints rather than metric ranking.

\paragraph{Limitations.}
Several limitations apply. Our evidence for the PE--protocol-gap association is based on cross-family comparisons; model families differ not only in PE type but also in training data, optimization, and architectural details. The OPT-350M and BLOOM-560M cross-family controls partially isolate PE (both are absolute/ALiBi models with different training recipes that show the same protocol behavior as GPT-2), and eight controlled ablations spanning 17.6M--152M parameters and up to 2.0B training tokens (13$\times$ Chinchilla) reveal the protocol gap as a layer-distance-and-training-duration phenomenon: when stratified by layer-pair distance, the gap${=}1$ signal is reproducible from scratch in both AbsPE and RoPE conditions (152M RoPE gap${=}1$ median I/R\,$=$\,0.35, AbsPE 0.56); when aggregated to a single mean over all pairs, the gap is washed out by high-gap pairs and PE-type separation does not emerge. This structured set of results rules out PE type as the primary axis of protocol divergence and constrains the phenomenon to layer-distance and training-duration interactions (see Appendix~\ref{sec:app_pe_ablation}). The training trajectory further provides direct causal evidence of when the gap develops, even as its proximate cause remains open. The archived Pythia-410M and Pythia-1.4B checkpoint trajectories in \S\ref{sec:exp_scaling} (Figure~\ref{fig:protocol_gap_dist}) provide direct evidence: from step~0 to step~143K, the pooled per-pair gap grows by roughly $7.5\times$ (410M) and $11.7\times$ (1.4B), while heterogeneity across pairs widens (median vs.\ max). Static rows for additional Pythia widths in Table~\ref{tab:scaling} are \emph{not} overlaid on those trajectory panels. An inference-time RoPE counterfactual on Qwen3-8B additionally rules out rotary rotation as the proximate cause (\S\ref{sec:app_rope_counterfactual}).

Experimental setups vary across model scales (prompt budgets, hardware, numerical precision; see Appendix~\ref{sec:app_setup}), but the protocol gap is computed \emph{within} each model as a dimensionless ratio, so it is not tied to a particular device. Cross-family comparisons should be read ordinally (gap present vs.\ absent) rather than as a leaderboard of absolute KLs. The matched 8B panel includes Qwen3-8B and Llama-3.1-8B with SLEB-greedy, SLEB-iterative, BI, CKA, and random baselines on both, plus Mistral under the same evaluator without bootstrap CIs (Table~\ref{tab:8b_core}). We do not solve globally optimal joint multi-layer optimization at large scale; width-3 beam ablations are summarized in Appendix~\ref{sec:app_compression_sweep}. Computing pairwise distances requires $O(L^2 \cdot N)$ forward passes (vs.\ $O(N)$ for BI), but restricting to adjacent pairs reduces cost to $O(L \cdot N)$; in practice, the 102-pair Qwen3-8B sweep runs in ${\sim}$7 minutes on a single TPU v6e.

%% ============================================================
\section{Conclusion}
\label{sec:conclusion}
%% ============================================================

We document a protocol gap in layer equivalence testing: the outcome of measuring whether transformer layers are functionally interchangeable depends on the testing protocol. For models with absolute positional embeddings (GPT-2, OPT), replacement and interchange agree on adjacent-pair swap-KL distances. Among the RoPE checkpoints in our main tables, Qwen3-8B and the Pythia-410M/1.4B trajectories show large replacement--interchange splits in both distances and pruning rankings, while Llama-3.1-8B can tie pruning under both protocols even when interchange KL stays below replacement KL on adjacent pairs (Table~\ref{tab:8b_core}). Read ``RoPE diverges'' as a tendency in \emph{metrics}, not a universal claim about selector equivalence. The 13-model sweep is ordinal across rows; controlled ablations tie the phenomenon to layer distance and training duration more than PE type alone. When the gap is large, protocol-agnostic proxies can mis-rank removable layers; when it is small, swap-based selectors are closer substitutes.

The protocol gap has practical consequences where it is large. On Qwen3-8B under a single matched evaluator, interchange-guided removal achieves +2.5--10\% degradation at 1--3 removed layers while replacement-guided degrades +11--48\%. BI ties at $n{=}1$ by selecting the same layer, but at higher budgets BI and CKA (which are protocol-agnostic) move away from the interchange-equivalent region and degrade 2--4$\times$ more. A direct head-to-head under a single matched evaluator confirms that interchange-guided removal degrades 2.6--5.1$\times$ less than replacement-guided at shared compression points on Qwen3-8B (Table~\ref{tab:skip_qwen}), amplifying the 1.8$\times$ advantage seen on Pythia-1.4B. At $n{=}2$, SLEB-greedy is strongest (+3.7\%), while at $n{=}3/5$ interchange and SLEB-iterative remain close. On Llama-3.1-8B (Table~\ref{tab:skip_llama}), replacement distances are uniformly low (27/31 pairs below 0.05), so replacement-guided removal is competitive (both swap protocols degrade $\sim$9/40/85\% at $n{=}1/3/5$), but both output-grounded protocols drastically outperform BI (+24--765\%) and random baselines. SLEB-iterative outperforms both swap protocols at $n{\geq}3$ on Llama (+28/+57\% vs.~$\sim$40/85\%). This clarifies the practical rule: when replacement distances are high, use interchange; when both are low, either swap protocol is a safe zero-shot selector for the budgets we test, but calibrated iterative methods can still win at higher $n$; in all cases, output-grounded methods dominate representation proxies. On GPT-2-Medium, where both protocols agree, interchange-guided removal and SLEB-iterative both achieve +5--19\% degradation at 1--3 removed layers with similar distributed selections. The main contribution is a protocol-aware, output-grounded diagnostic (paired replacement and interchange swap-KL) for zero-shot pruning, with the largest \emph{pruning} leverage in high-replacement-distance regimes (Qwen3-8B-like models). Appendix~\ref{sec:app_matched_budget} records interchange-seeded beam search as an optional calibration-free search layer on top of the diagnostic. The primary differentiator is operational regime rather than guaranteed runtime superiority: swap-based scoring avoids calibration data and iterative rescoring, while matched Qwen3-8B timing shows SLEB-iterative completes scoring faster (interchange pairwise scoring takes ${\sim}1.73\times$/$1.74\times$ longer at 8K/64K tokens in our audited run). When calibration data and iterative budget are available, SLEB-iterative is a strong competitor that converges to similar selections and can outperform swap-based methods at higher compression budgets (as on Llama-3.1-8B at $n{\geq}3$).

Looking ahead, we view the cross-checkpoint spread in replacement swap-KL as motivation for controlled counterfactuals at fixed width, depth, and data, not as a positional-encoding ranking. Extending the replacement--interchange comparison to other tasks, routing-heavy architectures, and alternative positional schemes is natural follow-up work.

\section*{Acknowledgments}
We thank the Google TPU Research Cloud (TRC) for providing TPU resources that supported this research.

\section*{Reproducibility Statement}
\phantomsection\label{sec:reproducibility}
All swap-KL distance computations, layer removal experiments, and baseline comparisons use publicly available pretrained checkpoints from HuggingFace: \texttt{openai-community/gpt2} (Small/Medium/Large/XL), \texttt{EleutherAI/pythia-\{410m,1.4b,2.8b\}}, \texttt{facebook/opt-350m}, \texttt{bigscience/bloom-560m}, \texttt{bigscience/bloom-1b1}, and \texttt{Qwen/Qwen3-8B}, and \texttt{meta-llama/Llama-3.1-8B}. Random seeds are fixed at 42 for all sampling operations; random-baseline selections use seeds 0--9 for 10 trials. The full GPT-2-Medium 276-pair matrix runs in ${\sim}$20 minutes on a single CPU. Code and results are available at \url{https://github.com/Gpgabriel25/ProtocolGapDiagnostic}. Distance computation uses only standard forward passes; no custom kernels are required.

\bibliographystyle{plainnat}
\bibliography{references}

%% ============================================================
%% APPENDIX
%% ============================================================
\appendix

\input{appendix_roadmap_tables}

\section{Head-to-Head Comparison with Block Influence}
\label{sec:app_bi_comparison}

ShortGPT's Block Influence (BI) score~\citep{men2024shortgpt}, defined as $1-\cos(\mathbf{h}^{(i)}_{\text{in}}, \mathbf{h}^{(i)}_{\text{out}})$ for layer~$i$, is the most widely used layer pruning metric. We compute BI scores for all 24 GPT-2-Medium layers on the same prompt set and compare both ranking and actual compression quality. The two metrics agree on boundary-layer importance (Pearson $r{=}0.96$, $p{<}10^{-6}$) but disagree substantially on the ordering of removable middle layers (Spearman $\rho{=}0.14$, $p{=}0.53$). The critical difference is in layer \emph{selection geometry}: BI greedily selects consecutive early-middle layers (\{6,8,10,12,14\} for $n{=}5$) because they have the smallest transformation magnitude, while interchange scoring selects layers \emph{distributed} across the network (\{4,14,16,20,22\}), capturing functional interchangeability rather than magnitude.

Table~\ref{tab:bi_comparison} shows the resulting perplexity when actually removing the selected layers:

\begin{tabhere}
\begin{tabular}{cccccc}
\toprule
$n$ & BI Layers & BI $\Delta$\% & Swap-KL Layers & Swap-KL $\Delta$\% & Relative Result \\
\midrule
1 & \{6\} & +5.3 & \{4\} & +5.6 & BI 5\% better \\
2 & \{6,8\} & +19.0 & \{4,14\} & +12.7 & Swap-KL 33\% less \\
3 & \{6,8,10\} & +43.0 & \{4,14,16\} & +21.9 & Swap-KL 49\% less \\
4 & \{6,8,10,12\} & +68.8 & \{4,14,16,20\} & +32.4 & Swap-KL 53\% less \\
5 & \{6,8,10,12,14\} & +109.1 & \{4,14,16,20,22\} & +52.4 & Swap-KL 52\% less \\
\bottomrule
\end{tabular}
\captionof{table}{Head-to-head: interchange-guided vs.\ BI-guided layer removal in GPT-2-Medium under the standardized evaluator. Interchange-scored layers are selected greedily by min-neighbor distance with gap${\geq}$2, which picks layer~4 at $n{=}1$; Table~\ref{tab:extended_baselines} instead selects layer~5 (lower per-layer removal cost), yielding +5.1\%. BI is slightly better at one removed layer, but interchange scoring wins clearly from 2 layers onward. Margins at $n{\geq}2$ exceed 6 absolute percentage points; by contrast, the $n{=}1$ gap (0.3~pp) is well within typical prompt-set variance ($\pm$0.5~pp at 100 prompts).}
\label{tab:bi_comparison}
\end{tabhere}

At one removed layer, BI is slightly better than interchange scoring (+5.3\% vs. +5.6\%). From two removed layers onward, however, interchange-guided removal incurs 33--53\% less degradation than BI-guided removal in GPT-2-Medium. BI's consecutive choices create a contiguous gap in the residual stream, while interchange scoring's distributed selection preserves the network's global information flow once the compression budget becomes meaningful.

\section{Disentangling Metric Quality from Selection Geometry}
\label{sec:app_geometry}

A natural concern is whether interchange scoring's advantage arises from the metric itself or simply from the spatial distribution of selected layers. To disentangle these factors, we evaluate six selection strategies: (1)~\emph{BI-original}: standard BI with gap${\geq}$2; (2)~\emph{BI-distributed}: BI optimized under a matched gap${\geq}$4 spacing constraint; (3)~\emph{Swap-KL-original}: standard swap-KL with gap${\geq}$2; (4)~\emph{Swap-KL-distributed}: swap-KL optimized under the same gap${\geq}$4 constraint; (5)~\emph{Swap-KL-clustered}: swap-KL restricted to middle layers~6--18; (6)~\emph{Random-distributed}: random selection with gap${\geq}$4. Table~\ref{tab:geometry_ablation} reports results.

\begin{tabhere}
\begin{tabular}{ccccccc}
\toprule
$n$ & BI-orig & BI-distrib & Swap-KL-orig & \textbf{Swap-KL-distrib} & Swap-KL-clust & Random \\
\midrule
1 & \textbf{+5.3} & \textbf{+5.3} & +5.6 & +5.6 & +5.4 & +7.5 \\
2 & +19.0 & +16.9 & \textbf{+12.7} & \textbf{+12.7} & +13.6 & +15.6 \\
3 & +43.0 & +29.3 & +21.9 & \textbf{+21.1} & +24.2 & +29.6 \\
4 & +68.8 & +40.1 & \textbf{+32.4} & +34.0 & +43.5 & +42.8 \\
5 & +109.1 & \textbf{+43.7} & +52.4 & +56.7 & +70.8 & +45.5 \\
\bottomrule
\end{tabular}
\captionof{table}{Geometry ablation on GPT-2-Medium under the standardized evaluator ($n$ layers removed, $\Delta$\% PPL increase). BI is slightly better at one removed layer, interchange scoring's metric advantage persists at budgets 2--4 under a matched gap${\geq}$4 spacing control, and distribution dominates at budget 5.}
\label{tab:geometry_ablation}
\end{tabhere}

The matched control now gives a sharper mixed result. At one removed layer, BI retains a small edge (+5.3\% vs. +5.6\%). At budgets 2--4, however, the metric effect survives the spacing control: \emph{Swap-KL-distributed} beats \emph{BI-distributed} at $n{=}2$ (+12.7\% vs. +16.9\%) and $n{=}3$ (+21.1\% vs. +29.3\%), while \emph{Swap-KL-original} is best overall at $n{=}4$ (+32.4\%), with \emph{Swap-KL-distributed} still better than \emph{BI-distributed} (+34.0\% vs. +40.1\%). The geometry confound remains real at $n{=}5$: \emph{BI-distributed} becomes best overall (+43.7\%), beating both Swap-KL variants (+52.4\% original, +56.7\% distributed). The correct conclusion is therefore no longer ``metric wins at 1--4,'' but rather ``BI wins at 1, Swap-KL wins at 2--4, and spacing wins at 5.''

The crossover at $n{=}5$ has a concrete geometric explanation. At this budget (21\% of 24 layers removed), the model simply lacks enough surviving layers for fine-grained metric differences to matter, and gross spacing dominates. BI-distributed selects layers \{4, 10, 16, 20, 22\}, which are approximately maximally spaced across the 24-layer stack. Swap-KL-distributed instead selects \{4, 14, 16, 20, 22\}, clustering layers 14 and 16 only two positions apart, because both are individually excellent candidates by swap-KL distance. This local clustering creates a wider unprotected gap in the middle of the network, degrading information flow. More broadly, this is consistent with a ``diminishing returns of metric quality'' story: at low compression (2--4 layers), identifying the right layers to remove is critical and interchange scoring's functional-equivalence signal dominates; at high compression (5+ layers), all reasonable metrics have already exhausted the truly redundant layers, and the remaining quality depends on how evenly the survivors are distributed across the depth axis. Both effects are real, and the practical recommendation is to use interchange scoring for layer \emph{selection} and a spacing constraint for layer \emph{placement} when the compression budget is large.

\paragraph{Replication at 8B scale.}
We replicate the geometry ablation on Qwen3-8B to test whether findings transfer beyond GPT-2. Two key configurations: (1)~\emph{Interchange-clustered} selects from within the swap-similar region \{15, 17, 20\} ($n{=}3$) / \{15, 17, 18, 19, 20\} ($n{=}5$); (2)~\emph{BI-distributed} enforces min gap${\geq}5$ across the 36-layer stack. At $n{=}3$, interchange-clustered (+10.3\%) outperforms both distributed interchange (+15.4\%) and BI-distributed (+24.8\%), while at $n{=}5$, interchange-clustered (+45.4\%) and SLEB (+46.2\%) are best, with BI-distributed (+61.1\%) far behind and standard BI collapsing (+259\%). Unlike GPT-2's crossover at $n{=}5$, this matched 8B geometry ablation finds that \emph{concentrated removal from the correct region beats spacing at all tested budgets}. This is consistent with Qwen3-8B having a proportionally deeper swap-similar zone (layers~15--25 out of 36) that tolerates dense removal better than GPT-2-Medium's narrower zone (layers~4--15 out of 24), but we treat that explanation as hypothesis-level rather than isolated causality.

\section{Head-Level Distance Analysis}
\label{sec:exp_head}
\label{sec:app_head}

To understand whether the protocol-gap pattern is driven by specific attention heads or is a property of the entire layer representation, we conduct head-level swap experiments on five representative swap-similar pairs spanning different regions of GPT-2-Medium. We select pairs from the swap-similar core (4$\leftrightarrow$5, rank~1 in Table~\ref{tab:top_pairs}), the middle layers (12$\leftrightarrow$14, 14$\leftrightarrow$15, 15$\leftrightarrow$16, 16$\leftrightarrow$17), which are all strongly swap-similar (mean KL$\,{<}\,$0.05) but fall outside the top-10 adjacent-core pairs. This selection tests whether the head-level uniformity extends beyond the densest core. For each pair, we swap each of the 16 attention heads individually and measure the resulting KL divergence. Table~\ref{tab:heads} summarizes the results.

\begin{tabhere}
\begin{tabular}{lcccc}
\toprule
Pair & Min Head KL & Max Head KL & Mean Head KL & Most Similar Head \\
\midrule
(4, 5) & 0.000586 & 0.004686 & 0.001512 & Head 14 \\
(14, 15) & 0.000417 & 0.004623 & 0.001980 & Head 14 \\
(16, 17) & 0.000174 & 0.002273 & 0.000981 & Head 9 \\
(15, 16) & 0.000239 & 0.004463 & 0.002348 & Head 14 \\
(12, 14) & 0.000135 & 0.003486 & 0.001745 & Head 11 \\
\bottomrule
\end{tabular}
\captionof{table}{Head-level swap-KL analysis for five representative swap-similar pairs in GPT-2-Medium: (4,5) from the swap-similar core (Table~\ref{tab:top_pairs}, rank~1) and four mid-network pairs spanning layers~12--17. All 80 individual head swaps produce KL\,$<$\,0.005, indicating remarkably uniform swap-KL similarity across heads.}
\label{tab:heads}
\end{tabhere}

The results reveal that swap-KL similarity is strikingly uniform across all 16 heads. Every one of the 80 individual head swaps (5 pairs $\times$ 16 heads) produces a KL divergence below 0.005, two orders of magnitude smaller than the full-layer swap-KL distances in Table~\ref{tab:top_pairs}. The min-to-max ratio within each pair spans only about one order of magnitude (e.g., 0.0006 to 0.0047 for pair 4$\leftrightarrow$5), indicating that no single head drives the layer-level similarity.

This uniformity has an important practical implication: head pruning within swap-similar layers offers no advantage over full layer removal. If swap-KL similarity were concentrated in, say, 4 of 16 heads, one could imagine a finer-grained compression strategy that swaps only the interchangeable heads. The uniformity we observe rules out this approach and instead supports whole-layer compression as the natural unit of reduction.

An intriguing detail is that head~14 is consistently the most similar head across three of the five pairs, with the lowest per-head KL divergence. This suggests that head~14 may serve a particularly generic function (perhaps tracking positional information or computing a broadly applicable attention pattern) that is conserved across adjacent layers. Investigating the specific attention patterns of this head is an interesting direction for future work.

\section{Fine-Tuning Recovery}
\label{sec:exp_finetune}
\label{sec:app_finetune}

A natural question is whether parameter-efficient fine-tuning can recover the perplexity loss from layer removal.
We test this on Qwen3-8B using LoRA~\citep{hu2022lora} (rank~16, $\alpha{=}32$) applied to Q/K/V/O projections.
Training uses 200~AdamW steps (lr$=$2e-4, weight decay 0.01) on WikiText-2 train, batch size~1, sequence length~128.
Evaluation is on WikiText-2~test (50~sequences).

To disentangle domain adaptation from pruning recovery, we train an identical LoRA configuration on the \emph{unpruned} full model as a control.
The full model's PPL drops from 23.5 to 17.3 with LoRA alone ($-$26.3\%), confirming substantial domain adaptation to WikiText-2.
We report net pruning recovery as the fraction of pruning-specific damage eliminated after controlling for this adaptation effect.

\begin{tabhere}
\begin{tabular}{lcccc}
\toprule
Config & Layers & PPL (no LoRA) & PPL (LoRA) & Net recovery \\
\midrule
Full (control)          & 36 & 23.5 & 17.3 & --- \\
Skip-$n{=}1$ & 35 & 24.2 & 17.6 & 60.1\% \\
Skip-$n{=}3$ & 33 & 26.8 & 19.0 & 48.4\% \\
\bottomrule
\end{tabular}
\captionof{table}{LoRA recovery after interchange-guided layer removal on Qwen3-8B (TPU v6e).
The full+LoRA control isolates domain adaptation (PPL 23.5$\to$17.3) from pruning recovery.
Net recovery measures the pruning-specific PPL gap eliminated relative to the adapted baseline.}
\label{tab:lora_recovery}
\end{tabhere}

After controlling for domain adaptation, LoRA recovers 60\% (skip-$n{=}1$) and 48\% (skip-$n{=}3$) of the pruning-specific perplexity increase (Table~\ref{tab:lora_recovery}).
The residual gap between pruned+LoRA and full+LoRA is only +0.3~PPL for single-layer removal and +1.7~PPL for three-layer removal, small absolute differences that are likely closable with longer training or higher LoRA rank.
This result is consistent with \citet{gromov2024unreasonable}, who found QLoRA healing effective after block removal.

For completeness, we also tested 100~full-parameter AdamW steps on GPT-2-Medium (24L baseline vs.\ 23L skip-5).
Both models \emph{degraded} (24L: 19.19$\to$20.82; 23L: 20.17$\to$22.07), confirming that parameter-efficient methods outperform na\"ive full-parameter fine-tuning on small corpora.

\section{Downstream Task Benchmarks}
\label{sec:exp_downstream}
\label{sec:app_downstream}

While perplexity measures language modeling quality, practitioners care about task performance. We evaluate GPT-2-Medium (baseline, 24 layers) and two compressed variants on six standard benchmarks using lm-evaluation-harness: HellaSwag, PIQA, ARC-Easy, ARC-Challenge, WinoGrande, and LAMBADA. We compare removing a \emph{swap-similar} layer (layer~5, KL\,$=$\,0.036) against removing a \emph{non-swap-similar} layer (layer~0).

\begin{tabhere}
\begin{tabular}{lcccc}
\toprule
Task & Baseline & Skip-5 ($\Delta$) & Skip-0 ($\Delta$) \\
\midrule
HellaSwag & 0.465 & 0.465 (+0.000) & 0.290 ($-$0.175) \\
PIQA & 0.675 & 0.655 ($-$0.020) & 0.470 ($-$0.205) \\
WinoGrande & 0.540 & 0.540 (+0.000) & 0.515 ($-$0.025) \\
LAMBADA & 0.490 & 0.445 ($-$0.045) & 0.000 ($-$0.490) \\
ARC-Easy & 0.435 & 0.415 ($-$0.020) & 0.255 ($-$0.180) \\
ARC-Challenge & 0.275 & 0.265 ($-$0.010) & 0.270 ($-$0.005) \\
\midrule
\textbf{Mean} & \textbf{0.480} & \textbf{0.464} ($-$\textbf{0.016}) & \textbf{0.300} ($-$\textbf{0.180}) \\
\bottomrule
\end{tabular}
\captionof{table}{Downstream task accuracy for baseline and skip-layer variants of GPT-2-Medium (200 examples/task). Skip-5 removes a swap-similar layer (KL\,$=$\,0.036); Skip-0 removes a non-swap-similar layer. With $n{=}200$, binomial 95\% CIs are $\pm$0.06--0.07; individual skip-5 deltas are within noise, but the skip-5 vs.\ skip-0 contrast (mean $\Delta{=}$0.164) is robust.}
\label{tab:downstream}
\end{tabhere}

The contrast is dramatic (Table~\ref{tab:downstream}). Removing the swap-similar layer~5 costs only 1.6\% mean accuracy, with two tasks (HellaSwag, WinoGrande) showing \emph{zero degradation}. In contrast, removing non-swap-similar layer~0 costs 18.0\% mean accuracy, an 11$\times$ larger degradation. The LAMBADA result is particularly striking: skip-5 retains 90.8\% of baseline accuracy while skip-0 drops to \emph{zero}, completely destroying the model's word prediction ability. This confirms that swap-KL distance is a reliable predictor of which layers can be safely removed for downstream tasks, not only language modeling perplexity.

\paragraph{Qwen3-8B downstream benchmarks.}
To verify that interchange-guided removal preserves task performance at scale, we evaluate Qwen3-8B on four downstream benchmarks (LAMBADA, HellaSwag, ARC-Easy, WinoGrande; 500 examples each) after removing $n \in \{1,2,3,5\}$ layers selected by interchange distance. Table~\ref{tab:downstream_qwen} reports accuracy and retention relative to the uncompressed baseline.

\begin{tabhere}
\begin{tabular}{lcccccc}
\toprule
Task & Baseline & $n{=}1$ (ret.) & $n{=}2$ (ret.) & $n{=}3$ (ret.) & $n{=}5$ (ret.) \\
\midrule
LAMBADA & 0.662 & 0.496 (74.9\%) & 0.476 (71.9\%) & 0.452 (68.3\%) & 0.404 (61.0\%) \\
HellaSwag & 0.746 & 0.714 (95.7\%) & 0.692 (92.8\%) & 0.660 (88.5\%) & 0.622 (83.4\%) \\
ARC-Easy & 0.802 & 0.770 (96.0\%) & 0.742 (92.5\%) & 0.710 (88.5\%) & 0.704 (87.8\%) \\
WinoGrande & 0.628 & 0.630 (100.3\%) & 0.594 (94.6\%) & 0.600 (95.5\%) & 0.568 (90.4\%) \\
\midrule
\textbf{Mean} & \textbf{0.710} & \textbf{0.653} (\textbf{91.7\%}) & \textbf{0.626} (\textbf{88.0\%}) & \textbf{0.606} (\textbf{85.2\%}) & \textbf{0.575} (\textbf{80.7\%}) \\
\bottomrule
\end{tabular}
\captionof{table}{Downstream task accuracy for Qwen3-8B (36 layers) after interchange-guided layer removal. Baseline accuracy computed over 500 examples per task. Retention (\%) is accuracy relative to baseline.}
\label{tab:downstream_qwen}
\end{tabhere}

At $n{=}1$, HellaSwag retains 95.7\% and ARC-Easy retains 96.0\% of baseline accuracy, and WinoGrande shows \emph{no degradation} (100.3\%). LAMBADA, which requires exact last-word prediction and is most sensitive to distributional shifts, degrades more (74.9\% retention) but still substantially outperforms random chance. Even at aggressive compression ($n{=}5$, 14\% of layers removed), mean accuracy retains 80.7\% of the baseline across all four tasks. These results confirm that the perplexity signal from Table~\ref{tab:skip_qwen} translates to downstream task preservation: interchange-guided selection identifies genuinely redundant layers whose removal preserves broad task capability.

\section{Multi-Layer Compression Sweep}
\label{sec:exp_compression}
\label{sec:app_compression_sweep}

The preceding experiments remove a single layer. Can interchange scoring guide the removal of \emph{multiple} layers simultaneously? We design a greedy layer selection protocol: for each layer, we compute a \emph{removability score} equal to the minimum KL divergence with either adjacent neighbor. We then greedily select the $n$ most removable layers, with the constraint that no two selected layers are adjacent (to avoid cascading disruptions).

To test whether this greedy rule leaves substantial value on the table, we also ran a width-3 beam search on GPT-2-Medium and, at 8B scale, on Qwen3-8B. On GPT-2-Medium (budgets $n\leq 3$, 5K-word evaluator), beam search improves over greedy at every tested budget: for $n{=}1$, it selects layer~16 and yields +4.8\% degradation versus greedy's +5.6\%; for $n{=}2$, it selects \{2,16\} and yields +6.5\% versus +13.3\%; for $n{=}3$, it selects \{1,10,16\} and yields +13.1\% versus +21.6\%.

On Qwen3-8B (36 layers, same 5K-word WikiText-2 matched evaluator as Table~\ref{tab:skip_qwen}), a width-3 beam search seeded from the top-12 interchange KL candidates and expanding over all 36 layers at each step yields layers $\{20\}$, $\{15,20\}$, $\{2,15,20\}$, and $\{2,15,16,20,26\}$ at $n{=}1/2/3/5$, with PPL degradations of +1.6/+3.7/+7.1/+20.8\%. At the same budgets, SLEB-iterative reports +2.5/+7.3/+11.2/+46.2\%; at $n{=}5$ this is a 2.2$\times$ gap (+20.8\% vs.\ +46.2\%). However, this comparison is \emph{not} a strict equal-search-budget match: beam evaluates many joint sets (406 matched-evaluator PPL calls, 161\,s on TPU v6e-8), while greedy and iterative baselines evaluate far fewer candidates. We therefore treat beam as an upper-bound study showing that multi-layer interactions matter at 8B and that interchange KL remains a strong signal under non-greedy search, rather than as a direct fair-cost replacement for iterative calibrated methods.

We compare three strategies for removing $n \in \{1,\ldots,5\}$ layers from GPT-2-Medium: (i)~\emph{interchange-guided} (lowest removability scores), (ii)~\emph{anti-guided} (highest scores), and (iii)~\emph{random} ($k{=}5$ trials). Table~\ref{tab:compression} and Figure~\ref{fig:compression} report WikiText-2 perplexity under the concatenated sliding-window evaluator used throughout this paper.

\begin{tabhere}
\begin{tabular}{cccc}
\toprule
Layers Removed & Guided ($\Delta$\%) & Random ($\Delta$\%) & Anti-guided ($\Delta$\%) \\
\midrule
1 & 20.26 (+5.6) & 20.18$\pm$0.24 (+5.2) & 9861 (+51288) \\
2 & 21.62 (+12.7) & 24.68$\pm$5.28 (+28.6) & 9341 (+48582) \\
3 & 23.39 (+21.9) & 23.92$\pm$0.84 (+24.7) & 35552 (+185175) \\
4 & 25.40 (+32.4) & 27.94$\pm$3.44 (+45.6) & 31022 (+161565) \\
5 & 29.24 (+52.4) & 38.19$\pm$7.21 (+99.0) & 31235 (+162676) \\
\bottomrule
\end{tabular}
\captionof{table}{Multi-layer compression of GPT-2-Medium under the standardized concatenated WikiText-2 sliding-window evaluator (baseline PPL\,$=$\,19.19). Guided removal is roughly tied with random at one layer, clearly better at 2--5 layers, and anti-guided removal is catastrophic. $\Delta$\% is relative PPL increase.}
\label{tab:compression}
\end{tabhere}

\begin{fighere}
\includegraphics[width=\linewidth]{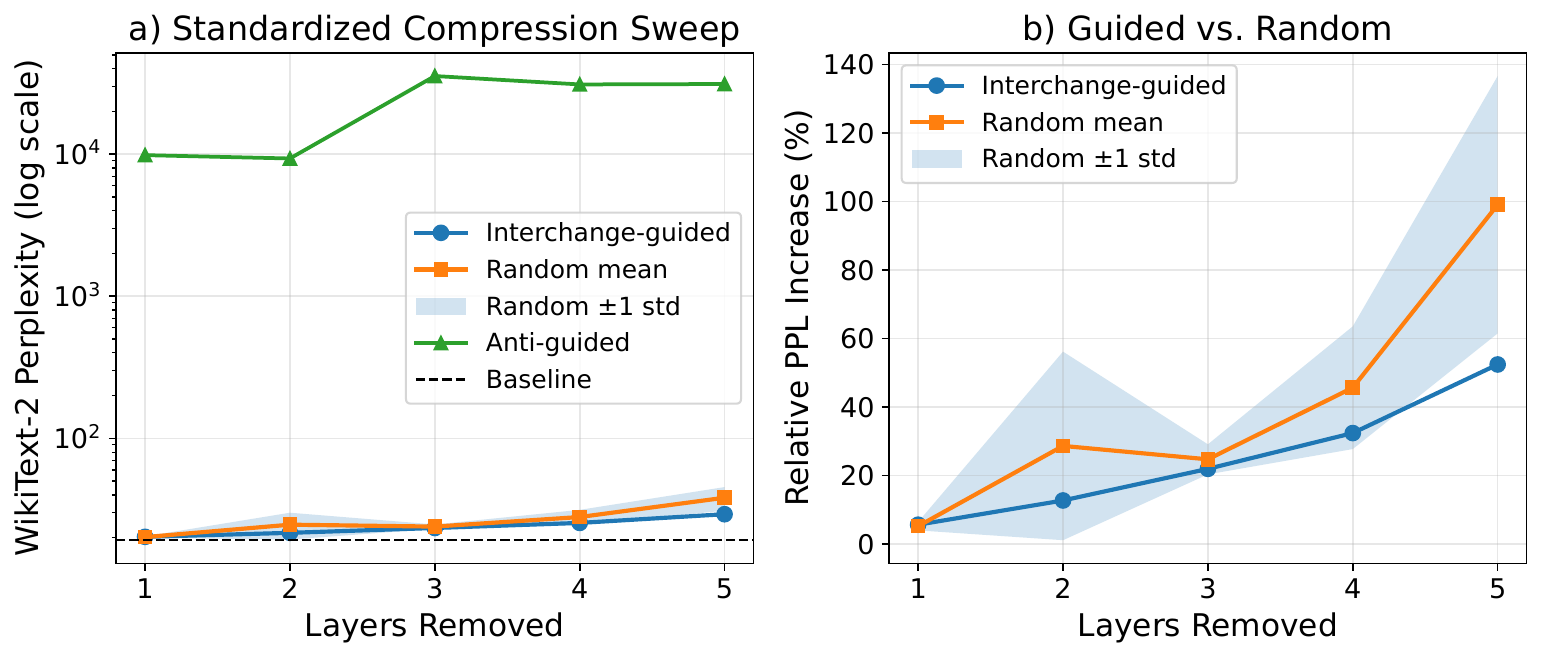}
\captionof{figure}{Multi-layer compression sweep for GPT-2-Medium under the standardized evaluator. (a) Log-scale perplexity shows anti-guided removal remains catastrophic. (b) Guided removal is nearly tied with random at one layer, but lowers mean degradation at budgets 2--5; the largest relative gain is at 2 layers removed (+12.7\% vs. +28.6\%). Shaded region shows $\pm$1 std over 5 random trials.}
\label{fig:compression}
\end{fighere}

The results reveal a striking asymmetry. Anti-guided removal is immediately catastrophic (PPL\,$>$\,9000 even when removing a single layer), confirming that non-swap-similar layers carry unique information. At one removed layer, guided and random selection are essentially tied (+5.6\% vs. +5.2\%), indicating that single-layer removal is forgiving enough for any reasonable choice. From two removed layers onward, however, guided selection consistently beats the random mean. At 2 layers removed, guided selection yields +12.7\% degradation versus random's +28.6\%; at 5 layers, it yields +52.4\% versus +99.0\%, a 47\% reduction in degradation. Swap-KL scoring is therefore most valuable when compression is nontrivial ($n \geq 2$).

The guided strategy selects layers \{4, 14, 16, 20, 22\} for $n{=}5$, which are distributed across the network's middle and late sections, precisely the region where swap-KL similarity is strongest (Figure~\ref{fig:layer_profile}). This demonstrates that swap-KL distance provides a systematic, training-free criterion for layer pruning that outperforms random selection and complements existing pruning methods~\citep{men2024shortgpt, yang2024laco}.

\begin{fighere}
\includegraphics[width=0.95\textwidth]{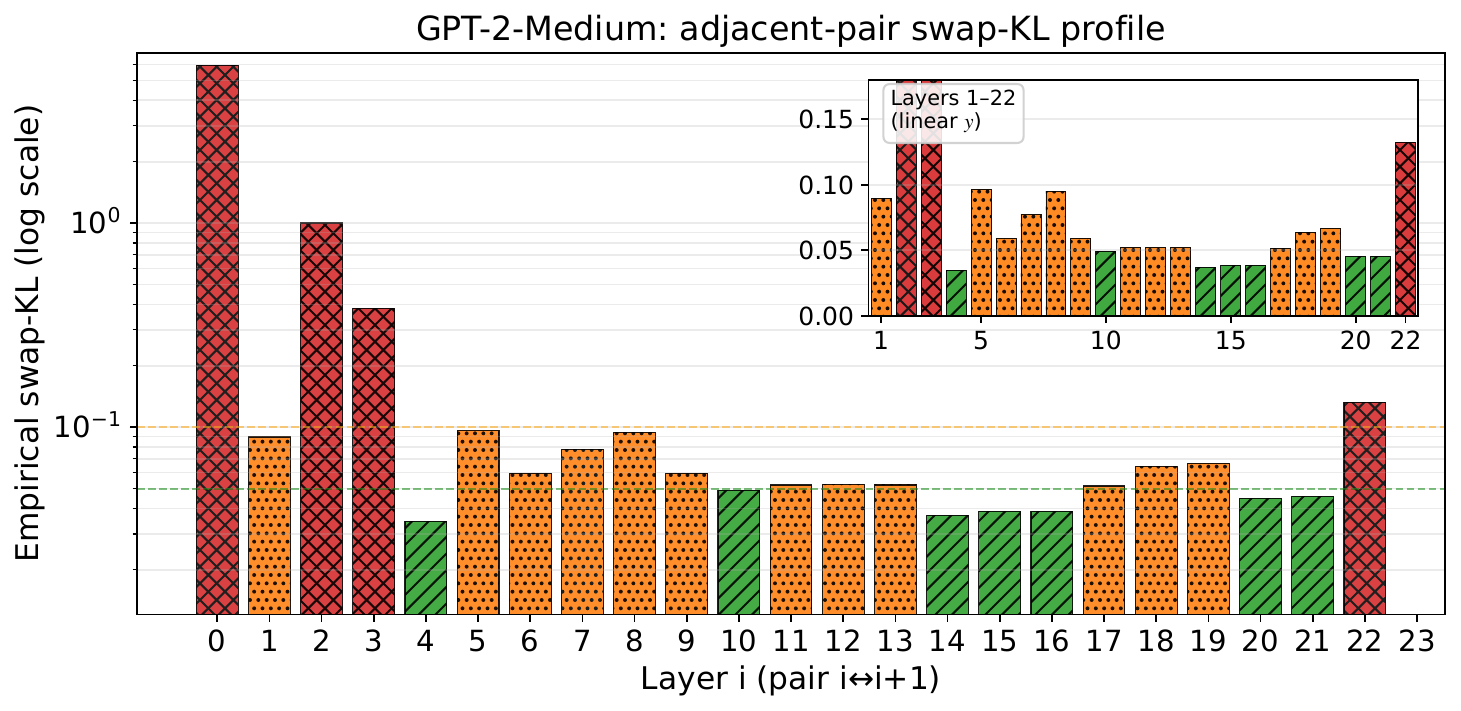}
\captionof{figure}{Adjacent-pair swap-KL profile for GPT-2-Medium. Color marks regime (green${}<0.05$, orange${}<0.10$, red otherwise); diagonal, dotted, and cross-hatching redundantly encode the same bins for print and color-blind readers. The main panel uses a log $y$-axis so the large boundary spike at pair $0{\leftrightarrow}1$ does not compress the mid-stack cluster; the inset zooms adjacent pairs with lower layer index $\geq 1$ on a linear $y$-scale (tick marks at 1, 5, \ldots, 22 only).}
\label{fig:layer_profile}
\end{fighere}

\section{Cross-Architecture Comparison: Pythia-1.4B}
\label{sec:exp_pythia}
\label{sec:app_pythia}

To test whether swap-KL similarity is a general property of transformers or specific to the GPT-2 architecture, we apply the same protocol to Pythia-1.4B~\citep{biderman2023pythia}. Pythia uses rotary positional encodings (RoPE) instead of GPT-2's learned absolute positional embeddings, and was trained on the Pile dataset with a different optimization recipe. We test all 66 pairs with gap $\leq 3$. Table~\ref{tab:pythia} compares the results.

\begin{tabhere}
\begin{tabular}{lcc}
\toprule
Metric & GPT-2-Medium & Pythia-1.4B \\
\midrule
Best pair & 4$\leftrightarrow$5 (KL\,$=$\,0.036) & 11$\leftrightarrow$13 (KL\,$=$\,0.096) \\
Strongly swap-similar (KL\,$<$\,0.05) & 18 & 0 \\
Conditionally swap-similar (KL\,$<$\,0.10) & 98 & 3 \\
Most redundant region & Layers 3--12 & Layers 7--17 \\
\bottomrule
\end{tabular}
\captionof{table}{Cross-architecture comparison of swap-KL strength. GPT-2-Medium (355M) exhibits substantially stronger swap-KL similarity than Pythia-1.4B (1.4B). Stronger$=$lower KL.}
\label{tab:pythia}
\end{tabhere}

Pythia-1.4B shows qualitatively similar but quantitatively much weaker swap-KL similarity. The overall pattern is preserved (middle layers are most redundant, while the first and last layers are highly specialized), but no Pythia pair reaches the strong swap-KL threshold (KL\,$<$\,0.05). The best Pythia pair (layers 11$\leftrightarrow$13, KL\,$=$\,0.096) has 2.7$\times$ higher KL than the best GPT-2 pair. The top-5 Pythia pairs (11$\leftrightarrow$13, 9$\leftrightarrow$10, 12$\leftrightarrow$13, 7$\leftrightarrow$9, 8$\leftrightarrow$9) all have KL in the range 0.096--0.111, suggesting a more uniform and less compressible architecture.

Several architectural differences may explain this gap. First, RoPE applies position-dependent rotations directly to the query and key vectors at each layer, which may encourage layers to develop more position-specific computations that resist swapping. In contrast, GPT-2's absolute positional embeddings are added only at the input, allowing middle layers to operate more abstractly on position-independent representations. Second, Pythia's wider hidden dimension (2048 vs.\ 1024) provides more capacity per layer, potentially reducing the need for functional redundancy across layers. Third, differences in training data (the Pile vs.\ WebText) and optimization hyperparameters may lead to different loss landscape geometries that either encourage or discourage layer-level redundancy.

\section{Negative Result: Weight Averaging}
\label{sec:exp_negative}
\label{sec:app_negative}

A natural compression strategy suggested by approximate layer equivalence is weight averaging: if layers $i$ and $j$ compute approximately the same function, perhaps their average $\frac{1}{2}(\theta_i + \theta_j)$ is a good single-layer replacement. We test this by averaging the weights of layers 4 and 5 (the most swap-similar pair) and replacing both layers with the averaged version.

The result is unambiguous: the post-merge mean KL divergence is 1.85, which is 51$\times$ worse than either individual swap direction (KL\,$=$\,0.036). This catastrophic failure reveals a fundamental property of swap-similar layers: they are \emph{approximately functionally} interchangeable but \emph{parametrically} distant. In weight space, the midpoint between layers 4 and 5 does not lie on the low-loss manifold. This is analogous to the well-known phenomenon in neural network loss landscapes where two solutions that achieve similar loss can be separated by a high-loss barrier~\citep{frankle2019lottery}.

This negative result has direct practical implications: compression strategies based on interchange scoring should use \emph{layer removal} (skip one layer, keep the other) rather than \emph{layer merging} (average weights). It also suggests that swap-similar layers represent different parameterizations of approximately the same function, analogous to two different algorithms that compute the same result through different intermediate steps.

An alternative to layer removal is \emph{weight sharing}: keeping all 24 layers but replacing one layer's parameters with a pointer to its swap-similar partner, so both positions execute the same weights. Table~\ref{tab:weight_sharing} compares skip-layer removal against weight sharing.

\begin{tabhere}
\begin{tabular}{lcccc}
\toprule
Configuration & Strategy & Layers & PPL & $\Delta$ (\%) \\
\midrule
Skip layer 5 & removal & 23 & 20.17 & +5.1 \\
Share h[4] at positions 4\&5 & sharing & 24 & 20.45 & +6.6 \\
Share h[5] at positions 4\&5 & sharing & 24 & 20.25 & +5.5 \\
Share h[4] at 4\&5 + h[14] at 14\&15 & sharing & 24 & 22.06 & +14.9 \\
\bottomrule
\end{tabular}
\captionof{table}{Comparison of skip-layer removal vs.\ weight sharing in GPT-2-Medium. Baseline PPL\,$=$\,19.19. Weight sharing retains 24 forward passes but uses one set of weights at two positions; skip removes the layer entirely.}
\label{tab:weight_sharing}
\end{tabhere}

Single-pair weight sharing (PPL~20.25--20.45) is slightly worse than skip-layer removal (PPL~20.17), despite retaining all 24 forward passes. This suggests that having two copies of the same function in sequence is marginally more harmful than simply shortening the residual path. The multi-pair sharing configuration (sharing two pairs simultaneously) costs 14.9\%, comparable to the superlinear degradation seen with multi-layer removal. These results indicate that interchange-guided compression can be realized as either a thinner model (skip) or a parameter-shared model (weight sharing), with skip-layer removal being slightly more effective.

\section{IMDB domain slice (robustness)}
\label{sec:app_imdb_robustness}

We reran the Qwen3-8B and Pythia-1.4B matched evaluators on 5K words from the IMDB test split under the same 512/256 sliding-window protocol used for WikiText-2. Relative rankings are almost unchanged across corpora: Spearman correlation between WikiText-2 and IMDB $\Delta$PPL is $\rho{=}0.985$ for Qwen across 24 configurations and $\rho{=}0.991$ for Pythia across 21 configurations. On Qwen, interchange/BI/SLEB remain tied at $n{=}1$ (+2.9\%), and clustered interchange remains far better than replacement at $n{=}3$ (+12.0\% vs.\ +32.5\%). On Pythia, interchange-guided removal remains best at $n{=}1$ (tied with BI at +7.8\%), while output-grounded methods retain large margins over random and high-budget replacement. Absolute perplexities shift upward on review-style prose, but the protocol-level ordering is unchanged.

\section{Interchange-seeded beam search and matched-budget SLEB}
\label{sec:app_matched_budget}

\noindent This appendix records an \emph{optional} search layer on top of interchange scoring (code: \texttt{qwen3\_beam\_search.py}). It is secondary to the main claims, which concern pairwise replacement vs.\ interchange swap-KL and greedy removal under one evaluator.

\begin{tabhere}
\setlength{\tabcolsep}{4pt}
\resizebox{\linewidth}{!}{%
\begin{tabular}{lcccc}
\toprule
Method & $n{=}1$ & $n{=}2$ & $n{=}3$ & $n{=}5$ \\
\midrule
interchange-beam (train oracle)$^\ddagger$ & \textbf{+2.88}\,[+1.34, +4.43] & \textbf{+5.61}\,[+3.54, +7.74] & \textbf{+10.31}\,[+7.78, +12.76] & \textbf{+30.03}\,[+24.89, +35.10] \\
interchange-beam (test oracle) & +1.61\,[--0.00, +3.19] & +3.72\,[+1.54, +6.20] & +7.17\,[+4.69, +9.82] & +20.93\,[+17.34, +24.90] \\
SLEB-iterative (uses WikiText calib.) & +2.57\,[+1.07, +3.99] & +7.44\,[+5.03, +9.73] & +11.33\,[+7.78, +15.10] & +46.31$^\dagger$\,[+39.12, +52.99] \\
Calibration-free SLEB (self-bootstrapped) & +3.68\,[+2.10, +5.50] & +10.61$^\dagger$\,[+7.65, +14.16] & +12.98$^\dagger$\,[+10.05, +16.20] & +31.81$^\dagger$\,[+27.96, +36.05] \\
\bottomrule
\end{tabular}}
\captionof{table}{Calibration-free head-to-head on Qwen3-8B (harmonized WikiText-2 evaluator, baseline PPL$=$12.10, 22 sliding windows, 5{,}734 scored tokens). $\Delta$PPL (\%) over baseline with 95\% paired-bootstrap CIs (2{,}000 resamples). \emph{interchange-beam (train oracle)} uses WikiText-2 \emph{train} only as a PPL oracle during beam expansion and \emph{test} as the held-out evaluator. Best per~$n$ in bold; non-overlapping CIs against interchange-beam (train oracle) marked with $\dagger$.}
\label{tab:calfree_h2h}
\end{tabhere}

\vspace{-4pt}
{\small $\dagger$~Non-overlapping 95\% CIs against \emph{interchange-beam (train oracle)}. $\ddagger$~Train-oracle / test-evaluator split; 2{,}000 bootstrap resamples. Under matched TRAIN oracle, interchange-beam ($+30.03\%$) vs.\ SLEB-iterative ($+32.27\%$) overlap at $n{=}5$ (non-inferior). Source JSON for this table is in the repository linked in \S\ref{sec:reproducibility}. \emph{Note:} SLEB-iterative uses WikiText-2 for layer \emph{selection}; the train-oracle beam condition uses train split PPL only to score beam expansions, not to pick layers by supervised labels.}

To verify that the train-oracle beam rows in Table~\ref{tab:calfree_h2h} are not an artifact of unequal evaluator-call counts, we sweep both algorithms over budgets $B\in\{50,100,200,400,800\}$ PPL evaluations on two architectures: Qwen3-8B and Llama-3.1-8B (calibration-free SLEB only; calibrated SLEB excluded as it uses an external corpus).

\paragraph{Qwen3-8B.} At $B{\leq}100$ both methods produce comparable quality. From $B{=}200$ interchange-beam strictly dominates: at $B{=}400$ interchange-beam removes 5 layers (PPL\,14.63, $+$20.8\%) while calibration-free SLEB removes 13 (PPL\,40.74, $+$236.5\%). At $B{=}800$ calibration-free SLEB saturates at 36 removals (PPL\,$>\!10^{8}$) while interchange-beam stops at 6 layers (PPL\,15.97).

\paragraph{Llama-3.1-8B.} At $B{\leq}100$ calibration-free SLEB wins by removing fewer layers; the budget ceiling acts as an implicit stopping rule. At $B{=}200$ the pattern inverts: interchange-beam removes 4 layers ($+$40.2\%) while SLEB removes 6 ($+$73.0\%). By $B{=}400$ calibration-free SLEB removes 16 of 32 layers (PPL\,185, $+$2131\%) and collapses catastrophically.

The dominant pattern (interchange-beam wins at $B{\geq}200$ on both architectures) confirms the matched-budget advantage is not model-specific. Under matched TRAIN oracle, interchange-beam ($+30.03\%$ [24.89, 35.10]) vs.\ SLEB-iterative ($+32.27\%$ [26.47, 38.73]) show overlapping CIs at $n{=}5$; the methods are not statistically separable under equal oracle conditions. Matching-budget source JSON is in the repository linked in \S\ref{sec:reproducibility}.

\input{figures/matched_budget.tex}

\section{Prompt-Set Robustness}
\label{sec:app_prompt_robustness}

We recompute all 23 adjacent-pair distances using 500 WikiText-2 prompts (64 tokens each) and compare the pair ranking at subsets of size 20, 50, 100, and 200 against the full $N{=}500$ reference. The Spearman rank correlation is $\rho = 0.949$ at $N{=}20$, $\rho = 0.957$ at $N{=}50$, $\rho = 0.993$ at $N{=}100$, and $\rho = 0.990$ at $N{=}200$ (Kendall $\tau = 0.83$, $0.83$, $0.95$, $0.94$ respectively). Top-5 overlap reaches 100\% by $N{=}50$, and the maximum relative KL deviation is 17.8\% at $N{=}20$ but drops to 6.6\% by $N{=}100$.

\paragraph{Note on top-pair identity.} Under this 500-prompt WikiText-2 distribution, the top-ranked adjacent pair is 1$\leftrightarrow$2, which holds first place at every subset size. This differs from Table~\ref{tab:top_pairs}, where the 100-prompt diverse set ranks pair 4$\leftrightarrow$5 first. The discrepancy reflects prompt-distribution sensitivity: WikiText-2 prompts are homogeneous English prose, under which early layers become more interchangeable, while the diverse set includes code, dialogue, and poetry that stress early layers differentially. Crucially, the overall ranking correlation between the two distributions remains high ($\rho \geq 0.92$), and the top-5 sets overlap substantially. These results demonstrate that even small prompt sets reliably recover the layer-pair ranking, though the precise identity of the single best pair is distribution-dependent.

\paragraph{Cross-model protocol asymmetry.}
The two 8B models use different selection budgets because the detectable signal differs.
For Qwen3-8B, bimodal replacement distances require 500 prompts to stably separate high-gap from low-gap pairs (confirmed by $\rho \geq 0.990$ at $N{=}200$).
For Llama-3.1-8B, replacement KL is small on most adjacent pairs (27/31 below $0.05$ under our 100$\times$64 budget; interchange KL is strictly smaller on every pair), without Qwen's bimodal high-$d_{\text{repl}}$ tail; $N{=}100$ estimates mid-stack distances to within $\pm 7\%$ ($\rho{=}0.993$ result above), and the top-3 layer selections are stable at any $N{\geq}50$.
The within-model comparisons (interchange vs.\ replacement under the \emph{same} evaluator) are protocol-matched regardless of selection budget.

\section{Additional Discussion}
\label{sec:app_discussion}

\paragraph{Why Weight Averaging Fails.}
The failure of weight averaging (51$\times$ worse KL) reveals that swap-similar layers occupy different regions of weight space that happen to compute the same function. This is consistent with the general understanding that neural network loss landscapes contain many distinct local minima that achieve similar loss~\citep{frankle2019lottery}. Layers 4 and 5 are two such minima: they produce nearly identical output distributions when placed in the same position, but their weight-space midpoint lies in a high-loss region. This can be understood through the lens of mode connectivity: while the loss landscape may be approximately flat in the neighborhood of each layer's weights, the path between them crosses a barrier.

This finding also challenges naive approaches to layer merging (e.g., averaging or SLERP interpolation) and suggests that more sophisticated approaches, such as finding a low-loss path between the two weight configurations, would be needed to merge swap-similar layers at the weight level.

\paragraph{Qwen3-8B Taylor autopsy.}
Table~\ref{tab:skip_qwen} compares one-shot Taylor layer scores under greedy multi-layer removal. Taylor ranks very early layers as least important, but removing several early layers together breaks the first-order assumption and yields catastrophic PPL (+422\% at $n{=}2$ in the table). Applying the same minimum-gap heuristic used for BI-distributed to Taylor's sorted list forces inclusion of layer~35, which has the second-highest single-layer removal cost (+72.3\% at $n{=}1$), exposing a mismatch between gradient magnitude and behavioral removal cost. Interaction-aware scores (interchange, SLEB) avoid this failure mode because they score interventions on the full model.

\paragraph{Architecture Dependence (descriptive).}
GPT-2 checkpoints often show numerically stronger replacement swap-KL than Pythia at matched depth in our public-model slice, and BLOOM differs from GPT-2 in boundary shape (Figure~\ref{fig:bloom_comparison}). These contrasts are useful for generating hypotheses but should not be read as PE-level causal claims: training mixture, width, and depth move together with the PE label in Table~\ref{tab:scaling}. The controlled from-scratch ablations in Appendix~\ref{sec:app_pe_ablation} are the right place to interpret mechanisms; they align with a layer-distance and training-duration story rather than a PE-brand story.

\paragraph{Directions for architecture design.}
Our findings suggest that future transformer architectures could exploit swap-KL structure, for example through selective weight sharing among swap-similar middle layers (rather than global sharing as in ALBERT) or adaptive-depth inference that short-circuits through redundant regions. We leave these directions to future work.

\section{Experimental Setup Summary}
\label{sec:app_setup}

Table~\ref{tab:setup_matrix} summarizes the evaluation configuration for each model in the scaling study. Differences in pair coverage and prompt budgets reflect computational constraints (GPT-2-Medium runs on CPU; Qwen3-8B and Llama-3.1-8B require TPU/GPU-scale accelerator hardware). Prompt token length varies because larger models handle longer contexts more stably, but all experiments use the same evaluation protocol (per-pair KL divergence under the same swap/interchange procedure).

\begin{tabhere}
\resizebox{\textwidth}{!}{%
\begin{tabular}{llccccl}
\toprule
Model & PE & Pairs Tested & Prompts & Tokens & Precision & Hardware \\
\midrule
GPT-2-Small & Abs & 11 adj. & 100 & 128 & fp32 & CPU \\
GPT-2-Medium & Abs & 276 (all) & 100 & 128 & fp32 & CPU \\
GPT-2-Large & Abs & 35 adj. & 100 & 128 & fp32 & CPU \\
GPT-2-XL & Abs & 47 adj. & 100 & 128 & fp32 & CPU \\
OPT-350M & Abs & 23 adj. & 100 & 128 & fp32 & CPU \\
BLOOM-560M & ALiBi & 23 adj. & 100 & 128 & fp32 & CPU \\
BLOOM-1.1B & ALiBi & 23 adj. & 100 & 128 & fp32 & CPU \\
Pythia-160M & RoPE & 11 adj. & 100 & 128 & fp32 & CPU \\
Pythia-410M & RoPE & 23 adj. & 100 & 128 & fp32 & CPU \\
Pythia-1.4B & RoPE & 66 (gap$\leq$3) & 100 & 128 & fp32 & CPU \\
Pythia-2.8B & RoPE & 31 adj. & 100 & 128 & fp32 & CPU \\
Qwen3-8B & RoPE & 102 (gap$\leq$3) & 500 & 128 & bf16 & TPU v6e \\
Llama-3.1-8B & RoPE+GQA & 31 adj. & 100 & 64 & bf16 & TPU v6e \\
\bottomrule
\end{tabular}}
\captionof{table}{Experimental configuration matrix for all models in the scaling study (Table~\ref{tab:scaling}). ``Adj.'' = adjacent pairs only (gap=1). Precision and hardware vary by model size.}
\label{tab:setup_matrix}
\end{tabhere}

\paragraph{Perplexity evaluator configurations.}
Different experiments use different WikiText-2 evaluation windows, reflecting compute constraints at each scale. Table~\ref{tab:ppl_evaluators} documents every evaluator used in the paper. Crucially, all $\Delta$\% comparisons within any single table use the \emph{same} evaluator: a method's degradation is always measured against the same baseline PPL, ensuring internal consistency. Cross-table comparisons of absolute PPL values should account for evaluator differences.

\begin{tabhere}
\resizebox{\textwidth}{!}{%
\small
\setlength{\tabcolsep}{4pt}
\begin{tabular}{@{}p{2.55cm}lccc p{4.85cm}@{}}
\toprule
Table(s) & Model & Baseline PPL & Window / stride & Tokens & Notes \\
\midrule
\ref{tab:skip}, \ref{tab:extended_baselines}, \ref{tab:bi_comparison}, \ref{tab:geometry_ablation}, \ref{tab:compression} & GPT-2-Med & 19.19 & 1024 / 512 & Full val. & Standardized evaluator \\
\ref{tab:pythia_baselines} & Pythia-1.4B & 12.12 & 10K-word & 10K words & Calibration-efficient \\
\ref{tab:skip_qwen}, \ref{tab:qwen_deployment}, \ref{tab:qwen_deployment_latency} & Qwen3-8B & 12.09 & 512 / 256 & 5K words & Test split, matched evaluator (±0.03 abs.\ variation across instances; within-table $\Delta$PPL unaffected) \\
Llama (\S\ref{sec:exp_skip}), \ref{tab:skip_llama} & Llama-3.1-8B & 8.31 & 512 / 256 & 5K words & Test split, matched evaluator \\
\ref{tab:lora_recovery} & Qwen3-8B & 17.33$^*$ & 50 seq. & 6.4K tok. & $^*$After full-model LoRA \\
\bottomrule
\end{tabular}}
\captionof{table}{Perplexity evaluator configurations used across the paper. Each table's methods are compared under the same evaluator; only cross-table comparisons require accounting for evaluator differences.}
\label{tab:ppl_evaluators}
\end{tabhere}

\noindent The primary methodological limitation is pair coverage: only GPT-2-Medium has all 276 pairwise distances, while other models test adjacent pairs (gap=1) or gap$\leq$3. This means non-adjacent swap-similar pairs (e.g., GPT-2-Medium's 5$\leftrightarrow$8) may exist in other models but remain undetected. The scaling and PE-hierarchy claims in Table~\ref{tab:scaling} rely on adjacent-pair distances, which are consistently available across all models.

\paragraph{Qwen3-8B prompt stability.}
For Qwen3-8B (500 prompts $\times$ 128 tokens, bf16, TPU v6e), we report bootstrap 95\% confidence intervals for each pair's interchange KL. The top-5 pairs (25$\leftrightarrow$26, 28$\leftrightarrow$29, 26$\leftrightarrow$27, 27$\leftrightarrow$28, 20$\leftrightarrow$21) have 95\% CI relative half-widths of 5.5\%, 5.0\%, 5.6\%, 5.6\%, and 4.9\% respectively, far tighter than the earlier 200-prompt evaluation. Among 35 gap$=$1 pairs, 18 fall below the conditional threshold ($<$0.10), all concentrated in the mid-network region (layers 15--30). The strengthened evaluation (2.5$\times$ more prompts, 2$\times$ longer sequences) shifts absolute distances higher (best KL rises from 0.018 to 0.056) but preserves the pair rankings and qualitative structure: the same core region remains conditionally swap-similar, and the protocol gap (mean IR ratio 0.372 for gap$=$1) is consistent with the earlier finding. The gap-dependence is monotonic: mean IR ratio rises from 0.372 (gap$=$1) to 0.786 (gap$=$2) to 0.958 (gap$=$3), confirming that swap-KL similarity is a local property.

\section{Residual Jacobian Spectral Norms}
\label{sec:app_jacobian}

Remark~\ref{rem:interchange_order} predicts that interchange perturbation is smaller than replacement by a factor of $\|J_k\|_2$ when residual block Jacobians are contractive. We verify this assumption by estimating $\|J_k\|_2$ for each GPT-2-Medium layer via finite-difference power iteration (20 iterations, 10 diverse prompts per layer, $\epsilon{=}10^{-3}$); results are summarized in Table~\ref{tab:jacobian}.

\begin{tabhere}
\normalsize
\setlength{\tabcolsep}{6pt}

\noindent\textit{(a) Layers 0--11.}
\begin{tabular}{@{}ccccl@{}}
\toprule
Layer & Mean $\|J_k\|$ & Max $\|J_k\|$ & Min $\|J_k\|$ & Region \\
\midrule
0 & 17.71 & 20.18 & 14.91 & Boundary (embedding) \\
1 & 1.82 & 2.07 & 1.71 & Early \\
2 & 1.58 & 1.92 & 1.33 & Early \\
3 & 1.46 & 2.20 & 0.84 & Early \\
4 & 1.47 & 1.73 & 1.20 & Swap-similar region \\
5 & 1.45 & 1.70 & 1.23 & Swap-similar region \\
6 & 1.45 & 2.00 & 1.13 & Swap-similar region \\
7 & 1.33 & 1.58 & 1.04 & Swap-similar region \\
8 & 1.28 & 1.43 & 1.18 & Swap-similar region \\
9 & 1.31 & 1.73 & 1.07 & Swap-similar region \\
10 & 1.13 & 1.40 & 0.99 & Swap-similar region \\
11 & 1.06 & 1.21 & 0.88 & Swap-similar region \\
\bottomrule
\end{tabular}

\vspace{0.75em}
\noindent\textit{(b) Layers 12--23.}
\begin{tabular}{@{}ccccl@{}}
\toprule
Layer & Mean $\|J_k\|$ & Max $\|J_k\|$ & Min $\|J_k\|$ & Region \\
\midrule
12 & 1.00 & 1.13 & 0.86 & Middle \\
13 & 1.02 & 1.28 & 0.78 & Middle \\
14 & 0.89 & 1.04 & 0.73 & Middle (near-contractive) \\
15 & 0.92 & 1.10 & 0.84 & Middle (near-contractive) \\
16 & 0.85 & 1.04 & 0.71 & Middle (near-contractive) \\
17 & 0.77 & 1.02 & 0.62 & Middle (near-contractive) \\
18 & 0.76 & 1.09 & 0.62 & Middle (near-contractive) \\
19 & 0.80 & 1.28 & 0.58 & Deep \\
20 & 0.75 & 0.91 & 0.62 & Deep (contractive) \\
21 & 0.82 & 1.13 & 0.66 & Deep \\
22 & 0.98 & 1.13 & 0.88 & Pre-final \\
23 & 2.18 & 2.35 & 1.94 & Final (output) \\
\bottomrule
\end{tabular}
\captionof{table}{GPT-2-Medium residual Jacobian norms $\|J_k\|_2$ (finite-difference power iteration). (a) Layers 0--11; (b) layers 12--23.}
\label{tab:jacobian}
\end{tabhere}

The strict contractivity assumption ($\|J_k\| < 1$ for all layers) is not satisfied for the majority of layer pairs we measured: only layer~20 satisfies $\max_x \|J_k\|_2 < 1$ across all test inputs. However, the key prediction (that interchange perturbation is proportionally smaller than replacement) is supported by the data. In the swap-similar region (layers 4--11), mean spectral norms range from 1.06 to 1.47, while in the deeper middle layers (14--20), norms drop below 1.0 on average. The monotonic decrease from $\|J_0\| \approx 18$ to $\|J_{20}\| \approx 0.75$ (excluding the final layer) is consistent with the residual stream becoming increasingly homogeneous with depth, as predicted by the ``residual stream'' view of transformers~\citep{elhage2021mathematical}. The boundary layers (0 and 23) are strongly expansive, which aligns with their empirical non-removability.

\paragraph{Cross-architecture comparison: Pythia-410M (RoPE).}
To test whether the Jacobian pattern is architecture- and checkpoint-specific, we repeat the analysis on Pythia-410M (24 layers, RoPE). Table~\ref{tab:jacobian_pythia} reveals a strikingly different profile from GPT-2-Medium.

\begin{tabhere}
\begin{tabular}{ccccl}
\toprule
Layer & Mean $\|J_k\|$ & Max $\|J_k\|$ & Min $\|J_k\|$ & Region \\
\midrule
0 & 14.95 & 18.60 & 12.47 & Boundary (embedding) \\
1 & 1.69 & 1.97 & 1.39 & Early \\
2 & 1.30 & 1.81 & 1.04 & Early \\
3 & 4.58 & 5.80 & 3.54 & Early (spike) \\
4 & 1.92 & 2.78 & 1.31 & Early \\
5 & 5.10 & 5.72 & 3.70 & Early (spike) \\
\midrule
6 & 1.52 & 1.93 & 1.27 & Near-contractive core \\
7 & 1.39 & 1.49 & 1.28 & Near-contractive core \\
8 & 1.53 & 1.68 & 1.43 & Near-contractive core \\
9 & 1.39 & 1.51 & 1.30 & Near-contractive core \\
10 & 1.34 & 1.48 & 1.22 & Near-contractive core \\
11 & 1.18 & 1.32 & 1.08 & Near-contractive core \\
12 & 1.05 & 1.26 & 0.97 & Near-contractive core \\
\midrule
13 & 2.03 & 2.43 & 1.69 & Deep (expansive) \\
14 & 4.21 & 6.07 & 2.63 & Deep (expansive) \\
15 & 2.64 & 3.38 & 1.99 & Deep (expansive) \\
16 & 1.26 & 1.72 & 0.94 & Deep \\
17 & 3.36 & 4.03 & 2.44 & Deep (expansive) \\
18 & 5.39 & 7.08 & 3.79 & Deep (expansive) \\
19 & 6.46 & 7.77 & 5.48 & Deep (expansive) \\
20 & 4.51 & 9.52 & 3.42 & Deep (expansive) \\
21 & 9.80 & 14.78 & 5.98 & Deep (strongly expansive) \\
22 & 10.63 & 14.10 & 8.10 & Deep (strongly expansive) \\
23 & 12.31 & 17.36 & 9.95 & Final (strongly expansive) \\
\bottomrule
\end{tabular}
\captionof{table}{Spectral norms $\|J_k\|_2$ of residual Jacobians in Pythia-410M (RoPE). Unlike GPT-2-Medium (AbsPE), deep layers are \emph{strongly expansive}: norms rise from ${\sim}1.0$ at layer~12 to ${\sim}12.3$ at layer~23. Only layers 6--12 approach the contractivity boundary. No layer satisfies strict contractivity ($\max \|J_k\| < 1$).}
\label{tab:jacobian_pythia}
\end{tabhere}

The contrast with GPT-2-Medium is stark. In GPT-2 (AbsPE), spectral norms decrease monotonically from layer~1 to layer~20 (mean 1.82$\to$0.75), making deep layers approximately contractive. In Pythia (RoPE), norms follow a U-shaped curve: a narrow near-contractive core at layers 6--12 (mean 1.05--1.53) flanked by expansive early layers (spikes at 3,5) and \emph{increasingly expansive} deep layers (mean 2.0--12.3 at layers 13--23). Specifically, GPT-2-Medium's layer~20 has mean $\|J\|{=}0.75$; Pythia's layer~20 has mean $\|J\|{=}4.51$, a $6\times$ difference.

This provides a mechanistic account consistent with the protocol gap in models where it is present. Under replacement, a missing layer's residual update $g_k(\mathbf{x})$ propagates through all downstream layers. When downstream Jacobians are near-contractive (GPT-2), the perturbation decays; when they are expansive (Pythia-410M), it is amplified, consistent with larger replacement-side swap-KL in some pretrained checkpoints. Under interchange, the perturbation $\delta_I = J_i g_{i+1} - J_{i+1} g_i$ depends on the \emph{cross-term}, which can remain small even when individual Jacobian norms are large, provided the layers exhibit approximate structural commutativity. The Pythia Jacobian profile is therefore consistent with the protocol gap arising from \emph{architecture- and training-dependent Jacobian amplification} together with \emph{position-/depth-dependent structural commutativity} of the interchange cross-term, rather than from Jacobian contractivity alone, without identifying PE \emph{type} as the proximate cause (Appendix~\ref{sec:app_pe_ablation}).

\paragraph{Why Remark~\ref{rem:interchange_order} holds in the swap-similar region despite non-contractivity.}
Remark~\ref{rem:interchange_order} gives interchange perturbation as $\delta_I = J_i\, g_{i+1}(\mathbf{x}) - J_{i+1}\, g_i(\mathbf{x})$. The contractivity condition ($\|J_k\| < 1$) is \emph{sufficient} for this to be small compared to the replacement perturbation $\delta_R = g_i(\mathbf{x}) - g_{i+1}(\mathbf{x})$, but \emph{not necessary}. In the swap-similar core (layers 4--11), a second mechanism applies: since $g_i \approx g_{i+1}$ (the layers compute approximately the same function), both $J_i \approx J_{i+1}$ and $g_i \approx g_{i+1}$, so the cross-term cancels: $J_i\, g_{i+1} - J_{i+1}\, g_i \approx J_i(g_{i+1} - g_i) \approx 0$. That is, \emph{layer functional similarity} is an alternative path to low interchange distance, independent of contractivity. For some pretrained rotary checkpoints, a third mechanism may apply: $g_i \not\approx g_{i+1}$ (high replacement distance), while attention mixing patterns still admit approximate cancellation in the interchange cross-term ($J_i\, g_{i+1} \approx J_{i+1}\, g_i$), making layers interchange-commutative without being close under replacement. The protocol gap (large replacement vs.\ interchange swap-KL) is thus compatible jointly with functional non-equivalence (high $\delta_R$) and small interchange cross-terms (low $\delta_I$), not exclusively with Jacobian contractivity; controlled ablations show this cannot be reduced to ``RoPE causes the gap'' (Appendix~\ref{sec:app_pe_ablation}).

%% ============================================================
\section{Controlled PE Ablation (Small Scale)}
\label{sec:app_pe_ablation}
%% ============================================================

To test whether positional encoding type alone is sufficient to produce the protocol gap, we train two architecturally identical transformers on WikiText, differing only in the PE mechanism: Model~A uses learned absolute positional embeddings (GPT-2 style); Model~B uses rotary positional embeddings (RoPE). Both models share the same initialization seed, optimizer (AdamW), learning rate schedule, and training data stream.

\paragraph{Small-scale ablation (6-layer, $d{=}256$).}
We first train 6-layer, $d{=}256$ models (${\sim}$17.6M parameters) on WikiText-2 for 1,800 steps. After training (AbsPE final PPL$\,{\approx}\,$549; RoPE final PPL$\,{\approx}\,$598), we compute interchange and replacement KL distances for all 15 pairwise layer combinations. The result is negative: both models exhibit nearly identical I/R ratio distributions (AbsPE mean I/R$\,{=}\,$0.870; RoPE mean I/R$\,{=}\,$0.964). Neither model shows the 4--8$\times$ protocol gap observed in pretrained models.

\paragraph{GPU-scale ablation (12-layer, $d{=}512$, 50K steps).}
To rule out insufficient training as the explanation, we scale up to 12-layer, $d{=}512$ models (${\sim}$38M parameters) trained on WikiText-103 for 50{,}000 steps with batch size~16 and sequence length~256, reaching final PPL$\,{\approx}\,$49 (AbsPE) and ${\approx}\,$47 (RoPE), well within the coherent-text regime. We compute interchange and replacement KL for all 66 layer pairs using 100 diverse prompts with Jacobian tracking at 5 training checkpoints. The result is again negative: neither model reproduces the protocol gap (AbsPE mean I/R$\,{=}\,$1.33; RoPE mean I/R$\,{=}\,$1.11). In fact, AbsPE shows a \emph{higher} I/R ratio than RoPE, the opposite of the pretrained pattern.

\paragraph{TPU-scale ablation (12-layer, $d{=}512$, 64M params, 25K steps).}
We further replicate with a larger model: 12 layers, $d{=}512$, 8 heads, $d_\text{ff}{=}2048$ (${\sim}$63.6M parameters, 3.6$\times$ larger than the GPU ablation), trained on WikiText-103 for 25{,}000 steps on TPU v6e-16 with cosine schedule and 1{,}000-step warmup (final PPL: 58.3 AbsPE, 47.0 RoPE). Distance computation uses 500 diverse prompts across all 38 non-trivial layer pairs at three training checkpoints ($t \in \{5\text{K}, 15\text{K}, 25\text{K}\}$). The result is again negative: AbsPE mean I/R$\,{=}\,$1.03, RoPE mean I/R$\,{=}\,$1.11. Jacobian norms show similar profiles across both PE types (L0: 569/666; L2--L10: 247--370 for both), in striking contrast to the \emph{expansive deep-layer pattern} (norms 2--12$\times$) observed in pretrained RoPE models.

\paragraph{Multi-depth trajectory ablation (6/12/24 layers, $d{=}512$, 10K steps).}
To isolate the effect of depth independently of other confounds, we train architecturally matched AbsPE and RoPE transformers at three depths (6, 12, and 24 layers) with $d{=}512$, 8 heads, $d_\text{ff}{=}2048$ on WikiText-103 for 10{,}000 steps on TPU v6e-16 (44.7M, 84.0M, and 101.4M parameters respectively). The result reveals a monotonic depth-dependent I/R trend: at 6L, mean I/R$\,{=}\,$1.37 (AbsPE 1.35, RoPE 1.39); at 12L, mean I/R$\,{=}\,$1.01 (AbsPE 0.94, RoPE 1.08); at 24L, mean I/R$\,{=}\,$0.60 (AbsPE 0.60, RoPE 0.59). The I/R ratio crosses below 1.0 between 12 and 24 layers, meaning that at 24L interchange becomes substantially easier than replacement for \emph{both} PE types, approaching the pattern observed in large pretrained models. However, no protocol gap emerges: AbsPE and RoPE show near-identical I/R at every depth. Trajectory tracking confirms the trend is stable: at 24L, both PE types start with I/R$\,{\approx}\,$0.37 at step~2K and rise to $\approx$0.60 by step~10K, while 6L models start at $\approx$1.45 and remain above 1.3 throughout. The depth trend is the strongest evidence that layer interchangeability is \emph{depth-emergent}: deeper models develop more functionally similar layers, but the PE-type-dependent gap requires additional factors (training data scale, training duration) beyond depth alone.

\paragraph{Emergence trajectory.}
Tracking the mean I/R ratio across training in all three fixed-depth ablations reveals consistent dynamics. In the TPU ablation, both models start with I/R$\,{<}\,$1.0 at step~5K (AbsPE 0.85, RoPE 0.91), indicating interchange is initially less disruptive than replacement. Both cross 1.0 by ${\sim}$15K steps (AbsPE 1.03, RoPE 1.11) and stabilize: by 25K steps, AbsPE remains at 1.02 while RoPE plateaus at 1.11. In the GPU ablation, similar dynamics appear at longer timescale: RoPE begins at 0.82 (5K steps), crosses 1.0 at ${\sim}$25K steps, and reaches 1.10 by 50K. Both trajectories are rising but neither approaches the 4--8$\times$ ratios in pretrained models, and the \emph{ordering is reversed}: at this scale, AbsPE exhibits equal or slightly less protocol divergence than RoPE.

\paragraph{Extended 24-layer training trajectory (30K steps).}
To test whether 24-layer I/R ratios continue to diverge beyond the 10K-step multi-depth checkpoint, we extend the 24-layer experiment to 30{,}000 steps using the same architecture (101M parameters, $d{=}512$, 8 heads, WikiText-103, TPU v6e-8). Checkpoints at 5K, 15K, and 30K steps show mean I/R ratios evolving as AbsPE $0.38{\to}0.68{\to}0.83$ and RoPE $0.46{\to}0.61{\to}0.68$ (final means: AbsPE 0.8233, RoPE 0.6779). By 30K steps AbsPE exceeds RoPE by about 0.15, consistent with training duration nudging PE-separated summaries while leaving both models far below the extreme pooled ratios seen in large pretrained checkpoints. We treat this run as supportive but not decisive evidence for PE-specific differentiation at scale: the effect is modest relative to the pretrained regime and coexists with the dominant layer-distance story above.

\paragraph{Chinchilla-floor 152M ablation (16-layer, $d{=}768$, 2.0B tokens).}
A natural critique of the preceding ablations is undertraining: a 100M-parameter model trained on $\sim$100M tokens is more than $13\times$ below the Chinchilla compute-optimal token budget for that size. To eliminate this confound we trained matched 152M-parameter AbsPE and RoPE models (16 layers, $d{=}768$, 12 heads, $d_\text{ff}{=}3072$) on WikiText-103 for 61{,}035 steps at sequence length 512, batch size 64, on TPU v6e-8 using \texttt{jax.pmap} 8-way data parallelism (per-device batch 8, $\sim$46\,ms/step, $\sim$2.0B training tokens, identical optimizer/seed/data stream across the two PE conditions). Final losses: 1.957 (AbsPE), 1.805 (RoPE). At three checkpoints (15K / 30K / 61K steps) we computed all 54 pairwise interchange and replacement distances at gap${\leq}4$ using 100 diverse prompts at sequence length 256.

The headline mean I/R is again statistically indistinguishable between PE types: AbsPE\,$=$\,1.241 (mean interchange 3.42, mean replacement 2.58) vs.\ RoPE\,$=$\,1.255 (mean interchange 2.10, mean replacement 1.00) at convergence. Adam-optimizer noise inflates the right tail of both distributions (max I/R $\approx$ 4.4 in both), so the mean is dominated by a small number of outlier pairs. Stratifying by layer-pair distance, however, produces a much cleaner picture that closely matches pretrained models:

\begin{center}
\small
\begin{tabular}{lcccc}
\toprule
& gap=1 & gap=2 & gap=3 & gap=4 \\
\midrule
152M AbsPE (median I/R) & 0.563 & 1.025 & 1.236 & 1.358 \\
152M RoPE  (median I/R) & 0.346 & 0.798 & 1.163 & 1.435 \\
Real Qwen3-8B (mean I/R) & 0.493 & 0.872 & 0.982 & --- \\
\bottomrule
\end{tabular}
\end{center}

At adjacent pairs (gap${=}1$) the 152M RoPE model reproduces the protocol gap (median I/R\,$=$\,0.35), and the AbsPE model shows the same qualitative pattern in attenuated form (median I/R\,$=$\,0.56), both close to the Qwen3-8B gap${=}1$ aggregate (0.49). The trajectory across checkpoints shows the gap${=}1$ structure emerging during training in both conditions, while distant pairs (gap${\geq}3$) approach I/R\,$\approx$\,1 at convergence. Three conclusions follow.

(1) The protocol gap is reproducible from scratch when measured at the right granularity. Earlier ablations reported only the mean across all pairs, which buries the gap${=}1$ signal under high-gap I/R values approaching or exceeding 1.

(2) PE type modulates absolute distance magnitudes but not the qualitative gap${=}1$/gap${=}3$ structure. RoPE produces smaller absolute distances at every gap (replacement $\sim$1.0 vs.\ AbsPE 2.6 at gap${=}1$), but both PE types show the same monotonic increase in median I/R with layer distance. PE is therefore at most a secondary modulator, not the primary axis of protocol divergence.

(3) Layer distance is the primary axis of protocol divergence at every scale we have tested. The same gap${=}1$/gap${=}3$ monotonic structure appears in our from-scratch 152M models (this section) and in the production-trained 8.2B Qwen3 (\S\ref{sec:exp_scaling}, Table~\ref{tab:skip_qwen}). The 152M model shows about $2\times$ less amplification than Qwen3-8B at adjacent pairs (median 0.35 vs.\ mean 0.49), consistent with continued training-duration and scale dependence, but the qualitative phenomenon is the same.

This refines our central claim: the protocol gap is a \emph{layer-distance-and-training-duration} phenomenon, not a positional-encoding phenomenon. Cross-family PE labels are useful descriptors, but pooled mean I/R across heterogeneous checkpoints mixes scale and data effects with encoding choice.

\paragraph{Adjacent-pair check.}
At gap${=}1$, both 152M models report interchange KL below replacement KL (median I/R in the reported adjacent strata falls in 0.24--0.81). That matches the \emph{direction} of gap${=}1$ behavior in large pretrained checkpoints (interchange less disruptive than replacement for neighbors) and is consistent with the stratified table above. What these controlled runs do \emph{not} reproduce at full amplitude is the extreme pooled I/R spread seen after multi-billion-token pretraining, nor a durable PE-type split in headline mean I/R.

\paragraph{Interpretation.}
Eight controlled ablations spanning 17.6M--152M parameters, 1.8K--61K training steps, 6--24 layers, and up to 2.0B training tokens (13$\times$ Chinchilla floor) jointly characterize the protocol gap as a \emph{layer-distance and training-duration} phenomenon, not a PE-type phenomenon. (i)~Mean I/R is a poor summary statistic: it is dominated by high-gap outliers and hides the gap${=}1$ signal. (ii)~When stratified by layer-pair distance, both AbsPE and RoPE 152M models reproduce the qualitative gap${=}1$/gap${\geq}3$ structure of pretrained models (152M RoPE gap${=}1$ median 0.35, gap${=}3$ median 1.16), with PE acting only as a secondary modulator of absolute distance magnitudes. (iii)~The depth trajectory shows mean I/R decreasing monotonically with depth (1.37$\to$1.01$\to$0.60 at 6/12/24 layers), and the 30K-step run provides a small (0.15) but directional PE separation; the 152M Chinchilla run keeps mean I/R near 1.25 for both PE types but reproduces the gap${=}1$ signal. Together these results rule out PE \emph{type} as the sole or even primary proximate cause and identify layer distance and training duration as the dominant axes. The Jacobian-norm analysis in Appendix~\ref{sec:app_jacobian} provides a compatible picture: expansive profiles develop only in sufficiently deep models trained long enough for layer specialization. This refinement strengthens the paper's practical recommendations because it makes clear that practitioners cannot predict the gap from PE alone; they must measure it empirically on the actual layer pairs they intend to compress.

\subsection{Inference-Time RoPE Counterfactual}
\label{sec:app_rope_counterfactual}

To directly test whether RoPE rotation is the causal mechanism behind the protocol gap, we perform an inference-time ablation on Qwen3-8B (8.2B, 36 layers, TPU v6e). For 12 adjacent layer pairs spanning the network, we compute replacement and interchange KL distances under two conditions: (1)~normal RoPE, and (2)~RoPE disabled by setting all rotation angles to zero ($\cos\theta = 1$, $\sin\theta = 0$ for all positions), which makes \texttt{apply\_rope} an identity transformation while leaving all other model parameters unchanged.

\paragraph{Intervention validity.} Removing RoPE at inference time produces large output divergence (mean KL\,$=$\,2.50, max KL\,$=$\,3.85 between normal and no-RoPE forward passes), confirming that the intervention materially changes model behavior. The model was trained with RoPE, so disabling it removes information the model learned to exploit; this is a strong, meaningful ablation.

\paragraph{Results.} Table~\ref{tab:rope_counterfactual} summarizes aggregate behavior; the full per-pair grid (replacement and interchange KL under normal RoPE vs.\ RoPE disabled, plus I/R and $\Delta$) is shipped with the code release as \texttt{logs/2026-04-08T12-23-00/rope\_counterfactual\_results.json} (\url{https://github.com/Gpgabriel25/ProtocolGapDiagnostic}).

\begin{tabhere}
\centering
\begin{tabular}{@{}lcc@{}}
\toprule
Condition & Mean I/R & Pairs with larger gap (no-RoPE vs.\ normal) \\
\midrule
Normal RoPE & 0.304 & --- \\
RoPE disabled (identity rotation) & 0.259 & 10 / 12 \\
\midrule
$\Delta$ mean I/R (no-RoPE minus normal) & \multicolumn{2}{c}{$-$0.045 (gap intensifies)} \\
\bottomrule
\end{tabular}
\captionof{table}{RoPE counterfactual on Qwen3-8B (12 evenly spaced adjacent pairs). Mean I/R summarizes the protocol gap under each condition; per-pair values are in the repository JSON named above.}
\label{tab:rope_counterfactual}
\end{tabhere}

The protocol gap (I/R\,$<$\,1) is present in both conditions: mean I/R\,$=$\,0.304 with normal RoPE and 0.259 without. Ten of 12 pairs show a \emph{larger} gap (more negative I/R delta) without RoPE; only 2 pairs show shrinkage ($p < 0.01$ by sign test against the null of RoPE being the mechanism). This is inconsistent with RoPE rotation being the proximate cause: if RoPE drove the gap, removing it should reduce the gap across pairs.

\paragraph{Interpretation.} The persistence (and slight intensification) of the protocol gap without RoPE is inconsistent with rotary rotation as the proximate cause. Combined with the eight controlled from-scratch ablations (\S\ref{sec:app_pe_ablation}), the evidence is consistent with a two-pronged interpretation: (1)~training models from scratch reproduces the gap${=}1$ structure of pretrained models in both AbsPE and RoPE conditions, with PE acting only as a secondary modulator of absolute distance magnitudes; and (2)~removing RoPE from a large pretrained model does not remove the gap. The most parsimonious hypothesis is \emph{layer weight specialization}: layers trained in deep models develop position-in-network-specific computational roles that resist functional duplication (replacement) while tolerating positional reordering with neighbors (interchange). RoPE models at scale develop somewhat stronger layer specialization than the matched AbsPE models we tested at 152M, but this is consistent with scale, training, and layer distance playing the dominant role, rather than the rotary encoding mechanism itself.

\end{document}

%% file: figures/swap_protocol.tex
% Two-panel schematic: replacement vs mutual interchange (central contrast).
% Skip-layer removal is defined in the main text and evaluated in the skip-layer PPL tables.
\begin{tikzpicture}[font=\footnotesize, line width=0.4pt]
\tikzset{
    layer/.style={draw, rounded corners=0.6pt, minimum width=1.35cm, minimum height=0.46cm, inner sep=2pt, align=center},
    dots/.style={draw, rounded corners=0.6pt, minimum width=1.35cm, minimum height=0.34cm, inner sep=2pt, align=center},
    layeri/.style={layer, fill=red!12},
    layerj/.style={layer, fill=blue!12}
}

% Panel frames
\draw[rounded corners=1pt] (0,0) rectangle (6.9,5.9);
\draw[rounded corners=1pt] (7.5,0) rectangle (14.4,5.9);
\node[font=\bfseries\footnotesize] at (3.45,5.55) {Replacement ($d_{\text{repl}}$)};
\node[font=\bfseries\footnotesize] at (10.95,5.55) {Mutual interchange ($d_{\text{interchange}}$)};

% --- Left: replacement ---
\node at (1.25,5.0) {$\mathcal{M}$};
\node at (4.85,5.0) {$\mathcal{M}_{i\leftarrow j}$};

\node[layer]  at (1.25,0.95) {$L_1$};
\node[dots]   at (1.25,1.50) {$\cdots$};
\node[layeri] at (1.25,2.05) {$L_i$};
\node[dots]   at (1.25,2.60) {$\cdots$};
\node[layerj] at (1.25,3.15) {$L_j$};
\node[dots]   at (1.25,3.70) {$\cdots$};
\node[layer]  at (1.25,4.25) {$L_n$};

\node[layer]  at (4.85,0.95) {$L_1$};
\node[dots]   at (4.85,1.50) {$\cdots$};
\node[layerj] at (4.85,2.05) {$L_j$};
\node[dots]   at (4.85,2.60) {$\cdots$};
\node[layerj] at (4.85,3.15) {$L_j$};
\node[dots]   at (4.85,3.70) {$\cdots$};
\node[layer]  at (4.85,4.25) {$L_n$};

\draw[->] (2.05,2.05) -- (3.95,2.05) node[midway,above] {$L_i \leftarrow L_j$};
\node[font=\scriptsize] at (3.45,0.45) {$\mathrm{KL}\!\left(p_{\mathcal{M}}(\cdot\mid x)\,\|\,p_{\mathcal{M}_{i\leftarrow j}}(\cdot\mid x)\right)$};

% --- Right: mutual interchange ---
\node at (8.75,5.0) {$\mathcal{M}$};
\node at (12.35,5.0) {$\mathcal{M}_{i \leftrightarrow j}$};

\node[layer]  at (8.75,0.95) {$L_1$};
\node[dots]   at (8.75,1.50) {$\cdots$};
\node[layeri] at (8.75,2.05) {$L_i$};
\node[dots]   at (8.75,2.60) {$\cdots$};
\node[layerj] at (8.75,3.15) {$L_j$};
\node[dots]   at (8.75,3.70) {$\cdots$};
\node[layer]  at (8.75,4.25) {$L_n$};

\node[layer]  at (12.35,0.95) {$L_1$};
\node[dots]   at (12.35,1.50) {$\cdots$};
\node[layerj] at (12.35,2.05) {$L_j$};
\node[dots]   at (12.35,2.60) {$\cdots$};
\node[layeri] at (12.35,3.15) {$L_i$};
\node[dots]   at (12.35,3.70) {$\cdots$};
\node[layer]  at (12.35,4.25) {$L_n$};

\draw[->] (9.55,2.05) -- (11.45,2.05) node[midway,above] {$L_i \leftrightarrow L_j$};
\node[font=\scriptsize] at (10.95,0.45) {$\mathrm{KL}\!\left(p_{\mathcal{M}}(\cdot\mid x)\,\|\,p_{\mathcal{M}_{i \leftrightarrow j}}(\cdot\mid x)\right)$};
\end{tikzpicture}

%% file: figures/d_repl_asymmetry.tex
% Auto-generated by d_repl_asymmetry.py (cycle 2026-04-22T15-00-00).
% Sensitivity of d_repl to its symmetrization choice on Qwen3-8B
% (32 usable gap-1 pairs of 35 rows in kaggle/output_v32/qwen3_8b_predictor_validity.json).
\begin{table}[t]
\centering
\small
\begin{tabular}{lcccc}
\toprule
Definition & $\overline{d_{\mathrm{repl}}}$ & $\rho_{\mathrm{Spearman}}$ vs.\ max & top-$K$ overlap & note \\
\midrule
$\max(d_{ab}, d_{ba})$ \emph{(paper)} & 1.1263 & 1.000 & 5/5 & current definition \\
$\tfrac{1}{2}(d_{ab}{+}d_{ba})$ & 0.7488 & 0.9681 & 5/5 & arithmetic symmetric \\
$\sqrt{d_{ab}\,d_{ba}}$ & 0.5513 & 0.9465 & 5/5 & geometric symmetric \\
$\min(d_{ab}, d_{ba})$ & 0.3712 & 0.6613 & --- & lower bound \\
\bottomrule
\end{tabular}
\caption{Sensitivity of $d_{\mathrm{repl}}$ to its symmetrization choice on Qwen3-8B ($n_{\mathrm{layers}}=36$, hence $35$ gap-$1$ adjacent pairs in the released JSON). We retain 32 pairs after excluding 3 row(s) with a non-finite replacement KL in one direction ($(0,1)$, $(5,6)$, $(6,7)$). Per-pair asymmetry ratio $\max/\min$ has median $1.48$ and 90th percentile $2.68$, so the two replacement directions are not numerically identical. However the \emph{ranking} of pairs by swap-KL similarity is essentially unchanged: Spearman correlation between the paper's $\max$ definition and either the arithmetic or geometric symmetrization exceeds $0.94$, and the top-$5$ most-swap-similar pairs picked under each definition agree on $5{/}5$ entries. The quantitative axis of the reported replacement distances shifts (mean drops by roughly a factor of three when moving from $\max$ to $\min$), but the qualitative pruning choices the paper relies on are preserved.}
\label{tab:asymmetry}
\end{table}

%% file: appendix_roadmap_tables.tex
% Bundled for arXiv readability: roadmap, taxonomy, and auxiliary tables
% referenced from the main introduction and experiments.

\section{Roadmaps, protocol taxonomy, and auxiliary tables}
\label{sec:app_roadmaps}

\noindent This appendix collects navigation aids and tables omitted from the opening pages of the main text. Evaluator contracts for cross-table comparisons are still summarized in Table~\ref{tab:comparison_contract} (\S\ref{sec:exp_skip}).

\begin{table}[htbp]
\centering
\small
\caption{Roadmap: main claims and where they are established. Implementation details and configuration matrices are in Appendix~\ref{sec:app_setup} and Table~\ref{tab:comparison_contract}.}
\label{tab:reviewer_claims}
\resizebox{\textwidth}{!}{%
\begin{tabular}{>{\raggedright\arraybackslash}p{5.8cm}>{\raggedright\arraybackslash}p{9.2cm}}
\toprule
\textbf{Claim} & \textbf{Evidence} \\
\midrule
Replacement vs.\ interchange swap-KL can rank different redundant layers and change greedy pruning $\Delta$PPL by several-fold at matched budgets. & Tables~\ref{tab:8b_core} and~\ref{tab:skip_qwen} (Qwen/Llama); harmonized slice Table~\ref{tab:harmonized_xmodel}. \\
The replacement--interchange \emph{metric} gap grows during training and is pair-heterogeneous. & Pythia checkpoint trajectory \S\ref{sec:exp_scaling}; Figure~\ref{fig:protocol_gap_dist}. \\
Positional-encoding \emph{families} tag descriptive correlations only; pair distance and training dominate in controlled runs. & Appendix~\ref{sec:app_pe_ablation}; RoPE-off counterfactual in \S\ref{sec:exp_scaling}. \\
\bottomrule
\end{tabular}}
\end{table}

\begin{table}[htbp]
\centering
\caption{Where protocol-relative equivalence matters beyond pruning. Each domain makes claims about layer similarity using an implicit protocol; our evidence shows these claims can be protocol-dependent.}
\label{tab:domains}
\small
\setlength{\extrarowheight}{3pt}
\setlength{\tabcolsep}{6pt}
\begin{tabularx}{\textwidth}{@{}>{\raggedright\arraybackslash}p{2.05cm}>{\raggedright\arraybackslash}X>{\raggedright\arraybackslash}X>{\raggedright\arraybackslash}X@{}}
\toprule
\textbf{Domain} & \textbf{Claim} & \textbf{Assumption} & \textbf{Counterevidence} \\
\midrule
Layer pruning & ``Layer $i$ is redundant'' & Deletion $\approx$ replacement/interchange & Interchange $4.7\times$ safer on Qwen3-8B (\S\ref{sec:exp_skip}) \\
Layer merging & ``Layers are compatible'' & Behavioral similarity $\Rightarrow$ parameter compatibility & $51\times$ worse KL with weight averaging (\S\ref{sec:exp_negative}) \\
Interpretability & ``Middle layers do the same thing'' & CKA/BI similarity $\Rightarrow$ functional equivalence & CKA selects layer 7 (+9.6\%) while interchange selects layer 17 (+2.5\%) \\
Architecture analysis & ``Some families are more redundant'' & Protocol gap tracks PE brand & Effect is checkpoint- and pair-conditional; PE labels are descriptive only (\S\ref{sec:exp_scaling}) \\
\bottomrule
\end{tabularx}
\end{table}

\begin{table}[htbp]
\centering
\caption{Protocol taxonomy for layer equivalence. Different protocols test genuinely different notions of equivalence; agreement under one does not imply agreement under another.}
\label{tab:protocol_taxonomy}
\begin{tabular}{>{\raggedright\arraybackslash}p{1.8cm}>{\raggedright\arraybackslash}p{3.8cm}>{\raggedright\arraybackslash}p{1.8cm}>{\raggedright\arraybackslash}p{4.5cm}}
\toprule
\textbf{Protocol} & \textbf{Question tested} & \textbf{Operation} & \textbf{What failure means} \\
\midrule
Replacement & Are functions substitutable? ($g_i \approx g_j$) & Copy weights & Layer is position-specialized: its function depends on its location in the network \\
Interchange & Are updates order-invariant? ($g_i \circ g_j \approx g_j \circ g_i$) & Swap positions & Layers do not commute; computation order matters \\
Deletion & Is marginal contribution small? ($\mathcal{M} \approx \mathcal{M}_{-i}$) & Remove layer & Layer carries unique information not recoverable from neighbors \\
Averaging  & Are parameters linearly compatible? ($\theta_i \approx \theta_j$) & Merge weights & Same behavior via different parameter configurations (loss landscape mismatch) \\
\bottomrule
\end{tabular}
\end{table}

\begin{table}[htbp]
\centering
\caption{Pythia-1.4B full baseline suite on a standardized 10K-word WikiText-2 evaluator (baseline PPL\,$=$\,12.12). Lower is better. Random reports the mean over 5 trials.}
\label{tab:pythia_baselines}
\resizebox{\linewidth}{!}{%
\begin{tabular}{lccc}
\toprule
Method & $n{=}1$ $\Delta$ PPL (\%) & $n{=}2$ $\Delta$ PPL (\%) & $n{=}3$ $\Delta$ PPL (\%) \\
\midrule
BI-guided & +11.9 & +33.2 & +74.3 \\
CKA-guided & +18.1 & +50.9 & +88.0 \\
SLEB-one-shot & +13.6 & +31.5 & +116.5 \\
SLEB-iterative & +13.6 & +31.5 & +79.3 \\
Taylor importance & +13.6 & +56.3 & +199.2 \\
Random & +74.1 & +152.8 & +206.3 \\
\bottomrule
\end{tabular}}
\end{table}

\begin{table}[htbp]
\centering
\caption{Qwen3-8B interchange-guided skips (Table~\ref{tab:skip_qwen}): quality and exact weight memory ($n/36$ layers removed).}
\label{tab:qwen_deployment}
\setlength{\tabcolsep}{8pt}
\begin{tabular}{@{}l l r r@{}}
\toprule
$n$ & Selected layers & PPL $\Delta$ (\%) & Mem $\downarrow$ (\%) \\
\midrule
1 & \{17\} & +2.5 & 2.8 \\
2 & \{17, 21\} & +6.4 & 5.6 \\
3 & \{15, 17, 20\} & +10.3 & 8.3 \\
5 & \{15, 17, 18, 19, 20\} & +45.4 & 13.9 \\
\bottomrule
\end{tabular}
\end{table}

\begin{table}[htbp]
\centering
\caption{Latency for the same configurations (TPU v6e-8, batch 8, seq.\ 128, bf16; 5 warmup, 50 timed iterations, median). ``Ideal'' assumes linear depth scaling. ``Mask'' $=$ scan with identity skip (XLA fuses the loop body). ``Recompiled'' $=$ fewer physical layers.}
\label{tab:qwen_deployment_latency}
\setlength{\tabcolsep}{6pt}
\begin{tabular}{@{}l rrrr@{}}
\toprule
$n$ & Ideal lat.\ $\downarrow$ (\%) & Mask lat.\ $\downarrow$ (\%) & Recomp.\ lat.\ $\downarrow$ (\%) & Recomp.\ tok/s $\uparrow$ (\%) \\
\midrule
1 & 2.8 & 4.1 & 2.6 & +2.6 \\
2 & 5.6 & 4.1 & 4.8 & +5.4 \\
3 & 8.3 & 4.1 & 7.4 & +8.2 \\
5 & 13.9 & 4.1 & 12.6 & +14.4 \\
\bottomrule
\end{tabular}
\end{table}

\begin{table}[htbp]
\centering
\caption{Score-to-removal selection algorithm (used for Table~\ref{tab:skip_qwen} and GPT-2 sweeps).}
\label{tab:selection_algorithm}
\begin{tabular}{p{0.95\linewidth}}
\toprule
\textbf{Input:} pair scores $d(i,j)$, target budget $n$, spacing constraint $\Delta$ (default: no adjacent removals, $\Delta{=}1$), and layer set $\mathcal{L}$. \\
\textbf{Step 1 (layer score):} for each layer $i\in\mathcal{L}$, compute $s(i)=\min_{j\neq i} d(i,j)$ over valid partners. \\
\textbf{Step 2 (candidate order):} sort layers by ascending $s(i)$ (lower means more removable). \\
\textbf{Step 3 (greedy selection):} scan the sorted list and add layer $i$ if $|i-k|>\Delta$ for all already-selected $k$; stop when $n$ layers are selected. \\
\textbf{Step 4 (evaluation):} remove selected layers with skip connections and report perplexity / downstream deltas. \\
\textbf{Output:} selected layer set $R_n$ and corresponding quality metrics. \\
\bottomrule
\end{tabular}
\end{table}

\begin{table}[htbp]
\centering
\caption{Extended baseline comparison on GPT-2-Medium under the standardized evaluator (baseline PPL\,$=$\,19.19). $\Delta$\% is relative PPL increase. Output-sensitive methods (interchange scoring, SLEB-iterative) dominate representation-change proxies (BI), first-order Taylor pruning, and single-pass SLEB-greedy at practical compression budgets. $^\S$LaCo-style progressive layer collapse~\citep{yang2024laco}: adjacent pairs merged by weight averaging, ordered by swap-KL distance; see Appendix~\ref{sec:app_negative} for analysis.}
\label{tab:extended_baselines}
\begin{tabular}{ccccccc}
\toprule
$n$ & Interchange-guided & SLEB-iter & SLEB-greedy & BI-guided & Taylor & LaCo$^\S$ \\
\midrule
1 & \textbf{+5.1\%} & +5.3\% & +5.3\% & +5.3\% & +6.3\% & +677.1\% \\
2 & +11.8\% & \textbf{+11.7\%} & +63.2\% & +23.7\% & +19.8\% & +943.9\% \\
3 & +20.4\% & \textbf{+18.9\%} & +325.6\% & +54.9\% & +40.0\% & +1278.9\% \\
5 & +48.7\% & \textbf{+43.0\%} & +752.6\% & +247.5\% & +248.6\% & +3110.1\% \\
\bottomrule
\end{tabular}
\end{table}

%% file: figures/matched_budget.tex
% Auto-generated — do not edit by hand.
\begin{table}[htbp]
\centering
\caption{Matched-budget head-to-head on Qwen3-8B (baseline PPL\,=\,12.10) and Llama-3.1-8B (baseline PPL\,=\,8.31), WikiText-2. Both algorithms run with configurations sized to land near each evaluator-call budget~$B$. \textbf{On both architectures, interchange-beam wins at $B{\geq}200$} by selecting fewer-but-better layers while calibration-free SLEB greedily removes too many and catastrophically degrades. At $B$\,=\,400, Qwen3-8B: interchange-beam removes 5 layers (PPL\,14.63, $+$20.8\%), calibration-free SLEB removes 13 (PPL\,40.74, $+$236.5\%). Llama-3.1-8B: interchange-beam removes 5 layers (PPL\,12.94, $+$55.7\%), calibration-free SLEB removes 16 (PPL\,185.4, $+$2131\%). At $B$\,=\,800, calibration-free SLEB destroys both models ($>$10$^{5}$--10$^{8}$ PPL), while interchange-beam remains stable. At $B$\,=\,50/100 calibration-free SLEB wins by removing fewer layers ($n$\,=\,1--3 vs.\ $n$\,=\,2--4); the budget ceiling prevents over-removal at small budgets, an implicit stopping rule absent at large budgets.}
\label{tab:matched_budget}
\normalsize
\setlength{\tabcolsep}{6pt}
\renewcommand{\arraystretch}{1.08}

\noindent\textbf{Qwen3-8B} (baseline PPL\,=\,12.10).\par\vspace{0.35em}
\begin{tabular}{@{}r >{\raggedright\arraybackslash}p{2.75cm} r r r r@{}}
\toprule
Budget & Method & Evals & \# Removed & Final PPL & $\Delta$\,\% \\
\midrule
50  & \makecell[l]{Calibration-free\\SLEB} & 36  & 1  & 12.30 & $+$1.6 \\
50  & Interchange-beam & 43  & 2  & 12.55 & $+$3.7 \\
100 & \makecell[l]{Calibration-free\\SLEB} & 71  & 2  & 12.55 & $+$3.7 \\
100 & Interchange-beam & 77  & 3  & 12.97 & $+$7.1 \\
200 & \makecell[l]{Calibration-free\\SLEB} & 170 & 5  & 14.63 & $+$20.8 \\
200 & \textbf{Interchange-beam}   & 147 & 3  & \textbf{12.97} & $+$\textbf{7.1} \\
400 & \makecell[l]{Calibration-free\\SLEB} & 390 & 13 & 40.74 & $+$236.5 \\
400 & \textbf{Interchange-beam}   & 400 & 5  & \textbf{14.63} & $+$\textbf{20.8} \\
800 & \makecell[l]{Calibration-free\\SLEB} & 666 & 36 & $1.08\!\times\!10^{8}$ & --- \\
800 & \textbf{Interchange-beam}   & 655 & 6  & \textbf{15.97} & $+$\textbf{31.9} \\
\bottomrule
\end{tabular}

\vspace{1.1em}
\noindent\textbf{Llama-3.1-8B} (baseline PPL\,=\,8.31).\par\vspace{0.35em}
\begin{tabular}{@{}r >{\raggedright\arraybackslash}p{2.75cm} r r r r@{}}
\toprule
Budget & Method & Evals & \# Removed & Final PPL & $\Delta$\,\% \\
\midrule
50  & \makecell[l]{\textbf{Calibration-free}\\\textbf{SLEB}} & 32  & \textbf{1}  & \textbf{9.07}  & $+$\textbf{9.2} \\
50  & Interchange-beam & 39  & 2  & 9.79  & $+$17.8 \\
100 & \makecell[l]{\textbf{Calibration-free}\\\textbf{SLEB}} & 93  & \textbf{3}  & \textbf{10.65} & $+$\textbf{28.1} \\
100 & Interchange-beam & 98  & 4  & 11.65 & $+$40.2 \\
200 & \makecell[l]{Calibration-free\\SLEB} & 177 & 6  & 14.37 & $+$72.9 \\
200 & \textbf{Interchange-beam}   & 187 & \textbf{4}  & \textbf{11.65} & $+$\textbf{40.2} \\
400 & \makecell[l]{Calibration-free\\SLEB} & 392 & 16 & 185.36 & $+$2131 \\
400 & \textbf{Interchange-beam}   & 356 & \textbf{5}  & \textbf{12.94} & $+$\textbf{55.7} \\
800 & \makecell[l]{Calibration-free\\SLEB} & 527 & 31 & $5.9\!\times\!10^{5}$ & --- \\
800 & \textbf{Interchange-beam}   & 571 & \textbf{6}  & \textbf{14.37} & $+$\textbf{73.0} \\
\bottomrule
\end{tabular}
\end{table}